\documentclass{article}

\usepackage[final,nonatbib]{neurips_2023}
\usepackage[backref=true, uniquename=false, uniquelist=false, useprefix=true, maxcitenames=1,natbib=true,bibstyle=trad-abbrv, citestyle=authoryear]{biblatex}
\renewbibmacro*{pageref}{%
  \iflistundef{pageref}
    {}
    {\addperiod\addspace\printtext{%
       \ifnumgreater{\value{pageref}}{1}
         {\bibstring{backrefpages}\ppspace}
         {\bibstring{backrefpage}\ppspace}%
       \printlist[pageref][-\value{listtotal}]{pageref}}}}

\renewcommand{\multicitedelim}{\addsemicolon\space}
\DeclareCiteCommand{\cite}[\mkbibparens]{\usebibmacro{prenote}}{\usebibmacro{citeindex}\usebibmacro{cite}}{\multicitedelim}{\usebibmacro{postnote}}
\defbibenvironment{bibliography}
  {\list
     {}
     {\setlength{\leftmargin}{\bibhang}%
      \setlength{\itemindent}{-\leftmargin}%
      \setlength{\itemsep}{\bibitemsep}%
      \setlength{\parsep}{\bibparsep}}}
  {\endlist}
  {\item}
\DeclareFieldFormat{citehyperref}{%
  \DeclareFieldAlias{bibhyperref}{noformat}
  \bibhyperref{#1}}

\DeclareFieldFormat{textcitehyperref}{%
  \DeclareFieldAlias{bibhyperref}{noformat}
  \bibhyperref{%
    #1%
    \ifbool{cbx:parens}
      {\bibcloseparen\global\boolfalse{cbx:parens}}
      {}}}

\savebibmacro{cite}
\savebibmacro{textcite}

\renewbibmacro*{cite}{%
  \printtext[citehyperref]{%
    \restorebibmacro{cite}%
    \usebibmacro{cite}}}

\renewbibmacro*{textcite}{%
  \ifboolexpr{
    ( not test {\iffieldundef{prenote}} and
      test {\ifnumequal{\value{citecount}}{1}} )
    or
    ( not test {\iffieldundef{postnote}} and
      test {\ifnumequal{\value{citecount}}{\value{citetotal}}} )
  }
    {\DeclareFieldAlias{textcitehyperref}{noformat}}
    {}%
  \printtext[textcitehyperref]{%
    \restorebibmacro{textcite}%
    \usebibmacro{textcite}}}

\usepackage[utf8]{inputenc} 
\usepackage[T1]{fontenc}    
\usepackage{hyperref}
\hypersetup{
  colorlinks,
  linkcolor={red!50!black},
  citecolor={blue!50!black},
  urlcolor={blue!50!black}
}
\usepackage{url}            
\usepackage{booktabs}       
\usepackage{amsfonts}       
\usepackage{nicefrac}       
\usepackage{microtype}      
\usepackage{xcolor}         
\usepackage{tikz}
\usetikzlibrary{shapes,snakes,backgrounds,arrows}

\usepackage{url}

\usepackage{caption} 
\usepackage{subcaption} 
\usepackage[inline]{enumitem} 
\usepackage{wrapfig} 
\usepackage[commands,theorems]{./MJH}
\usepackage{bbm}
\usepackage{upgreek}
\usepackage{siunitx}
\usepackage{algorithm}
\usepackage{algorithmic}
\usepackage{multirow}
\usepackage{tabularx}
\usepackage{pifont}
\newcommand{\xmark}{\color{black!50!red}{\ding{55}}}%
\newcommand{\cmark}{\color{black!50!green}{\ding{51}}}%
\usepackage{makecell}

\usepackage{adjustbox}
\usepackage{sidecap} 

\usepackage{cleveref}

\newcommand{\normLigne}[1]{\| #1 \|}
\def\rmD{\mathrm{D}}
\newcommand{\bm}[1]{\mathbbm{#1}}

\def\bfn{\mathbf{n}}
\def\bfk{\mathbf{k}}

\def\bfY{\mathbf{Y}}

\def\bfX{\mathbf{X}}

\def\bfB{\mathbf{B}}

\def\msa{\mathsf{A}}

\def\msk{\mathsf{K}}

\def\msc{\mathsf{C}}

\def\msu{\mathsf{U}}
\def\msv{\mathsf{V}}

\newcommand{\mcb}[1]{\mathcal{B}(#1)}

\def\Rbb{\mathbb{R}}

\def\Pbb{\mathbb{P}}

\def\rset{\mathbb{R}}

\def\nset{\mathbb{N}}

\def\bR{\mathbb{R}}

\def\rmd{\mathrm{d}}

\def\rmc{\mathrm{C}}

\newcommand{\M}{\mathcal M}
 \newcommand{\PE}{\mathbb{E}}

 \newcommand{\absLigne}[1]{\vert #1 \vert}

 \newcommand{\probaLigne}[1]{\mathbb{P}( #1 )}
 \newcommand{\expeLigne}[1]{\PE [ #1 ]}

\def\eqsp{\;}
\newcommand{\coint}[1]{\left[#1\right)}
\newcommand{\ocint}[1]{\left(#1\right]}
\newcommand{\ooint}[1]{\left(#1\right)}
\newcommand{\ccint}[1]{\left[#1\right]}

\newcommand{\metric}{\mathfrak{g}}

\def\Id{\operatorname{Id}}

 \def\bgamma{\bar{\gamma}}
 \newcommand{\ensembleLigne}[2]{\{#1\,:\eqsp #2\}}

 \newcommand{\complementary}{\mathrm{c}}

 \def\vareps{\varepsilon}

\DeclareMathOperator{\intersect}{intersect}
\DeclareMathOperator{\pt}{parallel transport}
\DeclareMathOperator{\reflect}{reflect}

\newcommand{\pms}[1]{\ensuremath{{\scriptstyle\pm #1}}}

\newcommand{\appendixhead}{
  \centerline{\textbf{\LARGE Supplementary to: }\vspace{0.15in}}
  \centerline{\textbf{\LARGE Metropolis Sampling for Constrained Diffusion Models}\vspace{0.25in}}
}
\let\origappendix\appendix 
\renewcommand\appendix{\pagenumbering{arabic}\origappendix}
\AtBeginEnvironment{appendices}{\crefalias{section}{appendix}}

\title{Metropolis Sampling for Constrained Diffusion Models}

\author{%
    Nic Fishman \\
    University of Oxford
\And
    Leo Klarner \\
    University of Oxford
\And
    Emile Mathieu\\
    University of Cambridge
\And
    Michael Hutchinson \\
    University of Oxford
\And
    Valentin De Bortoli \\
    ENS Ulm
}

\makeatother

\bibliography{main}
\begin{document}
\maketitle
\begin{center}
    \vspace{-2.4em}
    \texttt{\{njwfish,leojklarner\}@gmail.com}
    \vspace{0.3em}
\end{center}

\begin{abstract}
  Denoising diffusion models have recently emerged as the predominant paradigm for generative modelling on image domains. In addition, their extension to Riemannian manifolds has facilitated a range of applications across the natural sciences.  
  While many of these problems stand to benefit from the ability to specify arbitrary, domain-informed constraints, this setting is not covered by the existing (Riemannian) diffusion model methodology.
  Recent work has attempted to address this issue by constructing
  novel noising processes based on the reflected Brownian motion and logarithmic barrier methods. However, the associated samplers are either computationally burdensome or only apply to convex subsets of Euclidean space. 
  In this paper, we introduce an alternative, simple noising scheme based on Metropolis sampling that affords substantial gains in computational efficiency
  and empirical performance compared to the earlier samplers. Of independent interest, we prove that this new process corresponds to a valid discretisation
  of the reflected Brownian motion. We demonstrate the scalability and
  flexibility of our approach on a range of problem settings with convex and
  non-convex constraints, including applications from geospatial modelling,
  robotics and protein design.
\end{abstract}

\section{Introduction}

In recent years, denoising diffusion models
\citep{sohl2015deep,song2019generative,song2020score,ho2020denoising} have
emerged as a powerful paradigm for generative modelling, achieving
state-of-the-art performance across a range of domains. They work by
progressively adding noise to data following a Stochastic Differential Equation
(SDE)---the forward \emph{noising} process---until it is close to the invariant
distribution of the SDE. The generative model is then given by an approximation of
the associated time-reversed \emph{denoising} process, which is also an SDE
whose drift depends on the gradient of the logarithmic densities of the forward
process, referred to as the \emph{Stein score}.
Building on the success of diffusion models for image generation tasks,
\citet{debortoli2022riemannian} and \citet{huang2022Riemannian} have recently
extended this framework to a wide range of Riemannian manifolds, enabling the
specification of inherent structural properties of the modelled domain. This
has broadened the applicability of diffusion models to problems in the natural and engineering sciences, including the
conformational modelling of small molecules \citep{jing2022torsional,
  corso2022DiffDock}, proteins
\citep{trippe2022Diffusion,watson2022Broadly,yim2023se} and robotic platforms
\citep{urain2022se}. 

\begin{figure}[t]
\centering
\begin{subfigure}[b]{0.23\textwidth}
  \centering
  \includegraphics[width=\linewidth]{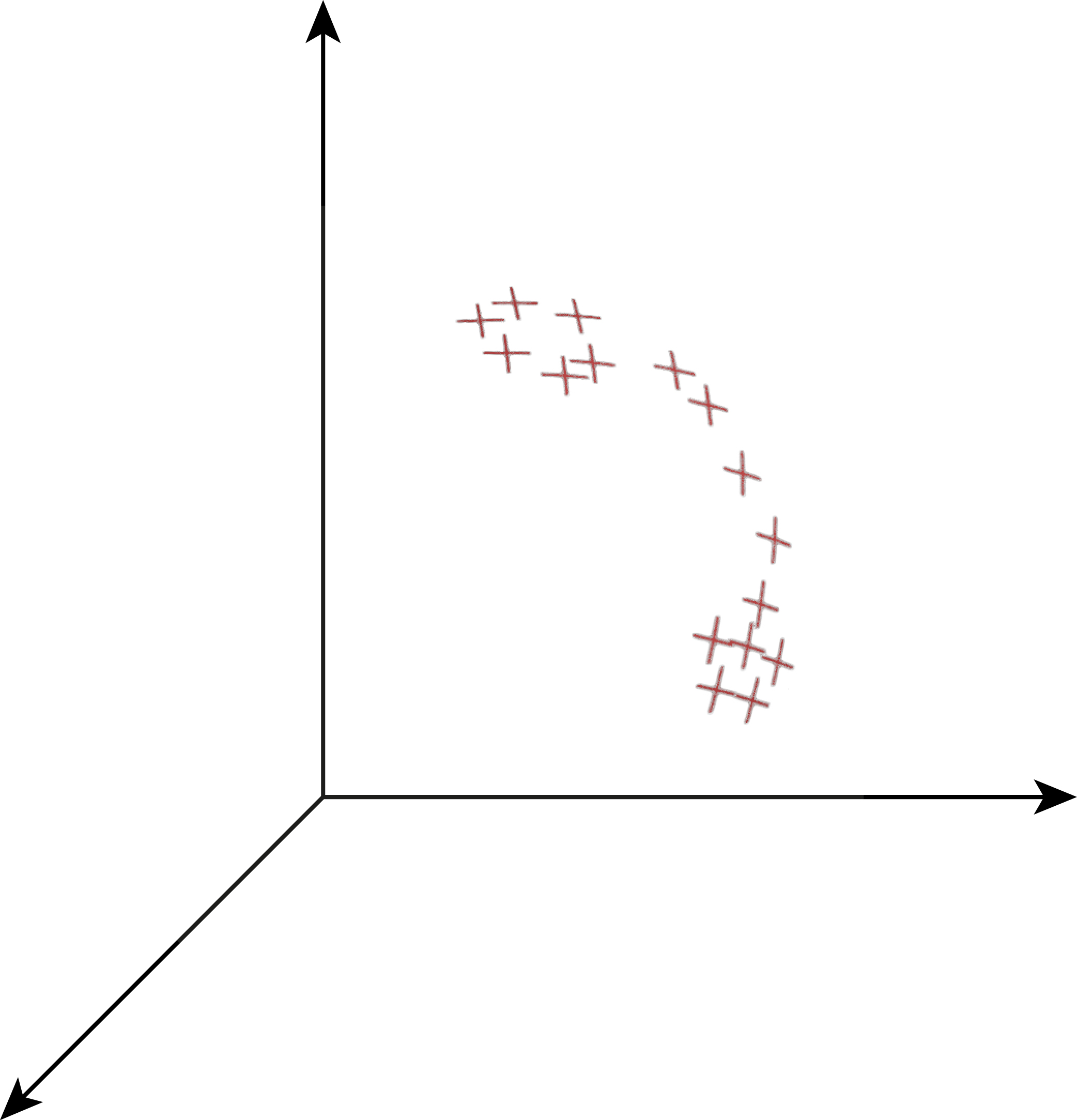}
  \caption{$\Rbb^3$}
  \label{fig:intro_figure_euclidean}
\end{subfigure}
\hfill
\begin{subfigure}[b]{0.23\textwidth}
  \centering
  \includegraphics[width=\linewidth]{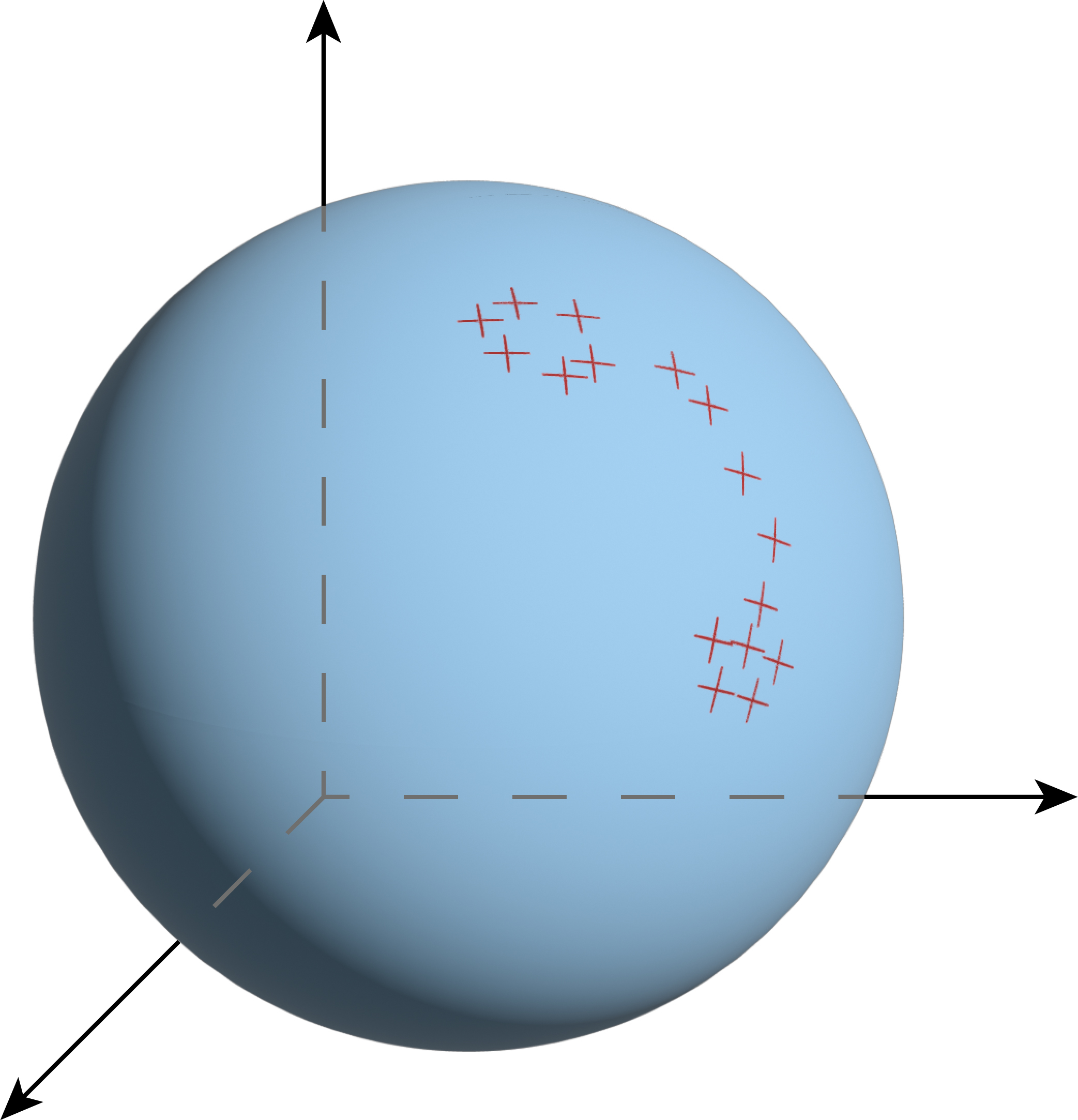}
  \caption{$\mathcal{S}^2\subset\Rbb^3$}
  \label{fig:intro_figure_manifold}
\end{subfigure}
\hfill
\begin{subfigure}[b]{0.23\textwidth}
  \centering
  \includegraphics[width=\linewidth]{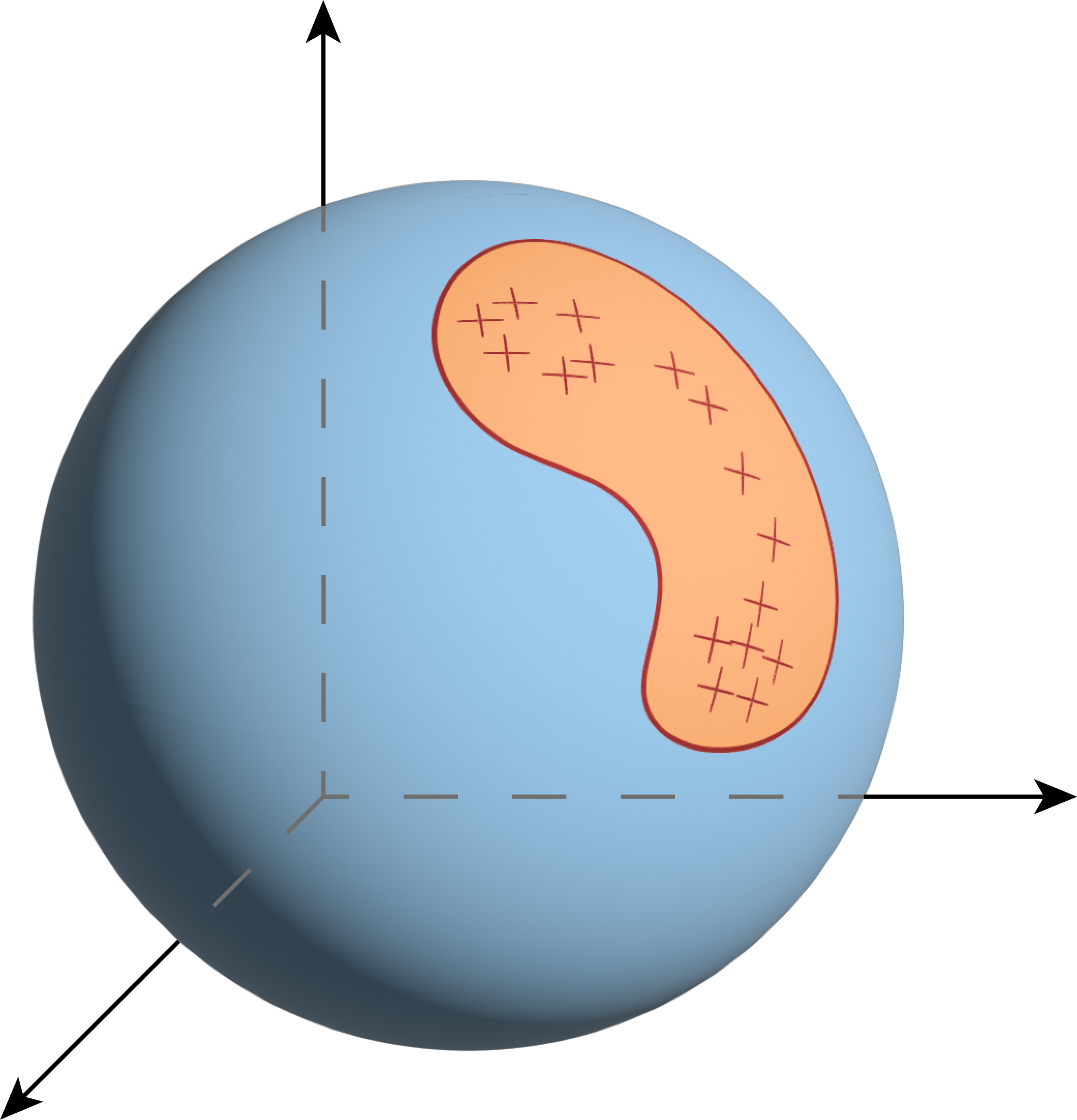}
  \caption{$\mathcal{M}\subset\mathcal{S}^2\subset\Rbb^3$}
  \label{fig:intro_figure_constrained}
\end{subfigure}
\caption{When operating in data-scarce settings, it may often be beneficial to specify as much prior knowledge of the modelled domain as possible.
    Consider a distribution over a subset $\mathcal{M}$ of the unit sphere $\mathcal{S}^2\subset\Rbb^3$. While a Euclidean diffusion model can approximate the distribution in $\Rbb^3$ (a), directly modelling it on $\mathcal{S}^2$ can make learning significantly easier (b).
Restricting the problem space even further by only constructing a diffusion model on $\mathcal{M}$ can lead to even better performance (c). 
}
\label{fig:intro_figure}
\end{figure}

However, in many data-scarce or safety-critical settings, researchers may want to restrict the modelled domain even further by specifying problem-informed constraints to make maximal use of limited experimental data or prevent unwanted behaviour \citep{Morris2002,han2006inverse,Thiele2013,lukens2020practical}. As illustrated in~\Cref{fig:intro_figure}, such domain-informed constraints can be naturally represented as a \emph{Riemannian manifold with boundary}. Training diffusion models on such constrained manifolds is thus
an important problem that requires principled noising processes---and corresponding discretisations---that stay within the constrained set.


Recent work by \citet{fishman2023diffusion} has attempted to derive such
processes and extend the applicability of diffusion models to
inequality-constrained manifolds by investigating the generative modelling
applications of classic sampling schemes based on log-barrier methods
\citep{Kannan2009,lee2017Geodesic,noble2022Barrier,kook2022sampling,gatmiry2022convergence,lee2018convergence}
and the reflected Brownian motion
\citep{williams1987Reflected,petit1997Time,shkolnikov2013Timereversal}.  While
empirically promising, the proposed algorithms can be computationally and
numerically burdensome, and require bespoke implementations for different
manifolds and constraints. Concurrently, \citet{lou2023reflected} have
investigated the use of reflected diffusion models for image applications. They
focus on the high-dimensional hypercube, as this setting admits a theoretically
grounded treatment of the \emph{static thresholding} method which is widely used
in image models such as \citet{saharia2022photorealistic}. More recently, \citet{liu2023mirror} have investigated the use of log-barrier-based mirror maps to transform a constrained domain into an unconstrained dual space for applications in image watermarking. While both methods exhibit robust scaling properties and impressive results, they only consider convex subsets of Euclidean space and do not extend to more general manifolds.

Here, we propose a new method for generative modelling on constrained manifolds
based on a Metropolis-based discretisation of the reflected Brownian motion. The
Metropolised process' chief advantage is that it is lightweight: the only
additional requirement over those outlined in \citet{debortoli2022riemannian}
that is needed to implement a constrained diffusion model is an efficient binary
function that indicates whether any given point is within the constrained
set. This Metropolised approximation of the reflected Brownian motion is substantially easier to implement, faster to compute and more numerically stable
than the previously considered sampling schemes. Our core theoretical
contribution is to show that this new discretisation converges to the reflected
SDE by using the invariance principle for SDEs with boundary
\citep{stroock1971diffusion}. To the best of our knowledge, this is the first
time that such a process has been investigated. We demonstrate that our method
attains improved empirical results on diverse manifolds with convex and non-convex constraints by applying it to a range of problems from geospatial modelling, robotics and protein design.

\section{Background}

\paragraph{Riemannian manifolds.}
A Riemannian manifold is defined as a tuple $(\M, \metric)$ with $\M$ a smooth
manifold and $\metric$ a metric defining an inner product on tangent spaces. In
this work, we will use the exponential map
$\exp_x: \mathrm{T}_x \M \rightarrow \M$, as well as the extension of the
gradient $\nabla$, divergence $\mathrm{div}$ and Laplace $\Delta$ operators to
$\M$. All of these quantities can be defined in local coordinates in terms of the
metric.
The extension of the Laplace operator to $\M$ is
called the Laplace-Beltrami operator, also denoted $\Delta$ when there is no
ambiguity. 
Using $\Delta$, we can define a Brownian motion on $\M$,
denoted $(\bfB_t)_{t \geq 0}$ and with density w.r.t. the volume form of $\M$
denoted $p_t$ for any $ t > 0$. 
We refer to~\Cref{sec:riemannain_intro} for a more detailed exposition, to \citet{lee2013smooth} for a thorough
treatment of Riemannian manifolds and to \citet{hsu2002stochastic} for details
on stochastic analysis on manifolds.
In the following, we consider a constrained manifold $\M$ defined by
\begin{equation} \label{eq:constrained} \c{M} = \ensembleLigne{x \in \c{N}}{f_i(x) < 0 , \ i \in \mathcal{I}}, 
\end{equation}
where $(\c{N}, \metric)$ is a Riemannian manifold, $\mathcal{I}$ is an
arbitrary finite indexing family and for any $i \in \mathcal{I}$,
$f_i\in \rmc(\c{N}, \rset)$. Since $\mathcal{I}$ is finite and $f_i$ continuous
for any $i \in \c{I}$, $\M$ is an open set of $\c{N}$ and inherits its metric
$\metric$. This captures simple Euclidean polytopes and complex constrained spaces like~\Cref{fig:intro_figure}.

\begin{figure}[t]
    \centering
    \includegraphics[width=\textwidth]{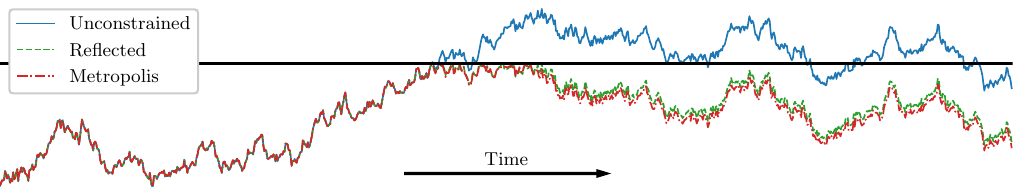}
    \caption{Visual comparison of a discretisation of the unconstrained Brownian
      motion \textcolor{black!60!blue}{(blue)} and two discretisations of the reflected Brownian motion: one based on a reflection scheme \textcolor{black!60!green}{(green)} and the other based on our Metropolis sampler \textcolor{black!60!red}{(red)}. The Metropolised trajectory is very close to that of the reflected one while being significantly easier to implement and cheaper to compute.}
    \label{fig:1d_trajectories}
\end{figure}

\paragraph{Denoising diffusion models.}
\label{sec:sgm}
Denoising diffusion models  \citep{song2019generative,ho2020denoising,song2020score} work as follows: let $(\bfX_{t})_{t \in \ccint{0,T}}$ be a
\textit{noising process} that corrupts the original data distribution $p_0$. We
assume that $(\bfX_t)_{t \geq 0}$ converges to $\mathrm{N}(0,\sigma^2\Id)$, with
$\sigma > 0$. Several such processes exist, but in practice we consider the Ornstein-Uhlenbeck (OU) process, also referred to as VP-SDE, which is defined by the following Stochastic Differential Equation (SDE)
\begin{equation} \label{eq:ornstein_uhlenbeck} \rmd \bfX_t = - \tfrac{1}{2}\bfX_t \rmd t + \sigma \rmd \bfB_t, \qquad \bfX_0 \sim p_0.
\end{equation}
Under conditions on $p_0$, for any $T > 0$,
$(\bfY_t)_{t \in \ccint{0,T}} = (\bfX_{T-t})_{t \in \ccint{0,T}}$ is also the
(weak) solution to a SDE 
\citep{anderson1982reverse,haussmann1986time,cattiaux2021time}
\begin{equation} \label{eq:time_reversal} \textstyle \rmd \bfY_t
  = \{\tfrac{1}{2}\bfY_t + \sigma^2 \nabla \log
  p_{T-t}(\bfY_t)\} \rmd t + \sigma \rmd \bfB_t, \
  \bfY_0 \sim p_T, \end{equation} where $p_t$ denotes the
density of $\bfX_t$. 
%
In practice, $\nabla \log p_t $ is approximated with a score network
$(t,x) \mapsto s_\theta(t,x)$ trained by minimising either a denoising score
matching ($\mathrm{dsm}$) loss or an implicit score matching ($\mathrm{ism}$) loss
\citep{vincent2011connection}
\begin{equation} \label{eq:implicit_sm}
  \textstyle{ \ell(\theta) = \mathbb{E}_{t \sim \mathcal{U}([0,T]), (\bfX_0, \bfX_t)}[  \lambda_t ( \frac{1}{2} \norm{ s_\theta(t, \bfX_t) }^2 + \mathrm{div}(s_\theta)(t, \bfX_t))],}
\end{equation}
where $\lambda_t >0$. For a flexible score network, the global minimiser
$\theta^\star = \mathrm{argmin}_{\theta}\mathcal{L}(\theta)$ satisfies
$s_{\theta^\star}(t, \cdot)=\nabla \log p_t$.  \citet{debortoli2022riemannian}
and \citet{huang2022Riemannian} have extended denoising diffusion models to the
Riemannian setting. The time-reversal formula \eqref{eq:time_reversal} remains
the same, replacing the Euclidean gradient with its Riemannian equivalent. The
\textrm{ism} loss can still be computed in that setting. However, the samplers
used in the Riemannian setting differ from the classical Euler-Maruyama
discretisation used in the Euclidean framework. \citet{debortoli2022riemannian}
use Geodesic Random Walks \citep{jorgensen1975central}, which ensure that the
samples remain on the manifold at every step. In this paper, we propose a
sampler with similar properties in the case of \emph{constrained} manifolds.

\paragraph{Reflected SDE.} We conclude this section by recalling the framework
for studying reflected SDEs, which is introduced via the notion of the
\emph{Skorokhod problem}. For simplicity, we focus on Euclidean space $\rset^d$ here, but note that reflected processes can be defined on arbitrary
smooth manifolds $\mathcal{N}$. In the case
of the Brownian motion, a solution to the Skorokhod problem is a process of the form
$(\bar{\bfB}_t, \bfk_t)_{t \geq 0}$. Locally, $(\bar{\bfB}_t)_{t \geq 0}$ can be
seen as a regular Brownian motion $(\bfB_t)_{t \geq 0}$ while
$(\bfk_t)_{t \geq 0}$ forces $(\bar{\bfB}_t)_{t \geq 0}$ to remain in
$\M$. Under mild additional regularity conditions on $\M$ and
$(\bar{\bfB}_t, \bfk_t)_{t \geq 0}$, see \citet{skorokhod1961stochastic},
$(\bar{\bfB}_t, \bfk_t)_{t \geq 0}$ is a solution to the \emph{Skorokhod
  problem} if for any $t \geq 0$
\begin{equation} \label{eq:rbm}
    \bar{\bfB}_t = \bar{\bfB}_0 + \bfB_t - \bfk_t \in \M,
\end{equation}
$\textstyle{\abs{\bfk}_t = \int_0^t \mathbf{1}_{\bar{\bfB}_s \in \partial \M}
  \rmd \abs{\bfk}_s}$ and
$\textstyle{\bfk_t = \int_0^t \bfn(\bar{\bfB}_s) \rmd \abs{\bfk}_s ,}$ where
$(\abs{\bfk}_t)_{t \geq 0}$ is the total variation of $(\bfk_t)_{t \geq 0}$\footnote{In this case  $(\bfk_t)_{t \geq 0}$ is not regular enough, but if it were in the class $\rmc^1$, its total variation would be given by $\int_0^t \abs{\partial_t \bfk_t} \rmd s$ in the one-dimensional case.}.
Let us provide some intuition on this definition. When
$(\bar{\bfB}_t)_{t \geq 0}$ hits the boundary $\partial \M$, 
$-\bfk_t$ pushes the process back into $\M$ along the inward normal $-\bfn(\bar{\bfB}_t)$,
according to
$\textstyle{\bfk_t = \int_0^t \bfn(\bar{\bfB}_s) \rmd \abs{\bfk}_s}$.  

The
condition
$\textstyle{\abs{\bfk}_t = \int_0^t \mathbf{1}_{\bar{\bfB}_s \in \partial \M}
  \rmd \abs{\bfk}_s}$ is more technical and can be seen as imposing that
$\bfk_t$ remains constant so long as $(\bar{\bfB}_t)_{t \geq 0}$ does not hit
$\partial \M$. We refer to \citet{fishman2023diffusion} and \citet{lou2023reflected} for a more thorough
introduction of these concepts in the context of diffusion models.


\section{Diffusion models for constrained manifolds via Metropolis sampling}
\label{sec:methodology}


In~\Cref{sec:practical_constraints}, we highlight the practical limitations of existing constrained diffusion models and propose an alternative Metropolis sampling-based approach. 
In \Cref{sec:theoretical_convergence}, we outline our proof that this process corresponds to a valid discretisation of the reflected Brownian motion, justifying its use in diffusion models.
An overview of the 
samplers we cover in this section is presented in~\Cref{tab:general_method_comparison}.

\subsection{Practical limitations of existing 
constrained diffusion models}
\label{sec:practical_constraints}

\begin{figure}[t]
    \centering
        \begin{subfigure}[t]{0.48\textwidth}
        \centering
        \includegraphics[width=\textwidth]{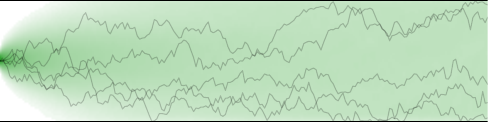}
        \caption{Reflected Brownian motion}
    \end{subfigure}
    \hfill
    \begin{subfigure}[t]{0.48\textwidth}
        \centering
        \includegraphics[width=\textwidth]{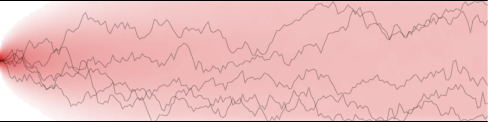}
        \caption{Metropolised approximation of Brownian motion}
    \end{subfigure}
    \caption{Evolution of the density of the reflected Brownian motion and its Metropolis sampling-based approximation on the unit interval starting from a delta mass.}
    \label{fig:1d_densities}
\end{figure}



\paragraph{Barrier methods.}
In the barrier approach, a constrained manifold is transformed into an
unconstrained space via a barrier metric. This metric is defined by
$\nabla^2 \phi(x)$ with $\phi(x) = \sum_{i\in \c{I}} \phi_i(d(x, f_i))$ where
\(d(x, f_i)\) is the minimum distance from the point \(x\) to the set defined by
\(f_i(x) = 0\) and $\phi_i$ is a monotonically decreasing function such that
$\lim_{z \rightarrow 0} \phi_i(z) = \infty$ . Under additional regularity
assumptions, $\phi$ is called a \emph{barrier function} (see \citet{nesterov1994interior}).  This definition ensures that the barrier function induces a well-defined exponential map on the manifold, making the Riemannian
diffusion model frameworks of \citet{debortoli2022riemannian} and
\citet{huang2022Riemannian} applicable.
In the log-barrier method of \citet{fishman2023diffusion}, evaluating $\phi$ requires computing \(d(x, \partial \M)\) (and its derivatives), which can be prohibitively expensive. Furthermore, since the exponential map under the induced manifold is difficult to compute, it is approximated by projecting the exponential map on the original manifold back onto the constraint set, incurring an additional bias.
\citet{liu2023mirror} propose a more tractable method by constructing a mirror map that transforms a constrained domain into an unconstrained dual space, in which one can train a standard Euclidean diffusion model. However, this approach is only applicable to convex subsets of $\bR^d$ and does not extend to arbitrary Riemannian manifolds. More generally, warping the geometry of the modelled domain can adversely impact the interpolative performance of log-barrier-based diffusion models, as the space between data points expands rapidly when approaching the boundary.

\paragraph{Reflected stochastic processes.}
\citet{fishman2023diffusion} and \citet{lou2023reflected} introduce diffusion models based on the
\emph{reflected Brownian motion} (RBM). In \citet{fishman2023diffusion}, the reflected SDE is discretised by \begin{enumerate*}[label=(\roman*)]
\item considering a classical step of the Euler-Maruyama discretization (or the Geodesic Random Walk in the Riemannian setting) and 
\item reflecting this step according to the boundary defined by $\partial \M$.
\end{enumerate*}
To compute the reflection, one must check whether the step crosses the boundary. If it does, the point of intersection needs to be calculated in order to reflect the ray and continue the step in the reflected direction. 
This can require an arbitrarily large number of reflections depending on the step size, 
the geodesic on the manifold, and the geometry of the bounded region within the manifold. 
We refer to \Cref{sec:reflected_discretisation} for the pseudocode of the reflection step and
additional comments.
An alternative approach to discretising a reflected SDE is to replace the reflection with a projection \cite{slominski1994approximation}. However, the projection requires the most expensive part of the reflection algorithm: computing the intersection of the geodesic with the boundary.
\citet{lou2023reflected} propose a more tractable approach that exploits the product structure of the unit hypercube to afford simulation-free sampling but does not extend to arbitrary Riemannian manifolds. Additionally, specifying convex constraints in their framework requires a bijection onto the hypercube, distorting the modelled geometry and incurring the same issues as outlined above.



\paragraph{Metropolis approximation.}
Existing approaches to constrained (Riemannian) diffusion models either only apply to convex subsets of $\bR^d$ or require manifold- and constraint-specific implementations that become computationally intractable as the complexity of the modelled geometry increases.
This limits their practicality even for relatively simple manifolds with well-defined exponential maps and linear inequality constraints such as for example polytopes.
\begin{wrapfigure}{r}{0.42\textwidth}
\vspace{0em}
\begin{minipage}{0.42\textwidth}
    \begin{algorithm}[H]
        \caption{\emph{Metropolis approx. of RBM}}
        \label{alg:metropolis_approx}
        \begin{algorithmic}
        \REQUIRE \(p \in \c{M}\), \(\{f_i\}_{i \in \mathcal{I}}\)
        \STATE Sample \(\v{v} \sim \mathrm{N}(0,\Id) \in \mathrm{T}_p\c{M}\)
        \STATE \(p' \gets \exp_{p}(\v{v})\)
        \IF{\(f_i(p') < 0\ \forall\ i\)}
            \STATE \(p\gets p'\)
        \ENDIF
        \RETURN \(p\)
        \end{algorithmic}
    \end{algorithm}
\end{minipage}
\vspace{1em}
\end{wrapfigure}
In the following, we introduce a method that aims to solve both of these problems.
The sampler we propose is similar to a classical Euler-Maruyama discretisation of
the Brownian motion, except that, whenever a step would carry the Brownian
motion out of the constrained region, we reject it  (see~\Cref{alg:metropolis_approx}).
This is a \emph{Metropolised} version of the usual discretisation and is trivial to implement compared to the
existing barrier, reflection and projection methods.
Hence, this method enables the principled extension of diffusion models to arbitrarily constrained manifolds at virtually \emph{no added implementational complexity or computational expense}.
\renewcommand{\arraystretch}{1.1}
\begin{table}[t]
  \setlength{\tabcolsep}{6.0pt}
  \centering
  \caption{Comparison of the advantages and disadvantages of the different constrained (Riemannian) diffusion models covered in~\Cref{sec:practical_constraints}.}
  \label{tab:general_method_comparison}
  \vspace{1em}
  \adjustbox{width=\textwidth}{
	\begin{tabular}{lcccc}
		& \multicolumn{2}{c}{\(\overbracket{\mspace{200mu}}^{\text{\small Both required for fast DSM loss}}\)} &  & \\ 
		\toprule
		\scshape \bfseries Diffusion Model & \scshape \thead[c]{Tractable \\conditional score} & \scshape \thead[c]{Simulation-free\\forward sampling} & \scshape \thead[c]{Modelling                            \\domain}  & \scshape \thead[c]{Preserves\\metric of \(\c{M}\)} \\
		\midrule
		Reflected Diffusions                                                                                               \\
		\quad\scshape \textcite{lou2023reflected}         & \cmark                    & \cmark & \(\text{\small convex}\;{\subset}\;\mathbb{R}^d\)       & \xmark \\
  		\quad\scshape \textcite{fishman2023diffusion}  & \xmark                    & \color{black!50!red}\(\c{O}(d^2)\) & Any \(\c{M}\)             & \cmark \\
		\quad\scshape Metropolis (ours) & \xmark                    & \color{black!50!red}\(\c{O}(d)\) & Any \(\c{M}\)             & \cmark \\

		\midrule
		Barrier Diffusions                                                                                                 \\
		\quad\scshape \textcite{fishman2023diffusion}  & \xmark                    & \xmark \hspace{0.1em} & \(\text{\small convex}\;{\subset}\;\text{any}\;\c{M}\)             & \xmark \\
		\quad\scshape \textcite{liu2023mirror}            & \cmark                    & \cmark & \(\text{\small convex}\;{\subset}\;\mathbb{R}^d\) & \xmark \\

		\bottomrule
	\end{tabular}
  } 
\end{table}
\renewcommand{\arraystretch}{1}

\subsection{Relating the Metropolis sampler to the reflected Brownian motion}
\label{sec:theoretical_convergence}
   

In this section, we prove that the proposed Metropolis sampling-based process (\Cref{alg:metropolis_approx}) corresponds to a valid discretisation of the reflected process, justifying its use in diffusion models. 
We present a concise overview of the core concepts
and main results here and postpone the full proof to \Cref{sec:rejection-convergence}. For simplicity, we focus on the Euclidean setting and discuss the assumptions our proof requires on $\M$, as well as its extension to more general manifolds, at the end of this section.
We begin with a definition of the
Metropolis approximation of RBM.

\begin{definition}
  For any $\gamma >0$ and $k \in \nset$, let $X_0^\gamma \in \M$ and
  $X_{k+1}^\gamma = X_k^\gamma + \sqrt{\gamma} Z_k^\gamma$ if
  $X_k^\gamma + \sqrt{\gamma} Z_k^\gamma \in \M$ and $X_k^\gamma$
  otherwise. The sequence $(X_k^\gamma)_{k \in \nset}$ is called the {Metropolis approximation of RBM}.
\end{definition}
For any $\gamma > 0$, we consider $(\bfX_t^\gamma)_{t \geq 0}$, the linear
interpolation of $(X^\gamma_k)_{k \in \nset}$ such that for any $k \in \nset$,
$\bfX_{k \gamma}^\gamma = X_k^\gamma$. The following result is the main theoretical
contribution of our paper.

\begin{theorem}
  \label{thm:weak_convergence_metropolis}
  Under assumptions on $\M$, 
  for any $T \geq 0$, $(\bfX_t^\gamma)_{t \in \ccint{0,T}}$ weakly converges to the RBM 
  $(\bar{\bfB}_t)_{t \in \ccint{0,T}}$ as $\gamma \to 0$.
\end{theorem}

The rest of the section is devoted to a high level presentation of the proof of
\Cref{thm:weak_convergence_metropolis}.  It is theoretically impractical to work
directly with the Metropolis approximation of RBM. Instead, we introduce an
auxiliary process, show this converges to the RBM, and finally prove that the
convergence of the auxiliary process implies the convergence of our Metropolis
discretisation.


\begin{definition}
  For any $\gamma > 0$ and $k \in \nset$, let 
  $\hat{X}_0^\gamma = x \in \M$ and 
  $\hat{X}_{k+1}^\gamma = \hat{X}_k^\gamma + \sqrt{\gamma} Z_k^\gamma$ with
  $Z_k^\gamma$ a Gaussian random variable conditioned on
  $\hat{X}_k^\gamma + \sqrt{\gamma} Z_k^\gamma \in \M$. The sequence
  $(\hat{X}_k^\gamma)_{k \in \nset}$ is called the Rejection approximation of RBM.
\end{definition}

\begin{wrapfigure}{R}{0.42\textwidth}
\vspace{-0.5em}
\begin{minipage}{0.42\textwidth}
    \begin{algorithm}[H]
        \caption{\emph{Rejection approx. of RBM}}
        \label{alg:rejection_sampling}
        \begin{algorithmic}
        \REQUIRE \(p \in \c{M}\), \(\{f_i\}_{i \in \mathcal{I}}\)
        \STATE Sample \(\v{v} \sim \mathrm{N}(0,\Id) \in \mathrm{T}_p\c{M}\)
        \STATE \(p' \gets \exp_{p}(\v{v})\)
        \WHILE{\(f_i(p') \geq 0\ \text{for any}\ i\)}
            \STATE Sample \(\v{v} \sim \mathrm{Id}(0,1) \in \mathrm{T}_p\c{M}\)
            \STATE \(p' \gets \exp_{p}(\v{v})\)
        \ENDWHILE
        \RETURN \(p'\)
        \end{algorithmic}
    \end{algorithm}
\end{minipage}
\vspace{-2em}
\end{wrapfigure}

We call this process \emph{Rejection approximation of RBM} since in practice, $Z_k^\gamma$ is
sampled using rejection sampling, see~\Cref{alg:rejection_sampling}. For any
$\gamma > 0$, we also consider $(\hat{\bfX}_t^\gamma)_{t \geq 0}$, the linear
interpolation of $(\hat{X}^\gamma_k)_{k \in \nset}$ such that for any
$k \in \nset$, $\hat{\bfX}_{k \gamma}^\gamma = \hat{X}_k^\gamma$. In~\Cref{sec:rejection-convergence}, we prove the following result.


\begin{theorem}
  \label{thm:weak_convergence_rejection}
  Under assumptions on $\M$, 
  for any $T \geq 0$, $(\hat{\bfX}_t^\gamma)_{t \in \ccint{0,T}}$ weakly converges to the Reflected Brownian Motion
  $(\bar{\bfB}_t)_{t \in \ccint{0,T}}$ as $\gamma \to 0$.
\end{theorem}


\begin{proof} 
  Our approach is
  based on the invariance principle of \citet{stroock1971diffusion}. More
  precisely, we show that we can compute an equivalent `drift' and `diffusion
  matrix' for the discretised process and that, as $\gamma \to 0$, the drift
  converges to zero and the diffusion matrix converges to $\Id$. In the
  Euclidean setting, this result, accompanied by mild regularity and growth
  assumptions, ensures that the discretization weakly converges to the original SDE. 
  However, the case with boundary is much more complicated, primarily
  because the approximate drift might explode near the boundary, thus we need to
  verify exactly how the drift behaves as $\gamma \to 0$ and as the process
  approaches the boundary. We show that the \emph{normalised} drift converges to the inward normal near the boundary. This ensures that \begin{enumerate*}[label=(\alph*)]
  \item in the interior of $\M$ the drift converges to zero, i.e.~locally in the
    interior of $\M$ the Brownian motion and the Reflected Brownian Motion
    coincide,
      \item on the boundary, the drift pushes the samples inside the manifold according to the inward normal, mimicking $(\bfk_t)_{t \geq 0}$ in \eqref{eq:rbm}.
   \end{enumerate*}
Finally, with results from \citet{stroock1971diffusion} and \citet{kang2017submartingale}, we show the convergence to the RBM.
Full details and derivations are provided in \Cref{sec:rejection-convergence}.
\end{proof}

Our next step is to show that the approximate drift and diffusion matrix of the Metropolised
process are upper and lower bounded by their counterparts
in the rejection process. While the upper-bound is easy to derive, the
lower-bound requires the following result.

\begin{proposition}
  \label{lemma:lower_bound_prob_main}
  Under assumptions on $\M$, $\forall \; \vareps > 0$, $\exists \; \bgamma >0$ such that for any
  $\gamma \in \ooint{0, \bgamma}$ and for any $x \in \M$,
  $\gamma \in \ooint{0, \bar{\gamma}}$ and $Z \sim \mathrm{N}(0, \Id)$ we have
  $\mathbb{P}(x + \sqrt{\gamma} Z \in \M) \geq 1/2-\vareps$, with
  $Z \sim \mathrm{N}(0, \Id)$.
\end{proposition}

\Cref{lemma:lower_bound_prob_main} tells us that \emph{locally} the boundary
looks like a half-space when integrating w.r.t. a Gaussian measure. A
corollary is that, for $\gamma > 0$ small enough and for any $k \in \nset$, the
probability that $X_{k+1}^\gamma = X_k^\gamma$ is upper bounded \emph{uniformly}
w.r.t. $X_k^\gamma \in \M$. The proof of \Cref{lemma:lower_bound_prob_main} uses
\Cref{thm:tubul-neighb} in~\Cref{sec:rejection-convergence}, whose proof relies
on the concept of \emph{tubular neighborhoods} \citep{lee2012smooth}.
Having established the lower and upper bound, we can conclude the proof by
noting that the approximate drift and the diffusion matrix in the rejection
and Metropolis case coincide as $\gamma \to 0$. This is enough to apply the
same results as before, giving the desired convergence.

\paragraph{Assumptions on $\M$.} Before concluding this section, we detail the
assumptions we make on $\M$. For \Cref{thm:weak_convergence_metropolis} to hold,
we assume that $\M = \ensembleLigne{x \in \rset^d}{\Phi(x) > 0}$ is bounded, 
with $\Phi \in \rmc^2(\rset^d, \rset)$ concave. We have that
$\partial \M = \ensembleLigne{x \in \rset^d}{\Phi(x) = 0}$. In addition, we
assume that for any $x \in \partial \M$, $\normLigne{\nabla \Phi(x)} = 1$. These
assumptions match those \citet{stroock1971diffusion} use for their study of the
existence of solutions to the RBM. While it seems possible to relax the
\emph{global} existence of $\Phi$ to a \emph{local} one, the regularity
assumption of the domain is key. This regularity is essential to
establish \Cref{lemma:lower_bound_prob_main} and the associated geometrical
result on tubular neighbourhoods. We also emphasize that the smoothness of the
domain is central in the results of \citet{kang2017submartingale} on the
equivalence of two definitions of RBMs which we rely on.
An extension of these results to a more general class of manifolds defined via the inequality constraints $f_i(x)<0$ and $f_i:\mathcal{M} \to\mathbb{R}$ is straightforward, yet highly technical, and hence we postpone a full derivation to future work. Note that contrary to the previous case, each $f_i$ is defined on a manifold. To take the underlying geometry of $\mathcal{M}$ into account, we need to extend the Taylor expansion of the function $\Phi$ (now $f_i$) to the manifold setting, see \citet[Lemma~S.8]{durmus2021riemannian}, for instance. We also use the notion of the tubular neighbourhood to decompose the space in a tangential and a normal part, which is still valid in the manifold setting \cite{lee2018introduction}. In all of the proofs (and the assumptions of \citet[Theorem~6.3]{stroock1971diffusion}) the norm between elements should be replaced by the Riemannian metric of $\mathcal{M}$. Finally, one needs to extend the proof of \citet{stroock1971diffusion}, which is possible since it only uses smoothness arguments that can be extended to the manifold setting \cite[Lemma~S.8]{durmus2021riemannian}.

\section{Related work on approximations of reflected SDEs}

Several schemes have been introduced to approximately sample from reflected
Stochastic Differential Equations. They can be interpreted as modifications of
classical Euler-Maruyama schemes used to discretise SDEs without
boundary. One of the most common approaches is to use the Euler-Maruyama
discretisation and project the solution onto the boundary if it escapes from the
domain $\M$. 
In this case, mean-square error rates of order \emph{almost} $1/2$
have been proven under various conditions
\citep{liu1995discretization,chitashvili1981strong,pettersson1995approximations,slominski1994approximation}. 
Concretely this means that
$\mathbb{E}[\| \bar{\bfB}_{t} - X_{n}^{t/n}\|^2] = O(n^{-1+\varepsilon})$ with
$\vareps>0$ arbitrary small and where $(X_k^\gamma)_{k \in \nset}$ is the
projection scheme.
The rate $1/2$ is tight 
\citep{pacchiarotti1998numerical}.
\citet{liu1993numerical} introduced a \emph{penalised}
method which pushes the solution away from the boundary and shows a mean-square error 
of order $1/4$, see also \citet{pettersson1997penalization}. Weak errors of order
$1$ have been obtained in \citet{bossy2004symmetrized} and \citet{gobet2001euler} by
introducing a reflection component in the discretisation or using some local
approximation of the domain to a half-space, see also \citet{pilipenko2014introduction}.
Closer to our work, \citet{burdzy2008discrete} consider three
different methods to approximate reflected Brownian motions on general domains
(two based on discrete methods and one based on killed diffusions). Only
qualitative results are provided. To the best of our knowledge, no previous work in the
probability literature has investigated the \emph{Metropolised} scheme we
propose. 
Our Metropolis scheme is also related to the ball walk \citep{applegate1991sampling}, which replaces the Gaussian random variable with a uniform over the ball (or the Dikin ellipsoid). \citet{applegate1991sampling} and \citet{lovasz2007geometry} have studied the asymptotic convergence rate of the ball walk, but, to the best of our knowledge, its limiting behaviour when the step size goes to zero has not been investigated.

\begin{table}[t]
    \centering
    \caption{Log-likelihood ($\uparrow$) of a held-out test set from a synthetic bimodal distribution over convex subsets of $\Rbb^d$ bounded by the hypercube \([-1,1]^d\) and unit simplex \(\Delta^d\). Means and standard deviations are computed over 3 different runs. Average training time is provided in hours.}
    \label{tab:synthetic_polytope}
    \vspace{1em}
    \setlength{\tabcolsep}{8pt}
    \begin{adjustbox}{max width=\textwidth}
    \begin{tabular}{cccccc}
    \toprule
     \multirow{2}{*}{\scshape Manifold} & \multirow{2}{*}{\scshape Dimension} & \multicolumn{2}{c}{\scshape Reflected} & \multicolumn{2}{c}{\scshape Metropolis} \\
     & & log-likelihood & runtime & log-likelihood & runtime \\
    \midrule 
    \multirow{3}{*}{$[-1,1]^d$} 
     &  2 & $2.25\pms{.01}$ & $8.95$ & $\mathbf{2.32\pms{.05}}$ & $\mathbf{0.72}$\\
     & 3 & $3.77\pms{.13}$ & $8.97$ & $\mathbf{4.15\pms{.15}}$ & $\mathbf{0.71}$\\
     & 10 & $7.42\pms{.77}$ & $10.1$ & $\mathbf{10.80 \pms{.34 }}$ & $\mathbf{0.90}$\\
    \midrule
    \multirow{3}{*}{$\Delta^d$} 
     & 2 & $1.01 \pms{.01}$ & $9.17$ & $\mathbf{1.06\pms{.02}}$ & $\mathbf{0.82}$ \\
     & 3 & $2.64\pms{.01}$ & $9.43$ &  $\mathbf{3.23\pms{.17}}$ &  $\mathbf{0.78}$ \\
     & 10 &  $7.00\pms{.13}$ & $10.5$ & $\mathbf{7.81\pms{.20}}$ & $\mathbf{0.97}$ \\
    \bottomrule
    \end{tabular}
    \end{adjustbox}
\end{table}

\begin{figure}
    \centering
    \label{convergence_results}
    \vspace{-1.35em}
    \includegraphics[width=0.9\textwidth]{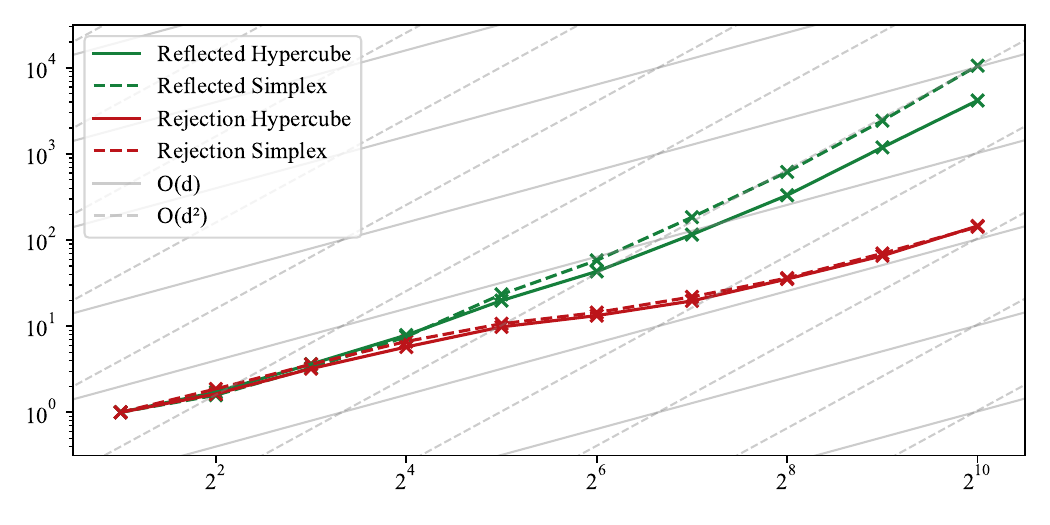}
    \put(-365, 35){\rotatebox{90}{\normalsize Convergence time rel. to \(d=1\)}}
    \put(-220, 0){\normalsize Dimension of manifold}
    \caption{Convergence time of the Reflected \textcolor{black!60!green}{(green)} and Metropolis \textcolor{black!60!red}{(red)} forward noising processes to the uniform distribution on the hypercube $[-1,1]^d$ and unit simplex $\Delta^d$. The lines indicate functions fit with the \textsc{PySR} symbolic regression package \cite{cranmer2023interpretable} and correspond to empirical runtime complexities of $\mathcal{O}(d^2)$ and $\mathcal{O}(d)$, respectively, matching the superimposed scaling law isocontours.
    }
    \label{fig:forward_process_timing}
\end{figure}

\section{Experimental results}

To demonstrate the practical utility and empirical performance of the proposed Metropolis diffusion models, we conduct a comprehensive evaluation on a range of synthetic and real-world tasks. 
In~\Cref{sec:exp_synthetic_tasks}, we assess the scalability of our method by applying it to synthetic distributions on hypercubes and simplices of increasing dimensionality.
In~\Cref{sec:exp_arms_and_backbones}, we extend the evaluation to real-world tasks on manifolds with convex constraints by applying our method to the robotics and protein design datasets presented in \citet{fishman2023diffusion}.
In~\Cref{sec:exp_wildfires}, we additionally demonstrate that our method extends to constrained manifolds with highly \emph{non-convex} boundaries---a setting that is intractable with existing approaches. 

As we found---in line with \citet{fishman2023diffusion}---that log-barrier diffusion models perform strictly worse than reflected approaches across all experimental settings, we focus on a more detailed comparison
with the latter here and postpone additional empirical results to~\Cref{sec:app_exp_barrier_results}. These include additional performance metrics and a comparison to an unconstrained Euclidean diffusion model on the synthetic datasets presented in~\Cref{sec:exp_synthetic_tasks}. 

For all experiments, we use a simple 6-layer MLP with sine activations and a score rescaling function to ensure that the score reaches zero at the boundary, scaling linearly into the interior of the constrained set as in \citet{liu2022estimating} and \citet{fishman2023diffusion}. We set $T=1$, $\beta_0=\num{1e-3}$ and tune $\beta_1$ to ensure that the forward process reaches the invariant distribution with a linear $\beta$-schedule. We use a learning rate of \num{2e-4} with a cosine learning rate schedule and an $\mathrm{ism}$ loss with a modified loss weighting function of $(1 + t)$, a batch size of 1024 and 8 repeats per batch. All models were trained on a single NVIDIA GeForce GTX 1080 GPU.
Additional details are provided in~\Cref{sec:app_implementational_details}. 

All source code that is needed to reproduce the results presented below is made available under \href{https://github.com/oxcsml/score-sde/tree/metropolis}{https://github.com/oxcsml/score-sde/tree/metropolis}, which requires a supporting package to handle the different geometries that is available under \href{https://github.com/oxcsml/geomstats/tree/polytope}{https://github.com/oxcsml/geomstats/tree/polytope}.

\subsection{Synthetic distributions on simple polytopes}
\label{sec:exp_synthetic_tasks}


In this section, we investigate the scalability of the proposed Metropolis diffusion models by applying them to synthetic bimodal distributions over the \(d\)-dimensional hypercube \([-1, 1]^d\) and unit simplex \(\Delta^{d}\). A quantitative comparison of the log-likelihood of a held-out test set is presented in~\Cref{tab:synthetic_polytope}, while a visual comparison is postponed to~\Cref{sec:app_exp_synthetic_tasks}.
We find that our Metropolis models outperform reflected approaches across all dimensions and constraint geometries by a substantial and statistically significant margin while training in one tenth of the time. 
The degree of improvement seems to scale with the dimensionality of the problem: the larger the dimension of the experiment, the larger the gain in performance from using our proposed Metropolis scheme.

We observe a similar difference in the scaling properties of reflected and Metropolis models when measuring the convergence times of the respective forward noising processes to the uniform distribution on hypercubes \([-1, 1]^d\) and simplices \(\Delta^{d}\) of increasing dimensionality. The results are presented in~\Cref{convergence_results} and show that the convergence time of the Metropolis process scales linearly in the dimension, while the reflected process scales quadratically.

\begin{figure}[t]
  \centering
  \begin{subfigure}[t]{0.23\textwidth}
    \centering
    \includegraphics[width=\linewidth]{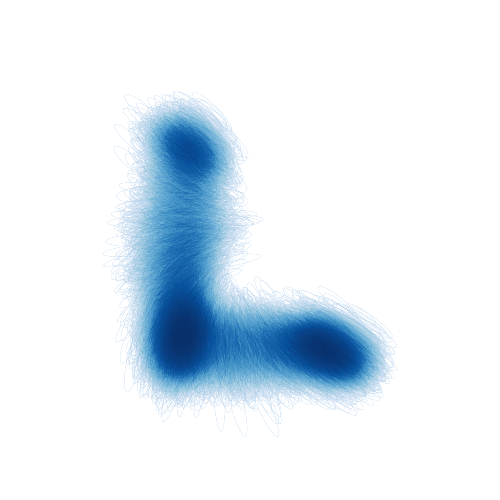}
    \caption{Data distribution.}
    \label{fig:robotic_arm_data}
  \end{subfigure}
  \hfill
  \begin{subfigure}[t]{0.23\textwidth}
    \centering
    \includegraphics[width=\linewidth]{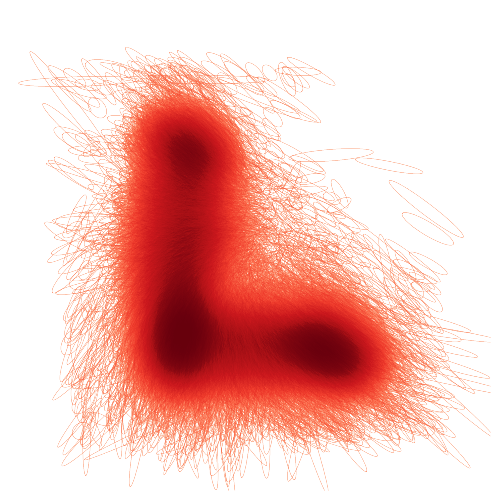}
    \caption{Metropolis samples.}
    \label{fig:robotic_arm_rejection}
  \end{subfigure}
  \hfill
  \begin{subfigure}[t]{0.23\textwidth}
    \centering
    \includegraphics[width=\linewidth]{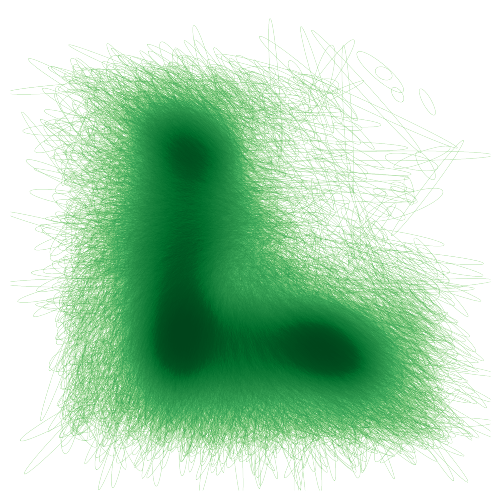}
    \caption{Reflected samples.}
    \label{fig:robotic_arm_reflected}
  \end{subfigure}
  \hfill
  \begin{subfigure}[t]{0.23\textwidth}
    \centering
    \includegraphics[width=\linewidth]{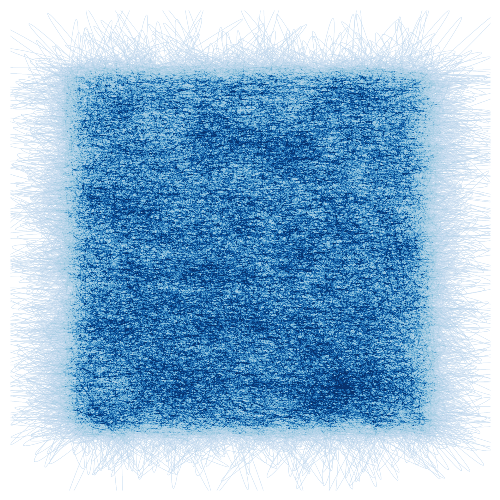}
    \caption{Uniform distribution.}
    \label{fig:robotic_arm_uniform}
  \end{subfigure}
  \caption{A qualitative visual comparison of $10^6$ samples from the data distribution, our Metropolis diffusion model, a reflected diffusion model and the uniform distribution for the constrained configurational modelling of robotic arms on \(\c{S}_{++}^2\times \Rbb^2\).}
  \label{fig:robotic_arm}
\end{figure}

\subsection{Modelling proteins and robotic arms under convex constraints}
\label{sec:exp_arms_and_backbones}

In addition to illustrating our method's scalability on high-dimensional synthetic tasks, we follow the experimental setup from \citet{fishman2023diffusion} to additionally demonstrate its practical utility and favourable empirical performance on two real-world problems from robotics and protein design.

\paragraph{Constrained configurational modelling of robotic arms.} The problem of modelling the configurations and trajectories of a robotic arm can be formulated as learning a distribution over the locations and manipulability ellipsoids of its joints, parameterised on \(\Rbb^d\times\c{S}_{++}^d\), where \(\c{S}_{++}^d\) is the manifold of symmetric positive-definite (SPD) \(d\times d\) matrices \citep{yoshikawa1985manipulability, jaquier2021geometry}. For practical robotics applications, it may be desirable to restrict the maximal velocity with which a robotic arm can move or the maximum force it can exert. This manifests in a trace constraint $C>0$ on \(\c{S}_{++}^d\), resulting in a constrained manifold \(\{M\in\c{S}_{++}^d:\sum_{i=1}^dM_{ii} < C\}\). Following \citet{fishman2023diffusion}, we parametrise this constraint via the Cholesky decomposition \citep{lin2019riemannian} and use the resulting setup to model the dataset presented in \citet{jaquier2021geometry}. 


\begin{figure}[t]
  \centering
  \begin{subfigure}[t]{0.23\textwidth}
    \centering
    \includegraphics[width=\linewidth]{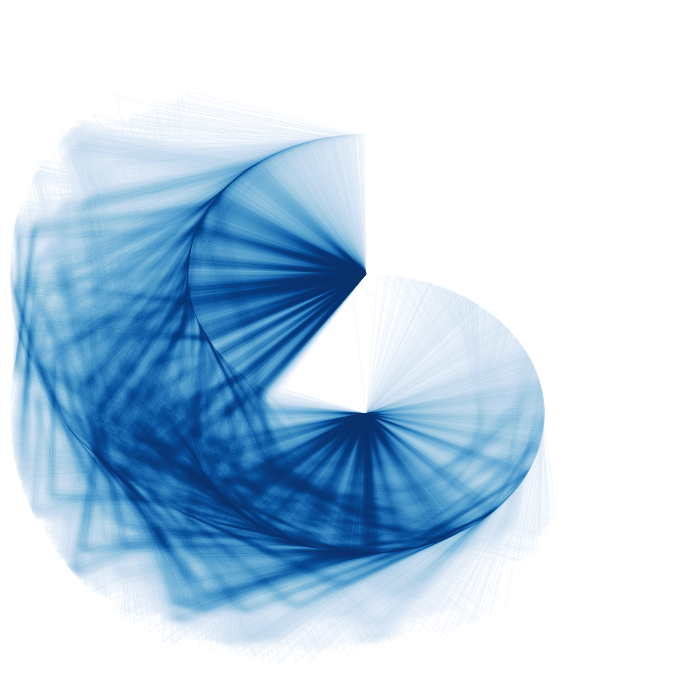}
    \caption{Data distribution.}
    \label{fig:loop_data}
  \end{subfigure}
  \hfill
  \begin{subfigure}[t]{0.23\textwidth}
    \centering
    \includegraphics[width=\linewidth]{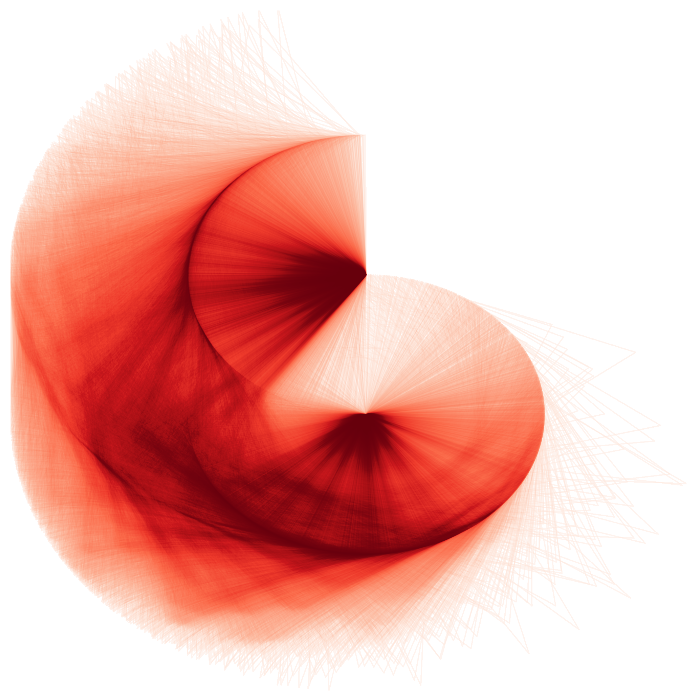}
    \caption{Metropolis samples.}
    \label{fig:loop_rejection}
  \end{subfigure}
  \hfill
  \begin{subfigure}[t]{0.23\textwidth}
    \centering
    \includegraphics[width=\linewidth]{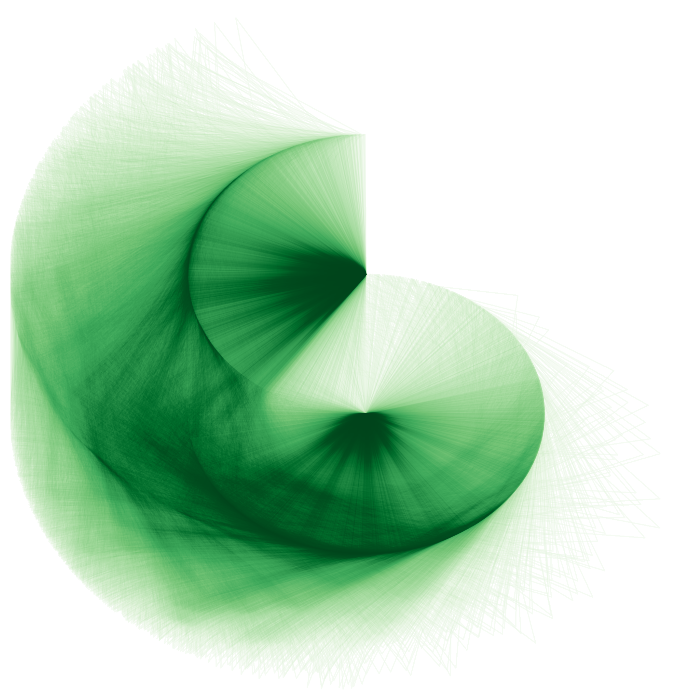}
    \caption{Reflected samples.}
    \label{fig:loop_reflected}
  \end{subfigure}
  \hfill
  \begin{subfigure}[t]{0.23\textwidth}
    \centering
    \includegraphics[width=\linewidth]{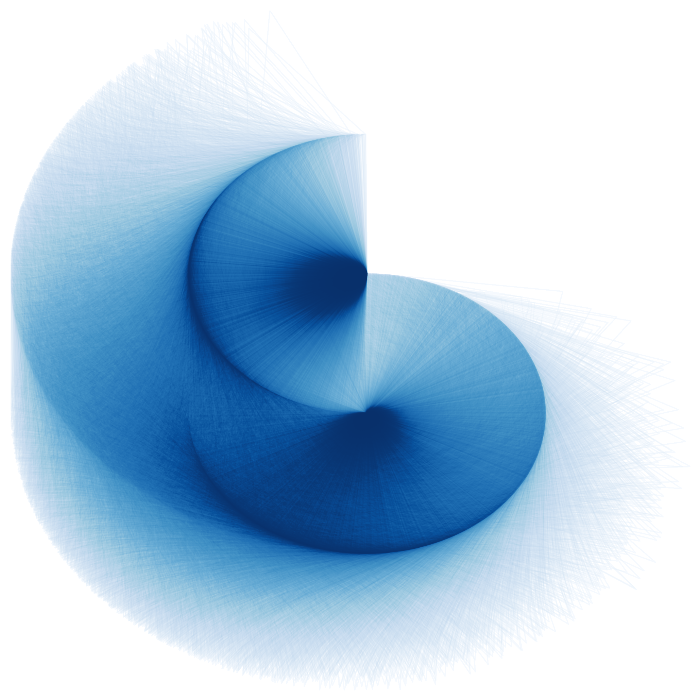}
    \caption{Uniform distribution.}
    \label{fig:loop_uniform}
  \end{subfigure}
  \caption{A qualitative comparison of $10^5$ samples from the data distribution,  our Metropolis diffusion model, a reflected diffusion model and the uniform distribution for the constrained conformational modelling of cyclic peptide backbones. For visual clarity, the figures only show the constrained planar projections encoded by \(\mathbb{P}\subset\Rbb^3\).}
  \label{fig:loop}
\end{figure}

\paragraph{Conformational modelling of protein backbones.} Modelling the conformational ensembles of proteins is a data-scarce problem with a range of important applications in biotechnology and drug discovery \citep{lane2023protein}. In many practical settings, it may often be unnecessary to model the structural ensembles of an entire protein, as researchers are primarily interested in specific functional sites that are embedded in a structurally conserved scaffold \citep{huang2016coming}. Modelling the conformational ensembles of such substructural elements requires positional constraints on their endpoints to ensure that they can be accommodated by the remaining protein. Using the parametrisation and dataset presented in \citet{fishman2023diffusion}, we formulate the problem of modelling the backbone conformations of a cyclic peptide of length $N=6$ as learning a distribution over the product of a polytope \(\mathbb{P}\subset\Rbb^{3}\) and the hypertorus \(\mathbb{T}^{4}\).

\begin{table}[H]
    \setlength{\tabcolsep}{8pt}
    \caption{
    Log-likelihood ($\uparrow$) of a held-out test set for the robotics and protein applications. Means and standard deviations are computed over 3 different runs. Average training time is provided in hours.
    }
    \label{tab:real_world_data}
    \centering
    \vspace{1em}

    \begin{adjustbox}{max width=\textwidth}
        
    \begin{tabular}{cccccc}
    \toprule
     \scshape \multirow{2}{*}{Dataset} & \scshape \multirow{2}{*}{Domain} & \multicolumn{2}{c}{\scshape Reflected} & \multicolumn{2}{c}{\scshape Metropolis} \\
     & &log-likelihood & runtime & log-likelihood & runtime \\
    \midrule
     Robotics & \(\c{S}_{++}^2\times\Rbb^2\) & $8.39 \pms{.06}$ & $9.52$ & $\mathbf{9.13 \pms{.03}}$ & $\mathbf{1.36}$\\
     Proteins & \(\mathbb{P}\subset\Rbb^3\times\mathbb{T}^4\) & $15.20\pms{.06}$ & 24.80 & $\mathbf{15.33 \pms{.02}}$ & $\mathbf{3.12}$ \\
    \bottomrule
    \end{tabular}
    \end{adjustbox}
\end{table}
We quantify the empirical performance of different methods by evaluating the log-likelihood of a held-out test set and present the resulting performance metrics in~\Cref{tab:real_world_data}.
Again, we find that our Metropolis model outperforms the reflected approach by a statistically significant margin while training 7-8 times as fast. 
Qualitative visual comparisons of samples from the true distribution, the trained diffusion models and the uniform distribution 
are presented in~\Cref{fig:robotic_arm,fig:loop}, with full univariate marginal and pairwise bivariate correlation plots postponed to~\Cref{sec:app_robotic_arms_spd,sec:app_loop_pairplots}.

\subsection{Modelling geospatial data within non-convex country borders}
\label{sec:exp_wildfires}

\begin{figure}[t]
\centering
\begin{subfigure}[t]{0.28\textwidth}
  \centering
  \includegraphics[width=\linewidth]{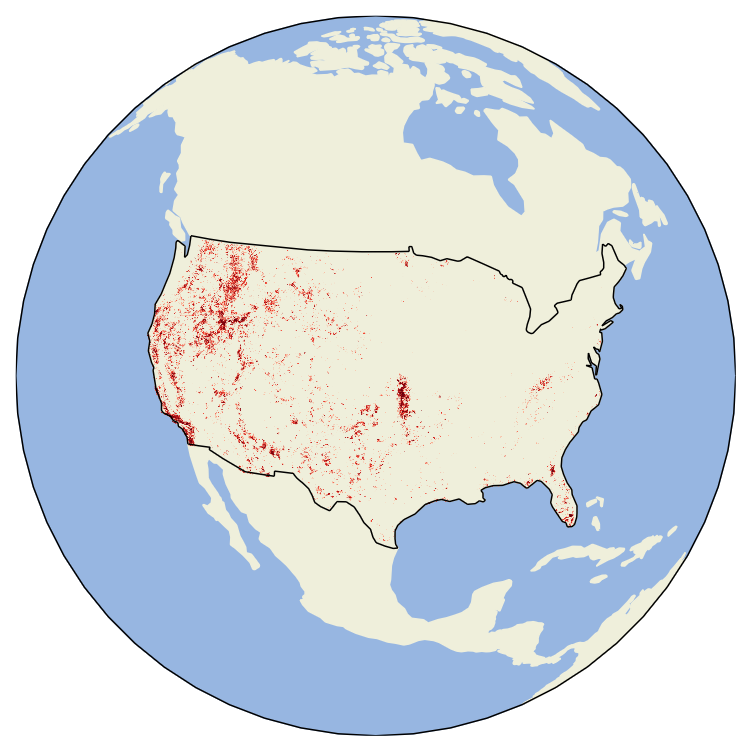}
  \caption{Data distribution.}
  \label{fig:wildfire_data_dist}
\end{subfigure}
\hfill
\begin{subfigure}[t]{0.28\textwidth}
  \centering
  \includegraphics[width=\linewidth]{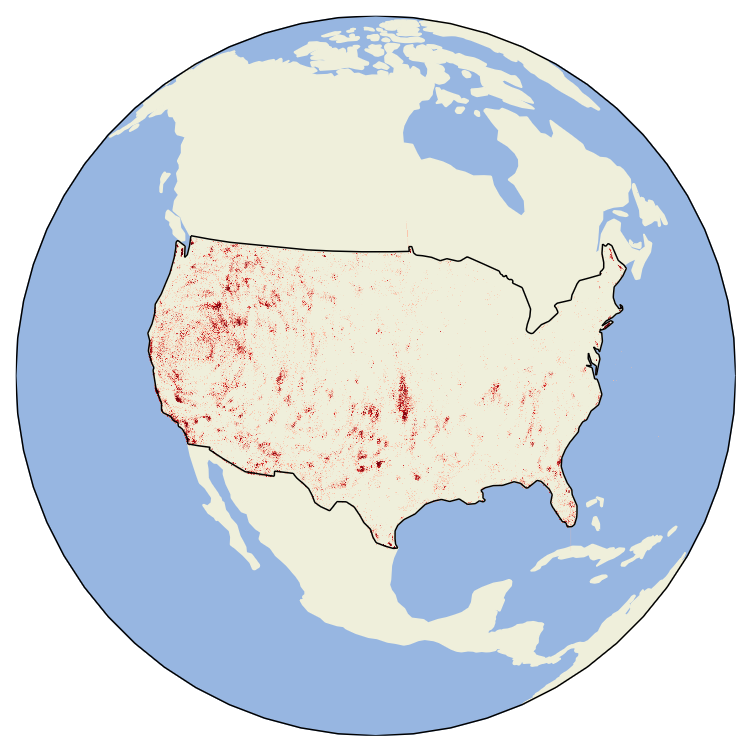}
  \caption{Metropolis samples.}
  \label{fig:wildfire_rejection_dist}
\end{subfigure}
\hfill
\begin{subfigure}[t]{0.28\textwidth}
  \centering
  \includegraphics[width=\linewidth]{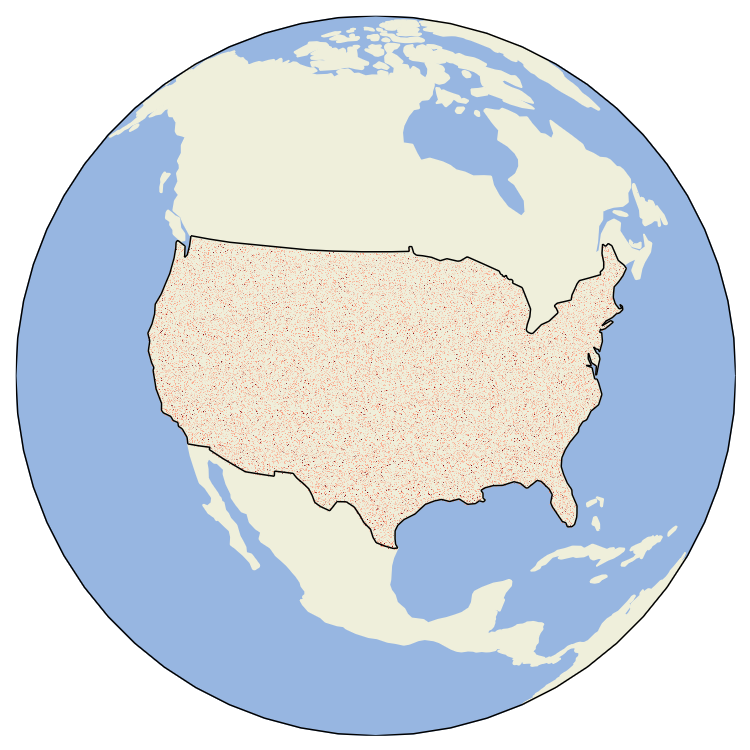}
  \caption{Uniform distribution.}
  \label{fig:wildfire_uniform_dist}
\end{subfigure}
\caption{Orthographic projection of $10^5$ samples from (a) the data distribution, (b) our Metropolis diffusion model, and (c) the uniform distribution, for geospatial data (wildfire incidence rates) within a non-convex boundary (the continental United States). The projections are aligned with the geometric centre of the boundary and zoomed in ten-fold for visual clarity.}
\label{fig:wildfire_headline_plot}
\end{figure}

Motivated by the strong empirical performance of our approach on tasks with challenging convex constraints, we investigated its ability to model
distributions whose support is restricted to manifolds with highly non-convex boundaries---a setting that is intractable with existing approaches. To this end, we derived a geospatial dataset based on wildfire incidence rates within the continental United States (see~\Cref{app_sec:wildfire_data} for full details) and trained a Metropolis diffusion model constrained by the corresponding country borders on the sphere $\mathcal{S}^2$. A qualitative visual comparison of samples from the true distribution, our model, and the uniform distribution is presented in~\Cref{fig:wildfire_data_dist,fig:wildfire_rejection_dist,fig:wildfire_uniform_dist} and a quantitative comparison to a Riemannian diffusion model on $\mathcal{S}^2$ \cite{debortoli2022riemannian} is given in~\Cref{tab:wildfire_riemann}. Both demonstrate that our approach is able to successfully capture challenging multimodal and sparse distributions on constrained manifolds with highly non-convex boundaries.

\begin{table}[h]
    \setlength{\tabcolsep}{8pt}
    \caption{MMD ($\downarrow$) of a held-out test set for the geospatial modelling dataset. Means and standard deviations are computed over 3 different runs. Average training time is provided in hours.
    }
    \vspace{5pt}
    \label{tab:wildfire_riemann}
    \centering
    \begin{tabular}{ccccc}
    \toprule
    \scshape Model & \scshape Domain & \scshape MMD & \scshape runtime & \scshape {\% in boundary} \\
    \midrule 
     Unconstrained & $\mathcal{S}^2$ & $0.1567\pms{0.013}$ & $\mathbf{0.81}$ & $63.3$\\
     Metropolis & $\mathbb{P}\subset\mathcal{S}^2$ & $\mathbf{0.1388\pms{0.015}}$  & $3.86$ & $\mathbf{100.0}$\\
    \bottomrule
    \end{tabular}
\end{table}
\section{Conclusion}

Accurately modelling distributions on constrained Riemannian manifolds is a challenging problem with a range of impactful practical applications. In this work, we have proposed a mathematically principled and computationally tractable extension of existing diffusion model methodology to this setting. Based on a \emph{Metropolisation} of random walks in Euclidean spaces and on Riemannian manifolds, we have shown that our approach corresponds to a valid discretisation of the reflected Brownian motion, justifying its use in diffusion models. To demonstrate the practical utility of our method, we have performed an extensive empirical evaluation, showing that it outperforms existing constrained diffusion models on a range of synthetic and real-world tasks defined on manifolds with convex boundaries, including applications from robotics and protein design. 
Leveraging the flexibility and simplicity of our method, we have also demonstrated that it extends beyond convex constraints and is able to successfully model distributions on manifolds with highly non-convex boundaries.
While we found our method to perform well across the synthetic and real-world applications we considered, we expect it to perform poorly on certain constraint geometries. For instance, the current implementation relies on an isotropic noise distribution which could impede its performance on exceedingly narrow constraint geometries, even with correspondingly small step sizes. In this context, an important direction of future research would be to investigate whether we can instead sample from more suitable distributions, e.g. a Dikin ellipsoid, while maintaining the simplicity and efficiency of the Metropolis approach. 


\newpage
\section*{Acknowledgements}

NF thanks the Rhodes Trust for supporting their studies at Oxford and this work. LK acknowledges support from the University of Oxford's Clarendon Fund.

\printbibliography

\newpage
\appendix
\appendixhead
\section{Overview}
\label{sec:app_overview}
In \Cref{sec:riemannain_intro}, we recall some basic concepts of Riemannian
geometry which are key to defining discretisations of the reflected Brownian motion.  In
\Cref{sec:reflected_discretisation}, we give some details on the reflection step
in reflected discretizations.  In \Cref{sec:rejection-convergence}, we prove the
convergence of the rejection and Metropolis discretizations to the true
reflected Brownian Motion.  The geospatial dataset with non-convex constraints based on wildfire incidence rates in the continental United States is presented
\Cref{app_sec:wildfire_data}. All supplementary experimental details and empirical results are
gathered in \Cref{sec:experimental_appendix}.

\section{Manifold concepts}
\label{sec:riemannain_intro}

In the following, we aim to introduce key concepts that underpin diffusion models on Riemannian manifolds, with a particular focus on notions relevant to the reflected Brownian motion that we build on in~\Cref{sec:reflected_discretisation}. For a more thorough treatment with reference to reflected diffusion models, we refer to \cite{fishman2023diffusion}. For a detailed presentation of smooth manifolds, see \citet{lee2013smooth}.

A Riemannian manifold is a tuple $(\M, \metric)$ with $\M$ a smooth manifold and $\metric$ a metric that imbues the manifold with a notion of distance and curvature and is defined as a smooth positive-definite inner product on each of the tangent spaces of the manifold:
\[\metric(p): \mathrm{T}_p\c{M} \times \mathrm{T}_p\c{M} \to \rset.\]
The tangent space $\mathrm{T}_p$ of a point \(p\) on a manifold is an extension of the notion of tangent planes and can be thought of as the space of derivatives of scalar functions on the manifold at that point. 

To establish how different tangent spaces relate to one another, we need to additionally introduce the concept of a \emph{connection}. This is a map that takes two vector fields and produces a derivative of the first with respect to the second, typically written as \(\nabla(X, Y) = \nabla_X Y\).
While there are infinitely many connections on any given manifold, the \emph{Levi-Cevita} emerges as a natural choice if we impose the following two conditions:
\begin{enumerate}[label=(\roman*)]
    \item \(X \cdot (\metric(Y,Z)) = \metric(\nabla_X Y, Z) + \metric(Y, \nabla_X Z)\),
    \item \(\sbr{X, Y} =  \nabla_X Y - \nabla_Y X\),
\end{enumerate}
where \(\sbr{\cdot, \cdot}\) is the Lie bracket. These conditions ensure that the connection is (i) metric-preserving and (ii) torsion-free, with the latter guaranteeing a unique connection and integrability on the manifold.

Using the metric and Levi-Cevita connection, we can define a number of key concepts:

\paragraph{Geodesic.} \emph{Geodesics} extend the Euclidean notion of `straight lines' to manifolds. They are defined as the unique path \(\gamma: (0,1) \to \c{M}\) such that $\nabla_{\gamma'}\gamma' = 0$ and are the shortest path between two points on a manifold, in the sense that 
\(\textstyle{L(\gamma) = \int_0^1 \sqrt{\metric(\gamma(t))(\gamma'(t), \gamma(t))} } \mathrm{d}t \)
is minimal. 

\paragraph{Exponential map.} The \emph{exponential map} on a manifold is given by  the mapping between an element \(\v{v}\in T_p\c{M}\) of the tangent space at point \(p\) and the endpoint of the unique geodesic \(\gamma\) with \(\gamma(0) = p\) and \(\gamma'(0) = \v{v}\).

\paragraph{Intersection.} The \emph{intersection} along a geodesic is the first point at which the geodesic intersects the boundary. We recall that the boundary is defined by $f=0$. We can define this via an optimisation problem: compute the minimum $t > 0$ such that we have that $\exp_\metric(x, t z)$ is a root of $f$: $f(\exp_\metric(x, t z)) = 0$. We will say that $\exp_\metric(x, t z) = \intersect_\metric (x, z; f) $ and that $t = \arg\intersect_\metric (x, z; f)$. 

\paragraph{Parallel transport.}
We say that a vector field \(X\) is \emph{parallel} to a curve \(\gamma: (0,1)\to\c{M}\) if
\(\nabla_{\gamma'} X = 0\),
where \(\gamma': (0,1)\to \mathrm{T}_{\gamma(t)}\c{M}\).
For two points on the manifold \(p,q\in\c{M}\) that are connected by a curve \(\gamma\), and an initial vector \(X_0 \in \mathrm{T}_p\c{M}\), there is a unique vector field \(X\) that is parallel to \(\gamma\) such that \(X(p) = X_0\). This induces a map between the tangent spaces at \(p\) and \(q\)
\(\tau_\gamma: \mathrm{T}_p\c{M} \to \mathrm{T}_q\c{M}\), which is referred to as the \emph{parallel transport} of tangent vectors between \(p\) and \(q\) and satisfies the condition that for \(\v{v}, \v{u} \in \mathrm{T}_p\c{M}\)
\(\metric(p)(\v{v}, \v{u}) = \metric(q)(\tau_\gamma(\v{v}), \tau_\gamma(\v{u})).\)

\paragraph{Reflection.}
For an element \(\v{v}\in T_p\c{M}\) in the tangent space of the manifold at point $p$ and a constraint characterised by its unit normal vector \(\v{n}\in T_p\c{M}\), the \emph{reflection} of \(\v{v}\) in the tangent space is given by $\v{v}' = \v{v} - 2\metric(\v{v},\v{n}) \v{n}$.

\section{Full Reflected Discretisation}
\label{sec:reflected_discretisation}

Here, we reproduce the central algorithm for the full discretisation of the reflected Brownian motion (\Cref{alg:app_reflected_random_walk}) derived for Euclidean models in \cite{lou2023reflected} and for Reimannian models in \cite{fishman2023diffusion}. Its central component is the \emph{Reflected Step Algorithm} (\Cref{alg:app_reflected_step_algorithm}), which gives a generic computation for the reflection in any manifold. 
Due to the need to balance speed and numerical instability issues around the boundary, an efficient practical implementation of the reflected step is highly non-trivial, even for simple polytopes in Euclidean space.
More complex geometries and boundaries make this problem significantly worse: a constraint on the trace of SPD matrices
under the log-Cholesky metric of \cite{lin2019riemannian} requires solving
complex non-convex optimisation problems for each sample at each discretised
sampling step in both the forward and reverse process. This motivates our work in this paper.

These problems motivated the development of our Metropolis approximation, which significantly simplifies the random walk. Instead of requiring the intersection, parallel transport and reflection, we simply need to be able to evaluate the constraint functions $f_i$. We highlight this simplicity in \Cref{alg:app_metropolis_random_walk}.

\begin{algorithm}[H]  
    \begin{algorithmic}
        \REQUIRE \(x \in \c{M}\), \(\v{v} \in \mathrm{T}_x\c{M}\), \(\{f_i\}_{i \in \mathcal{I}}\)
        \STATE \(\ell \gets \norm{\v{v}}_\metric\)
        \STATE \(\v{s} \gets \v{v} / \norm{\v{v}}_\metric\)
        \WHILE{\(\ell \geq 0\)}
            \STATE \(d_i =  \arg\intersect_\metric (x, z; f_i )\)
            \STATE \(i \gets \arg\min_i\; d_i \)
            \STATE \(\alpha \gets \min\del{d_i, \ell} \)
            \STATE \(x' \gets \exp_\metric\del{x, \alpha\v{s}}\)
            \STATE \(\v{s} \gets \pt_\metric\del{x, \v{s}, x'} \)
            \STATE \(\v{s} \gets \reflect\del{\v{s}, f_i}\)
            \STATE \(\ell \gets \ell - \alpha\)
            \STATE \(x \gets x'\)
        \ENDWHILE
        \STATE \textbf{return} \(x\)
    \end{algorithmic}
    \caption{\emph{Reflected Step Algorithm}. The algorithm operates by
      repeatedly taking geodesic steps until one of the constraints is violated, or
      until the step is fully taken. Upon hitting the boundary, we parallel-transport
      the tangent vector to the boundary and then reflect it against it. We
      then start a new geodesic from this point in the new direction. The \({\arg} \intersect_t\) function computes the distance one must travel along a geodesic in direction \(\v{s}\) until constraint \(f_i\) is violated. For a discussion of \(\pt\), \(\exp_\metric\) and \(\reflect\) see \cref{sec:riemannain_intro}.}
    \label{alg:app_reflected_step_algorithm}
\end{algorithm}

\begin{algorithm}[H]
  \caption{\emph{Reflected Random Walk}. Discretisation of the SDE
    $\rmd \bfX_t = b(t, \bfX_t) \rmd t + \rmd \bfB_t - \rmd \bfk_t$.}
   \label{alg:app_reflected_random_walk}
\begin{algorithmic}
    \STATE {\bfseries Require:} $T, N, X_0^\gamma, \{f_i\}_{i \in \mathcal{I}}$
  \STATE $\gamma = T / N$
  \FOR{$k \in \{0, \dots, N-1\}$}
  \STATE $Z_{k+1} \sim \mathrm{N}(0, \Id)$
  \STATE $X_{k+1}^\gamma = \text{ReflectedStep}[X_{k}^\gamma, \sqrt{\gamma} Z_{k+1}, \{f_i\}_{i \in \mathcal{I}}]$
  \ENDFOR
  \STATE {\bfseries return} $\{ X_k^\gamma\}_{k=0}^{N}$
\end{algorithmic}
\end{algorithm}

\begin{algorithm}
  \caption{\emph{Metropolis Random Walk}. Discretisation of the SDE
    $\rmd \bfX_t = b(t, \bfX_t) \rmd t + \rmd \bfB_t - \rmd \bfk_t$.}
   \label{alg:app_metropolis_random_walk}
\begin{algorithmic}
    \STATE {\bfseries Require:} $T, N, X_0^\gamma, \{f_i\}_{i \in \mathcal{I}}$
  \STATE $\gamma = T / N$
  \FOR{$k \in \{0, \dots, N-1\}$}
  \STATE $Z_{k+1} \sim \mathrm{N}(0, \Id)$
  \STATE \(X' \gets \exp_\metric\del{X_{k}^\gamma, \sqrt{\gamma} Z_{k+1}}\)
  \IF{$\max_{i\in\mathcal{I}} f_i(X') \leq 0$}
    \STATE $X_{k+1}^\gamma = X'$
  \ELSE
    \STATE $X_{k+1}^\gamma = X_{k}^\gamma$
  \ENDIF
  \ENDFOR
  \STATE {\bfseries return} $\{ X_k^\gamma\}_{k=0}^{N}$
\end{algorithmic}
\end{algorithm}
\newpage
\section{Convergence to the reflected process}\label{sec:rejection-convergence}
\def \Kker {\mathrm{K}}

In this note, we assume that $\M = \ensembleLigne{x \in \rset^d}{\Phi(x) > 0}$
is compact, with $\Phi \in \rmc^2(\rset^d, \rset)$. We have that
$\partial \M = \ensembleLigne{x \in \rset^d}{\Phi(x) = 0}$. In addition, we
assume that for any $x \in \partial \M$, $\normLigne{\nabla \Phi(x)} = 1$ and
that $\Phi$ is concave. The closure of $\M$ is denoted $\cl{\M}$. The assumption
that $\Phi$ is concave is only used in \Cref{thm:tubul-neighb}-\ref{item:d} and
can be dropped. We consider it for simplicity.

Let $(\hat{X}_k^\gamma)_{k \in \nset}$ given for any $\gamma >0$ and $k \in \nset$ by
$\hat{X}_0^\gamma = x \in \cl{\M}$ and for
$\hat{X}_{k+1}^\gamma = \hat{X}_k^\gamma + \sqrt{\gamma} Z_k^\gamma$ with $Z_k^\gamma$ a
Gaussian random variable conditioned on
$\hat{X}_k^\gamma + \sqrt{\gamma} Z_k^\gamma \in \cl{\M}$.  In practice, $Z_k^\gamma$
is sampled using rejection sampling. We define
$\hat{\bfX}^\gamma : \rset_+ \to \cl{\M}$ given for any $k \in \nset$ by
$\hat{\bfX}^\gamma_{k\gamma} = \hat{X}_k^\gamma$ and for any
$t \in \coint{k\gamma, (k+1)\gamma}$, $\hat{\bfX}^\gamma_t = \hat{X}_{k}^\gamma$. Note
that $(\bfX_t)_{t \in \ccint{0,T}}$ is a $\rmD(\ccint{0,T}, \cl{\M})$ valued
random variable, where $\rmD(\ccint{0,T}, \cl{\M})$ is the space of
right-continuous with left-limit processes which take values in $\cl{\M}$. We denote
$\hat{\Pbb}^\gamma$ the distribution of $(\hat{\bfX}_t^\gamma)_{t \in \ccint{0,T}}$ on
$\rmD(\ccint{0,T}, \cl{\M})$.

Our goal is to show the following theorem.

\begin{theorem}
  \label{thm:weak_convergence_appendix}
  For any $T \geq 0$, $(\hat{\bfX}_t^\gamma)_{t \in \ccint{0,T}}$ weakly converges to
  $(\bfX_t)_{t \in \ccint{0,T}}$ such that for any $t \in \ccint{0,T}$
  \begin{equation}
    \label{eq:skorkhod_pbm}
    \textstyle
    \bfX_t = x + \bfB_t - \bfk_t , \qquad \absLigne{\bfk}_t = \int_0^t \bm{1}_{\bfX_s \in \partial \M} \rmd \absLigne{\bfk}_s , \qquad \bfk_t = \int_0^t \bfn(\bfX_s) \rmd \absLigne{\bfk}_s . 
  \end{equation}
\end{theorem}

\begin{proof}
  In order to prove the result, we prove that the distribution of the Markov
  chain seen as an element of $\rmD(\ccint{0,T}, \cl{\M})$ converges to a
  solution of the Skorokhod problem \eqref{eq:skorkhod_pbm}. In particular, we
  first show that the limiting distribution satisfies a submartingale problem
  following \cite[Theorem 6.3]{stroock1971diffusion}. The transition from a
  solution of a submartingale problem to a weak solution of the Skorokhod
  problem is given by \cite[Theorem 1, Proposition 2.12]{kang2017submartingale}
  and \cite[Corollary 2.10]{ramanan2006reflected}. In order to apply
  \cite[Theorem 6.3]{stroock1971diffusion}, we define an intermediate drift and
  diffusion matrix, see \eqref{eq:intermediate_drift} and
  \eqref{eq:intermediate_diffusion}. To prove the theorem one needs to control
  the drift and diffusion matrix inside $\M$ and show that it converges to $0$
  and $\Id$ respectively. The technical part of the proof comes from the control
  of the drift coefficient near the boundary. In particular, we show that if the
  intermediate drift is large then we are close to the boundary and the
  intermediate drift is pointing inward. To investigate the local properties of
  the drift near the boundary we rely on the notion of tubular neighborhood, see
  \cite[Theorem 6.24]{lee2012smooth}.
\end{proof}

Some key properties of the tubular neighborhood are stated in
\Cref{sec:tubular_properties}. We then establish a few technical lemmas about the
tail probability of some distributions in \Cref{sec:technical_lemmas}. Controls
on the diffusion matrix and lower bounds on the probability of belonging in $\M$
are given in \Cref{sec:lower_control}. Properties of large drift terms are given
in \Cref{sec:large_drift_properties}. The convergence of the drift and diffusion
matrix on compact sets is given in \Cref{sec:convergence_on_compact}. The
convergence of the boundary terms is investigated in
\Cref{sec:convergence_boundary}. Finally, we conclude the proof in
\Cref{sec:submartingale}.

\subsection{Properties of the tubular neighborhood}
\label{sec:tubular_properties}

Using the results of \cite{lee2012smooth}, we establish the existence of an open
set of $\cl{\M}$ (for the induced topology of $\rset^d$) satisfying several
important properties. 


\begin{theorem}
  \label{thm:tubul-neighb}
  There exist $\msu \subset \cl{\M}$ open and $C\geq 1, \bar{r} >0$ such that
  for any $\gamma \in \ooint{0, \bar{\gamma}}$ with $\bgamma=1$ the following
  properties hold:
  \begin{enumerate}[label=(\alph*)]
  \item For any $x \in \msu$, there exist a unique $\bar{x} \in \partial \M$ and
    $\bar{\alpha} >0$ such that
    $x = \bar{x} + \bar{\alpha} \nabla \Phi(\bar{x})$. \label{item:b}
  \item For any $\bar{\alpha} \in \ccint{0,\bar{r}}$ and $\bar{x} \in \partial \M$
    such that $\bar{x} + \bar{\alpha} \nabla \Phi(\bar{x}) \in \cl{\M}$, let
    $x = \bar{x} + \bar{\alpha} \nabla \Phi(\bar{x})$ and $\msc(x, \gamma)$ such
    that $x + \sqrt{\gamma}z \in \msc(x,\gamma)$ if
    \begin{equation}
      -\bar{\alpha}\gamma^{-1/2} \leq \alpha < \bar{r} \gamma^{-1/2} , \qquad \normLigne{v}^2 \leq ( \alpha \gamma^{1/2} + \bar{\alpha})/(C\gamma), \label{eq:def_msc}
    \end{equation}
      with $z = \alpha \nabla \Phi(\bar{x}) + v$, with $v \perp \nabla \Phi(\bar{x})$. Then $\msc(x, \gamma) \subset \cl{\M}$. \label{item:c}
    \item
      $\msv = \ensembleLigne{\bar{x} + \alpha \nabla \Phi(\bar{x})}{\bar{x} \in
        \partial \M, \ \alpha \in \coint{0,\bar{r}}}$ is open in $\cl{\M}$. \label{item:c_prime}
    \item For any $x \in \msu$,
      $x + \sqrt{\gamma}z \in \cl{\M} \cap \msc(x,\gamma)^\complementary$ then
      $\alpha \geq \bar{r}\gamma^{-1/2}$ or
      $\normLigne{v}^2 \geq ( \alpha \gamma^{1/2} + \bar{\alpha})/(C\gamma)$ and
      $\alpha \gamma^{1/2} + \bar{\alpha} \geq 0$, with
      $z = \alpha \nabla \Phi(\bar{x}) + v$, with $\bar{x}$ given in
      \ref{item:b} and $v \perp \nabla \Phi(\bar{x})$. \label{item:d}.
  \item There exists $R > 0$ such that
    $\ensembleLigne{x \in \cl{\M}}{d(x, \partial \M) \leq 2R } \subset \msv$. \label{item:a}      
  \end{enumerate}
\end{theorem}

\begin{proof}
  Let $\gamma \in \ooint{0, \bgamma}$ with $\bgamma = 1$.  First, note that for
  any $\bar{x} \in \partial \M$, the normal space is given by
  $\ensembleLigne{\alpha \nabla \Phi(\bar{x})}{\alpha \in \rset}$.  Using this
  result and \cite[Theorem 6.24]{lee2012smooth} there exists $\tilde{r}_0 > 0$
  such that
  $\msu_0 = \ensembleLigne{\bar{x} + \alpha \nabla \Phi(\bar{x})}{\bar{x} \in
    \partial \M, \ \alpha \in \ooint{-\tilde{r}_0, \tilde{r}_0}} \subset
  \rset^d$ is open\footnote{This is the tubular neighborhood theorem which is
    key to the rest of the proof.}. We have that for any
  $\alpha \in \coint{-r_0, 0}$ and $\bar{x} \in \partial \M$
  \begin{align}
    \Phi(\bar{x}+\alpha \nabla \Phi(\bar{x})) &= \textstyle{\Phi(\bar{x}) + \alpha \normLigne{\nabla \Phi(\bar{x})}^2 + \int_0^1 \nabla^2 \Phi(\bar{x}+t\alpha \nabla \Phi(\bar{x}))(\alpha \nabla \Phi(\bar{x}))^{\otimes 2} \rmd t } \\
    &\leq \alpha + \tilde{C}_0 \alpha^2 < 0,
  \end{align}
  with $r_0 = \min(\tilde{r}_0, 1/(2\tilde{C}_0 ))$, 
  where we have used that $\Phi(\bar{x}) = 0$, $\normLigne{\nabla \Phi(\bar{x})}=1$ and 
  defined
  $\tilde{C}_0 = \sup \ensembleLigne{\normLigne{\nabla^2 \Phi(\bar{x}+\alpha \nabla \Phi(\bar{x}))}}{\bar{x}
    \in \partial \M, \ \alpha \in \ccint{-\tilde{r}_0,\tilde{r}_0}}$.
  Reciprocally, we have for any $\alpha \in \coint{0,r_0}$ and $\bar{x} \in \partial \M$
  \begin{align}
    \Phi(\bar{x}+\alpha \nabla \Phi(\bar{x})) &= \textstyle{\Phi(\bar{x}) + \alpha \normLigne{\nabla \Phi(\bar{x})}^2 + \int_0^1 \nabla^2 \Phi(\bar{x}+t\alpha \nabla \Phi(\bar{x}))(\alpha \nabla \Phi(\bar{x}))^{\otimes 2} \rmd t } \geq \alpha - C_0 \alpha^2, 
  \end{align}
  where we have used that $\Phi(\bar{x}) =0$,
  $\normLigne{\nabla \Phi(\bar{x})}=1$ and defined
  $C_0 = \sup \ensembleLigne{\normLigne{\nabla^2 \Phi(\bar{x}+\alpha \nabla
      \Phi(\bar{x}))}}{\bar{x} \in \partial \M, \ \alpha \in
    \ccint{-r_0,r_0}}$. Let $r_1 = \min(r_0, 1/(2C_0))$. Then,
  $\msu_1 = \ensembleLigne{\bar{x} + \alpha \nabla \Phi(\bar{x})}{\bar{x} \in
    \partial \M, \ \alpha \in \ooint{-r_1, r_1}} \subset \rset^d$ is open and
  \begin{equation}
    \msu_1 \cap \cl{\M} = \ensembleLigne{\bar{x} + \alpha \nabla \Phi(\bar{x})}{\bar{x} \in \partial \M, \
    \alpha \in \coint{0, r_1}} . 
\end{equation}
In what follows, we define $\msu = \msu_1 \cap \cl{\M}$. Note that $\msu$ is
open for the induced topology and that $\partial \M \subset \msu$. In
particular, $\partial \M$ is compact, $\msu^\complementary$ is closed and
$\partial \M \cap \msu^\complementary = \emptyset$. Hence, there exists $r > 0$
such that $d(\partial \M, \msu^\complementary) \geq 4r$. Without loss of
generality we can assume that $r \leq 1/2$.  We also have
$\ensembleLigne{x \in \cl{\M}}{d(x, \partial \M) \leq 2r } \subset \msu$. The
proof of \ref{item:b} follows from the definition of $\msu_0$. In the rest of
the proof, we define
\begin{equation}
  \label{eq:def_quantity_upper}
  C^{1/2} = 2\max(1, \sup \ensembleLigne{\normLigne{\nabla^2 \Phi(\bar{x} + u) }}{\bar{x} \in \partial \M, \ \normLigne{u}^2 \leq r(r+1)}) , \quad \bar{r} = \min(1/(2C^{1/2}), r/2).
\end{equation}
 Let us prove \ref{item:c}. Consider
$x + \sqrt{\gamma} z \in \msc(x,\gamma)$ with $\msc(x,\gamma)$ given by
\eqref{eq:def_msc} and $x = \bar{x} + \bar{\alpha} \nabla \Phi(\bar{x})$ and
$z = \alpha \nabla \Phi(\bar{x}) + v$ with $v \perp \nabla \Phi(\bar{x})$.
In particular, we recall that we have 
    \begin{equation}
      -\bar{\alpha}\gamma^{-1/2} \leq \alpha < \bar{r} \gamma^{-1/2} , \qquad \normLigne{v}^2 \leq ( \alpha \gamma^{1/2} + \bar{\alpha})/(C\gamma) . 
    \end{equation}
    This implies that
    \begin{equation}
      \label{eq:consequence_upper}
      \bar{\alpha} + \sqrt{\gamma} \alpha \leq 2 \bar{r} , \qquad \gamma \normLigne{v}^2 \leq 2 \bar{r} / C . 
    \end{equation}
    First, using that $C \geq1$, $\normLigne{\nabla \Phi(\bar{x})}=1$,
    \eqref{eq:consequence_upper} and \eqref{eq:def_quantity_upper}, we have
\begin{equation}
  \normLigne{x+\sqrt{\gamma}z - \bar{x}}^2 = (\bar{\alpha} + \sqrt{\gamma}\alpha)^2 + \gamma \normLigne{v}^2 \leq r^2 + r/C \leq r(r+1) . 
\end{equation}
Then, we have that
\begin{align}
  \Phi(x+\sqrt{\gamma}z) &= \textstyle{\Phi(\bar{x}) + \bar{\alpha} + \sqrt{\gamma} \alpha  + \int_0^1 \nabla^2 \Phi(\bar{x}+t(x+\sqrt{\gamma}z-\bar{x}))(x+\sqrt{\gamma}z-\bar{x})^{\otimes 2} \rmd t } \\
  & \geq \bar{\alpha} + \sqrt{\gamma} \alpha - (C^{1/2}/2)  ((\bar{\alpha} + \sqrt{\gamma} \alpha)^2 + \gamma \normLigne{v}^2) ,
\end{align}
where we recall that
\begin{equation}
  C^{1/2} = 2\max(1, \sup \ensembleLigne{\normLigne{\nabla^2 \Phi(\bar{x} + u) }}{\bar{x} \in \partial \M, \ \normLigne{u}^2 \leq r(r+1)}) , \qquad \bar{r} = \min(1/(2C^{1/2}), r/2).
\end{equation}
First, using that $r\leq 1/2$ and \eqref{eq:consequence_upper}, we have 
$\bar{\alpha} + \sqrt{\gamma}\alpha \leq 2r \leq 1$. Since, 
$\normLigne{v}^2 \leq (\bar{\alpha} +\sqrt{\gamma} \alpha)/(C\gamma)$ and  we have
that $\normLigne{v}^2 < 1/(C \gamma)$. Let
$P(X) = X - (C^{1/2}/2)X^2 - (C^{1/2}/2)\gamma\normLigne{v}^2$. We have that $P(x) \geq 0$ if and
only if $x \in \ccint{x_{\min}, x_{\max}}$ with
\begin{equation}
  x_{\min} = (1 -(1 -C\gamma \normLigne{v}^2)^{1/2})/C^{1/2}, \qquad x_{\max} = (1 +(1 -C\gamma \normLigne{v}^2)^{1/2})/C^{1/2}.
\end{equation}
Using that for any $t \in \ooint{0,1}$, $(1-t)^{1/2} \geq 1 - t$ we have that
\begin{equation}
  x_{\min} \leq \gamma C \normLigne{v}^2/2 , \qquad x_{\max} \geq 1/C^{1/2} . 
\end{equation}
Since $\normLigne{v}^2 \leq (\sqrt{\gamma} \alpha + \bar{\alpha})/(\gamma C)$,
we have that $\bar{\alpha} + \sqrt{\gamma} \alpha \geq x_{\min}$. In addition,
using that $\bar{\alpha} + \sqrt{\gamma} \alpha \leq 2 \bar{r} \leq 1/C^{1/2} \leq x_{\mathrm{max}}$,
we get that $P(\bar{\alpha} + \sqrt{\gamma}\alpha) \geq 0$ and therefore
$x + \sqrt{\gamma}z \in \cl{\M}$ since $\Phi(x+ \sqrt{\gamma}z) \geq 0$. This
concludes the proof of \ref{item:c}. Note that the condition
$\alpha \geq -\gamma^{-1/2} \bar{\alpha}$ is implied by the condition
$\normLigne{v}^2 \leq (\sqrt{\gamma} \alpha + \bar{\alpha})/(\gamma C)$.
Using that
$\msv \subset \ensembleLigne{x \in \cl{\M}}{d(x,\partial \M) \leq 2r} \subset
\msu$, \ref{item:c_prime} is a direct consequence of \cite[Theorem
6.24]{lee2012smooth}].
Next,
we prove \ref{item:d}.    Let
$x + \sqrt{\gamma}z \in \cl{\M} \cap \msc(x,\gamma)^\complementary$. If
$\alpha <-\bar{\alpha}\gamma^{-1/2}$ then since $\Phi$ is concave, we have 
  \begin{align}
    \Phi(x+\sqrt{\gamma}z) = \textstyle{\Phi(\bar{x}) + \bar{\alpha} + \sqrt{\gamma} \alpha + \int_0^1 \nabla^2 \Phi(\bar{x} + t(x+\sqrt{\gamma}z-\bar{x}))(x+\sqrt{\gamma}z-\bar{x})^{\otimes 2} \rmd t } < 0 ,
  \end{align}
  where we have used that $\Phi(\bar{x}) = 0$. This is absurd, hence either
  $\alpha \geq \bar{r}\gamma^{-1/2}$ or
  $\normLigne{v}^2 \geq ( \alpha \gamma^{1/2} + \bar{\alpha})/(C\gamma)$ and
  $\alpha \gamma^{1/2} + \bar{\alpha} \geq 0$, which concludes the proof. The
  proof of \ref{item:a} is similar to the proof that
  $\ensembleLigne{x \in \cl{\M}}{d(x, \partial \M) \leq 2r } \subset \msu$.
\end{proof}

The main message of \Cref{thm:tubul-neighb} is that using
\Cref{thm:tubul-neighb}-\ref{item:d}, if we move in the direction of
$\nabla \Phi(\bar{x})$ (the inward normal) with magnitude $\alpha$ then we are
allowed to move in the orthonormal direction with magnitude $\alpha^{1/2}$. In
the next paragraph, we discuss this fact in details and shows it is necessary
for the rest of our study.


\paragraph{The necessity of \Cref{thm:tubul-neighb}-\ref{item:c}.}

\begin{figure}
        \includegraphics[width=.45\linewidth]{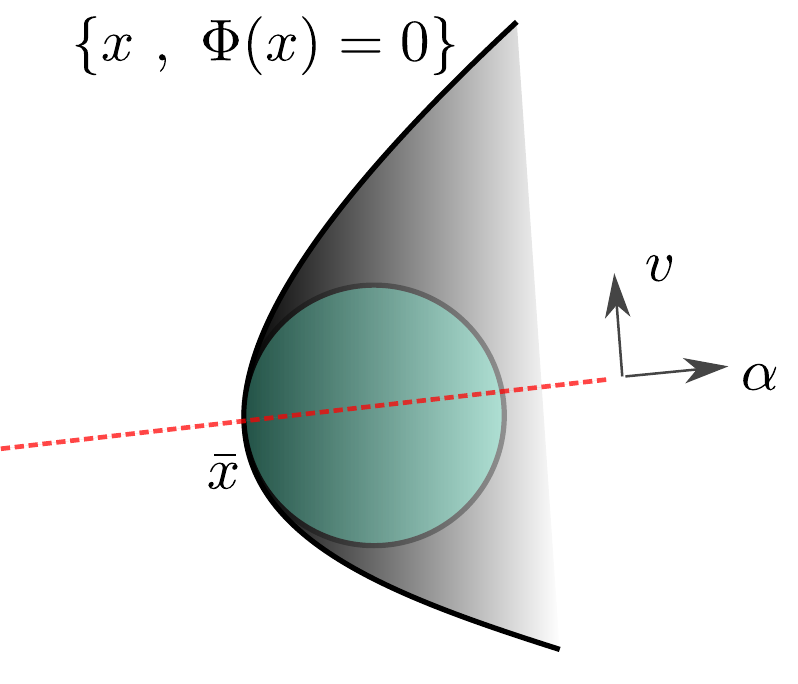} \hfill
        \includegraphics[width=.5\linewidth]{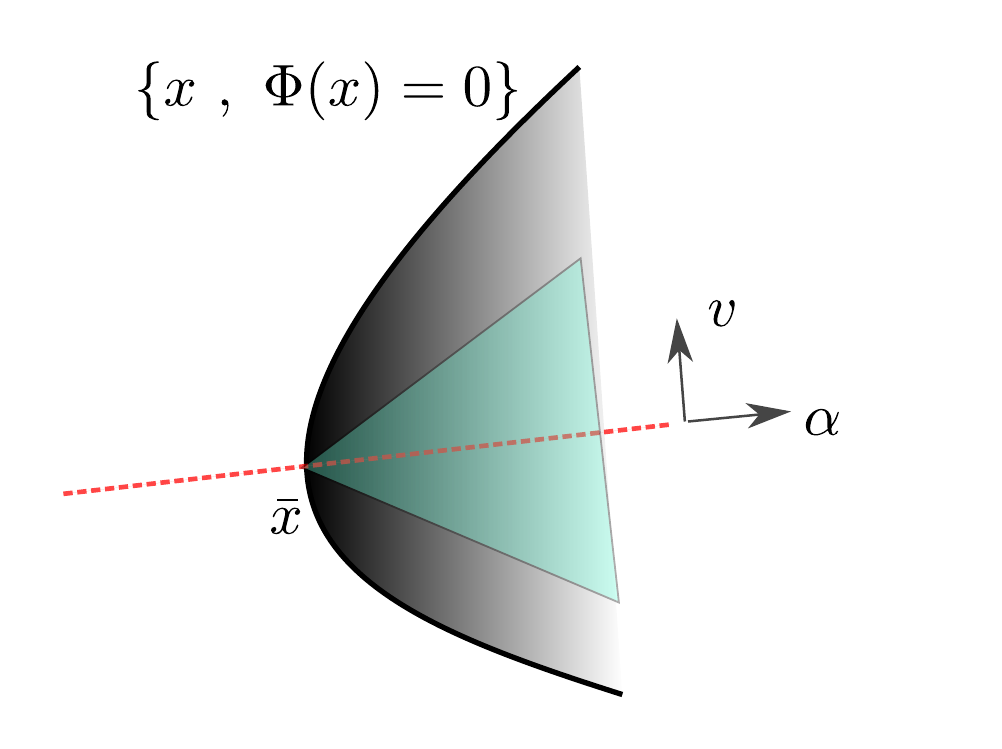}
        \caption{The grey shaded area represents $\cl{\M}$ while the blue shaded
          area represents $\msc(x, \gamma)$ for an arbitrary value of $\gamma$
          and $x =\bar{x} \in \partial \M$.}
        \label{fig:comparison_area}
      \end{figure}

At first sight one can wonder if the statement of
\Cref{thm:tubul-neighb}-\ref{item:c} could be simplify. In particular, it would
be simpler to replace this statement with: for any
$\bar{\alpha} \in \ccint{0,\bar{r}}$ and $\bar{x} \in \partial \M$ such that
$\bar{x} + \bar{\alpha} \nabla \Phi(\bar{x}) \in \cl{\M}$, let
$x = \bar{x} + \bar{\alpha} \nabla \Phi(\bar{x})$ and $\msc(x, \gamma)$ such
that $x + \sqrt{\gamma}z \in \msc(x,\gamma)$ if
    \begin{equation}
      -\bar{\alpha}\gamma^{-1/2} \leq \alpha < \bar{r} \gamma^{-1/2} , \qquad \normLigne{v}^2 \leq ( \alpha \gamma^{1/2} + \bar{\alpha})^2/(C\gamma), \label{eq:def_msc_illus}
    \end{equation}
    with $z = \alpha \nabla \Phi(\bar{x}) + v$, with
    $v \perp \nabla \Phi(\bar{x})$. Then $\msc(x, \gamma) \subset \cl{\M}$. Note
    that $\normLigne{v}^2 \leq ( \alpha \gamma^{1/2} + \bar{\alpha})/(C\gamma)$
    is replaced by
    $\normLigne{v}^2 \leq ( \alpha \gamma^{1/2} + \bar{\alpha})^2/(C\gamma)$,
    see \Cref{fig:comparison_area} for an illustration. However, in that case
    \Cref{thm:tubul-neighb}-\ref{item:d} becomes: in addition, if
    $x + \sqrt{\gamma}z \in \cl{\M} \cap \msc(x,\gamma)^\complementary$ then
    $\alpha \geq r\gamma^{-1/2}$ or
    $\normLigne{v}^2 \geq ( \alpha \gamma^{1/2} + \bar{\alpha})^2/(C\gamma)$ and
    $\alpha \gamma^{1/2} + \bar{\alpha} \geq 0$.

    In what follows, when controlling the properties of large drift, see the
    proof of \Cref{prop:lower-bound-drift} and the proof of
    \Cref{prop:condition_duo_iv}, we need to control quantities of the form
    $\probaLigne{x + \sqrt{\gamma} Z \in \msc(x, \gamma)^\complementary \cap
      \cl{\M}} /\sqrt{\gamma}$\footnote{The division by $\sqrt{\gamma}$ comes
      from the definition of the intermediate drift
      \eqref{eq:intermediate_drift}.}  Using the original
    \Cref{thm:tubul-neighb}-\ref{item:d} it is possible to show that this
    quantity is bounded.  However, if one uses the updated version of
    \Cref{thm:tubul-neighb}-\ref{item:d} then one needs to show that there
    exists $M \geq 0$ and $\bgamma >0$ such that for any
    $\gamma \in \ooint{0, \bgamma}$ (here we have assumed that
    $\bar{\alpha} = 0$, i.e.~ $x \in \partial \M$ for simplicity)
    \begin{equation}
      \textstyle \int_{0}^{r/\gamma^{-1/2}} \int_{\nabla \Phi(\bar{x})^\perp} \bm{1}_{\normLigne{v}^2 \geq \alpha^2} \varphi(v) \varphi(\alpha) \rmd v \rmd \alpha \leq M \sqrt{\gamma} , 
      \end{equation}
      which is absurd.

\subsection{Technical lemmas}
\label{sec:technical_lemmas}

We start with a few technical lemmas which will allow us to control some
Gaussian probabilities outside of a compact set. We denote
$\Psi: \rset_+ \times \nset \to \ccint{0,1}$ such that for any $k \in \nset$,
$\Psi(\cdot, k)$ is the tail probability of a $\chi$-squared random variable
with parameter $k$, i.e.~ for any $k \in \nset$ and $t \geq 0$ we have
\begin{equation}
  \label{eq:tail_probability}
  \Psi(t, k) = \probaLigne{\normLigne{Z}^2 \geq t},
\end{equation}
with $Z$ a Gaussian random variable in $\rset^k$ with zero mean and identity
covariance matrix. We will make extensive use of the following lemma which is a
direct consequence of \cite[Section 4, Lemma 1]{laurent2000adaptive}.

\begin{lemma}
  \label{lemma:tail_chi_square}
  For any $k \in \nset$ and $t \in \rset_+$ with $t \geq 5k$, $\Psi(t, k) \leq \exp[-t/5]$.
\end{lemma}

\begin{proof}
  Let $k \in \nset$.  First, note that for any $x \geq k$, we have that
  $k + 2 (k x)^{1/2} + 2x \leq 5x$.  Combining this result and \cite[Section 4,
  Lemma 1, Equation (4.3)]{laurent2000adaptive}, we have that for any $x \geq k$
  \begin{equation}
    \probaLigne{\normLigne{X}^2 \geq 5 x} \leq \exp[-x] ,
  \end{equation}
  with $X$ a $\rset^k$-valued Gaussian random variable with zero mean and
  identity covariance matrix. This concludes the proof upon letting $t = 5x$.
  \end{proof}

  Let $\varphi: \ \rset^p \to \rset_+$ given for any $u \in \rset$ by
  $\varphi(u) = (2 \uppi)^{-p/2} \exp[-\normLigne{u}^2/2]$\footnote{In the rest
    of the supplementary, we never precise the dimension $p \in \nset$ which can
    be deduced from the variable.}, i.e.~the density of a real Gaussian random
  variable with zero mean and unit variance. While
  \Cref{lemma:tail_chi_square_int} appears technical, it will be central to
  provide quantitative upper bounds on the \emph{rejection} probability, see
  \Cref{lemma:lower_bound_prob} for instance.

\begin{lemma}
  \label{lemma:tail_chi_square_int}
  For any $k \in \nset$, $\alpha_0>0$, $\beta_0 \in \ocint{0,1}$ and $\delta > 0$ we have 
  \begin{equation}
   \psi(\delta) =  \textstyle{\sup \ensembleLigne{\int_0^{+\infty} \Psi(\alpha_0 t/\delta, k)^{\beta_0} \varphi(t - t_0/\delta) \rmd t}{t_0 \geq 0}} \leq C_0 \delta , 
  \end{equation}
  with $C_0 = 5(2\uppi)^{-1/2}(k+1)/(\alpha_0 \beta_0)$.
\end{lemma}

\begin{proof}
  Let $k \in \nset$, $\alpha_0 >0$, $\beta_0 \in \ocint{0,1}$ and $\delta > 0$. Let
  $t_{\delta} = 5 k \delta /\alpha_0$. Note that if $t \geq t_\delta$ then,
  $\alpha_0 t /\delta \geq 5k$. In addition, we have 
  \begin{align}
    \textstyle
    \int_0^{+\infty} \Psi(\alpha_0 t/\delta, k)^{\beta_0} \varphi(t - t_0/\delta) \rmd t & \textstyle \leq (2 \uppi)^{-1/2} \int_0^{+\infty} \Psi(\alpha_0 t/\delta, k)^{\beta_0} \rmd t \\
    & \textstyle \leq (2 \uppi)^{-1/2} \int_0^{t_\delta} \Psi(\alpha_0 t/\delta, k)^{\beta_0} \rmd t + (2 \uppi)^{-1/2} \int_{t_\delta}^{+\infty} \Psi(\alpha_0 t/\delta, k)^{\beta_0} \rmd t .
  \end{align}
  Using that for any $w >0$,
  $\int_0^{+\infty} \exp[-w t] \rmd t \leq (1/w)$, that for any
  $u \geq 0$, $\Psi(u, k) \leq 1$ and that if $u \geq 5k$,
  $\Psi(u,k) \leq \exp[-u/5]$, we get for any $t_0 \geq 0$
  \begin{equation}
    \textstyle
\int_0^{+\infty} \Psi(\alpha_0 t/\delta, k) \varphi(t - t_0/\delta) \leq (2\uppi)^{-1/2} [5k \delta/\alpha_0 + 5 \delta/(\alpha_0 \beta_0)] \leq (5(2\uppi)^{-1/2}(k+1)/(\alpha_0\beta_0)) \delta ,
\end{equation}
which concludes the proof.
\end{proof}

Finally, we have the following lemma, which is similar to
\Cref{lemma:tail_chi_square} but will be used to control quantities related to
the norm.

\begin{lemma}
  \label{lemma:tail_chi_square_int_duo}
  For any $k \in \nset$, $\alpha_0>0$, $\beta_0 \in \ocint{0,1}$ and $\delta > 0$ we have 
  \begin{equation}
   \psi(\delta) =  \textstyle{ \int_0^{+\infty} \Psi(\alpha_0 t/\delta, k)^{\beta_0} t \varphi(t) \rmd t} \leq C_0 \delta^2 , 
  \end{equation}
  with $C_0 = 25(2\uppi)^{-1}(k^2+1)/(\alpha_0\beta_0)^2$.
\end{lemma}

\begin{proof}
  Let $k \in \nset$, $\alpha_0 >0$, $\beta_0 \in \ocint{0,1}$ and $\delta > 0$.
  Let $t_{\delta} = 5 k \delta /\alpha_0$. Note that if $t \geq t_\delta$ then,
  $\alpha_0 t /\delta \geq 5k$. In addition, we have
  \begin{align}
     \textstyle{ \int_0^{+\infty} \Psi(\alpha_0 t/\delta, k)^{\beta_0} t \varphi(t) \rmd t} &\leq  (2 \uppi)^{-1} \textstyle{ \int_0^{t_\delta} \Psi(\alpha_0 t/\delta, k)^{\beta_0} t \rmd t} + (2\uppi)^{-1} \textstyle{ \int_{t_\delta}^{+\infty} \Psi(\alpha_0 t/\delta, k)^{\beta_0} t  \rmd t} .
  \end{align}
  In addition, using that if $u \geq 5k$ then $\Psi(u,k) \leq \exp[-u/5]$, we get
  \begin{align}
    \textstyle (2\uppi)^{-1}  \int_{t_\delta}^{+\infty} \Psi(\alpha_0 t/\delta, k)^{\beta_0} t  \rmd t &\leq (2\uppi)^{-1} \textstyle \int_0^{+\infty} \exp[-\alpha_0\beta_0 t /(5\delta)] t \rmd t \leq (2\uppi)^{-1} 25 \delta^2 / (\alpha_0 \beta_0)^2 . 
  \end{align}
  Finally, using that for any $u \geq 0$, $\Psi(u,k)\leq1$, we have 
  \begin{equation}
    \textstyle (2 \uppi)^{-1} \textstyle{ \int_0^{t_\delta} \Psi(\alpha_0 t/\delta, k)^{\beta_0} t \rmd t} \leq (2 \uppi)^{-1} 25 k^2 \delta^2 / \alpha_0^2 ,
  \end{equation}
  which concludes the proof.
\end{proof}

\subsection{Lower bound on the inside probability and control of moments of order two and higher}
\label{sec:lower_control}

\paragraph{Lower bound on the inside probability.} We begin with the following
lemma which controls the expectation of $1 + \normLigne{Z}$ \emph{outside} of
$\msc(x,\gamma)$. We recall that $\msv$ is defined in
\Cref{thm:tubul-neighb}-\ref{item:c_prime}. 

\begin{lemma}
  \label{lemma:control_outside}
  Let $\bgamma = 1$.  Let $x \in \msv$, $Z \in \sim \mathrm{N}(0, \Id)$ and
  $\gamma \in \ooint{0, \bar{\gamma}}$ then we have
  \begin{equation}
    \max(\expeLigne{\bm{1}_{x + \sqrt{\gamma} Z \in \cl{\M} \cap \msc(x,\gamma)^\complementary}}, \expeLigne{\langle Z, \nabla \Phi(\bar{x})\rangle \bm{1}_{x + \sqrt{\gamma} Z \in \cl{\M} \cap \msc(x,\gamma)^\complementary}}) \leq \psi(\gamma) ,
  \end{equation}
  with $\psi: \ \rset_+ \to \rset_+$ such that 
  $\limsup_{t \to 0} \psi(t)/t^{1/2} < +\infty$.
\end{lemma}

\begin{proof}
  Let $\bar{r} > 0$ given by \Cref{thm:tubul-neighb}. First, we have that
  \begin{align}
    &\textstyle\int_{\rset} \int_{\rset^{d-1}} (1 + \abs{\alpha} + \normLigne{v}) \bm{1}_{\alpha \geq \bar{r} /\gamma^{1/2}} \varphi(\alpha) \varphi(v) \rmd \alpha \rmd v \\
    &\textstyle\qquad \leq d \int_{\rset} (1 + \abs{\alpha}) \bm{1}_{\alpha \geq \bar{r} /\gamma^{1/2}} \varphi(\alpha)  \rmd \alpha \leq d (\Psi(\bar{r}^2/\gamma,1) + \exp[-\bar{r}^2/(2\gamma)]) . \label{eq:true_uno}
  \end{align}
  Second, using \Cref{lemma:tail_chi_square_int}, we have that
  \begin{align}
    &\textstyle\int_{\rset} \int_{\rset^{d-1}}\bm{1}_{\normLigne{v}^2\geq (\bar{\alpha}+\sqrt{\gamma}\alpha)/(C\gamma)} \bm{1}_{\bar{\alpha}+\sqrt{\gamma}\alpha \geq 0} \varphi(\alpha) \varphi(v) \rmd \alpha \rmd v \\
    &\qquad \textstyle \leq \int_{\rset} \bm{1}_{\bar{\alpha}+\sqrt{\gamma}\alpha \geq 0} \Psi((\bar{\alpha}+\sqrt{\gamma}\alpha)/(C\gamma), d-1) \varphi(\alpha) \rmd \alpha \\
    &\qquad \textstyle \leq \int_{0}^{+\infty} \Psi(\alpha/C\gamma^{1/2}, d-1) \varphi(\alpha-\bar{\alpha}/\gamma^{1/2}) \rmd \alpha \leq \Psi_1(\gamma^{1/2}) . \label{eq:true_duo}
  \end{align}
  Second, using \Cref{lemma:tail_chi_square_int_duo}, we have that
   \begin{align}
     &\textstyle\int_{\rset} \int_{\rset^{d-1}}\alpha \bm{1}_{\normLigne{v}^2\geq (\bar{\alpha}+\sqrt{\gamma}\alpha)/(C\gamma)} \bm{1}_{\bar{\alpha}+\sqrt{\gamma}\alpha \geq 0} \varphi(\alpha) \varphi(v) \rmd \alpha \rmd v \\
     &\qquad = \textstyle\int_{\rset} \alpha \Psi((\bar{\alpha}+\sqrt{\gamma}\alpha)/(C\gamma),d-1) \bm{1}_{\bar{\alpha}+\sqrt{\gamma}\alpha \geq 0} \varphi(\alpha)  \rmd \alpha \\
     &\qquad \textstyle \leq \int_0^{+\infty} \Psi(\alpha/C\gamma^{1/2},d-1) \alpha \varphi(\alpha) \rmd \alpha \leq \Psi_2(\gamma^{1/2}). \label{eq:true_trio}
   \end{align}
   Note that we have
   $\limsup_{\gamma \to 0} \Psi_2(\gamma^{1/2}) + \Psi_1(\gamma^{1/2}) <
   +\infty$.  We conclude upon combining \eqref{eq:true_uno},
   \eqref{eq:true_duo} and \eqref{eq:true_trio} with
   \Cref{thm:tubul-neighb}-\ref{item:d} and the fact that $\normLigne{\Phi(\bar{x})} =1$.
\end{proof}

The following lemma allow us to give a lower bound to the quantity
$\expeLigne{\bm{1}_{x+\sqrt{\gamma}Z \in \cl{\M}}}$ uniformly w.r.t $x \in \cl{\M}$.

\begin{lemma}
  \label{lemma:lower_bound_prob}
  There exists $\bgamma >0$ such that for any $\gamma \in \ooint{0, \bgamma}$
  and for any $x \in \cl{\M}$, $\gamma \in \ooint{0, \bar{\gamma}}$ and
  $Z \sim \mathrm{N}(0, \Id)$ we have
  \begin{equation}
    \expeLigne{\bm{1}_{x + \sqrt{\gamma} Z \in \cl{\M}}} \geq 1/4 \eqsp . 
  \end{equation}
\end{lemma}

\begin{proof}
  Let $\gamma \in \ooint{0, \bgamma}$. 
  If $x \not \in \msv$ then $\mathrm{B}(x,2R) \subset \M$ using
  \Cref{thm:tubul-neighb}-\ref{item:a} and therefore
  $\expeLigne{\bm{1}_{x + \sqrt{\gamma} Z \in \cl{\M}}} \geq 1/4$ for
  $\bgamma > 0$ small enough. Now, assume that $x \in \msv$. Using
  \Cref{lemma:control_outside}, we have that
  $\expeLigne{\bm{1}_{x + \sqrt{\gamma} Z \in \cl{\M}\cap \msc(x,
      \gamma)^\complementary}} \leq \psi(\gamma)$.  In addition, using
  \Cref{thm:tubul-neighb}-\ref{item:c}, we have that for any $\gamma >0$
  \begin{align}
    \expeLigne{\bm{1}_{x + \sqrt{\gamma} Z \in \cl{\M}}} &\geq \expeLigne{\bm{1}_{x + \sqrt{\gamma} Z \in \msc(x, \gamma)}} \\
    &\geq \textstyle{\int_{-\bar{\alpha} \gamma^{-1/2}}^{r \gamma^{-1/2}} \int_{\nabla \Phi(\bar{x})^\perp} \bm{1}_{\normLigne{v}^2\leq (\bar{\alpha}+\gamma^{1/2}\alpha)/(C\gamma)} \varphi(\alpha)\varphi(v) \rmd \alpha \rmd v} \\
                                                             &\geq \textstyle{\int_{-\bar{\alpha} \gamma^{-1/2}}^{r \gamma^{-1/2}} (1 - \Psi((\bar{\alpha}+\gamma^{1/2}\alpha)/(C\gamma),d-1)) \varphi(\alpha) \rmd \alpha} \\
    & \textstyle{\geq (1/2) - \Psi(r^2/\gamma, 1) - \int_{-\bar{\alpha} \gamma^{-1/2}}^{+\infty} \Psi((\bar{\alpha}+\gamma^{1/2}\alpha)/(C\gamma),d-1) \varphi(\alpha) \rmd \alpha. }
  \end{align}
  Hence, using \Cref{lemma:tail_chi_square} and
  \Cref{lemma:tail_chi_square_int}, there exists $\bgamma >0$ such that for any
  $\gamma \in \ooint{0, \bgamma}$, 
  $\Psi(r^2/\gamma, 1) + \int_0^{+\infty} \Psi(\alpha/(C\gamma^{1/2}),d)
  \varphi(\alpha-\gamma^{1/2}\bar{\alpha}) \rmd \alpha \leq 1/4$, which
  concludes the proof.
\end{proof}

Note that the result of \Cref{lemma:lower_bound_prob} can be improved to
$1/2-\vareps$ for any $\vareps > 0$. In particular this result tells us that for
$\gamma > 0$ small enough, $\cl{\M}$ looks like the \emph{hyperplane} from the
point of view of the Gaussian with variance $\gamma$ centered on $\partial \M$.


\paragraph{Bound on moments of order two and higher.}

In what follows, we define for any $\gamma >0$,
$\Delta^\gamma: \ \cl{\M} \to \rset_+$ given for any $x \in \cl{\M}$ by
\begin{equation}
  \label{eq:intermediate_Delta}
 \textstyle{ \Delta^\gamma(x) = (1/\gamma) \int_{\rset^d} \bm{1}_{x + \sqrt{\gamma} z \in \M} \normLigne{\sqrt{\gamma} z}^4 \varphi(z) \rmd z  / \int_{\rset^d} \bm{1}_{x + \sqrt{\gamma} z \in \M} \varphi(z) \rmd z. }
\end{equation}

\begin{proposition}
  \label{prop:condition_a}
  We have $\lim_{\gamma \to 0} \sup \ensembleLigne{\Delta^\gamma(x)}{x \in \cl{\M}} = 0$.
\end{proposition}

\begin{proof}
  Let $\bgamma >0$ given by \Cref{lemma:lower_bound_prob}. Let  $x \in \cl{\M}$ and
  $\gamma \in \ooint{0, \bgamma}$. We have using \Cref{lemma:lower_bound_prob}
  \begin{equation}
    \textstyle{\int_{\rset^d} \bm{1}_{x + \sqrt{\gamma} z \in \M} \varphi(z) \rmd z \geq 1/4 .}
  \end{equation}
  We also have that
  \begin{equation}
    \textstyle{
      (1/\gamma) \int_{\rset^d} \bm{1}_{x + \sqrt{\gamma} z \in \M} \normLigne{\sqrt{\gamma} z}^4 \varphi(z) \rmd z \leq 3 \gamma d^2 .
      }
  \end{equation}
  Therefore, we get that for any $\gamma \in \ooint{0, \bgamma}$,
  $\Delta^\gamma(x) \leq 12 \gamma d^2$, which concludes the proof.
\end{proof}

In what follows, we define for any $\gamma >0$,
$\hat{\Sigma}^\gamma: \ \cl{\M} \to \mathrm{S}_d^+(\rset)$ given for any $x \in \cl{\M}$ by
\begin{equation}
  \label{eq:intermediate_diffusion}
 \textstyle{ \hat{\Sigma}^\gamma(x) =  \int_{\rset^d} \bm{1}_{x + \sqrt{\gamma} z \in \M} z \otimes z  \varphi(z) \rmd z  / \int_{\rset^d} \bm{1}_{x + \sqrt{\gamma} z \in \M} \varphi(z) \rmd z. }
\end{equation}

\begin{proposition}
  \label{prop:condition_d}
  There exists $\bgamma >0$ such that for any $x \in \cl{\M}$ and
  $\gamma \in \ooint{0, \bgamma}$ we have
  \begin{equation}
    \normLigne{\hat{\Sigma}^\gamma(x)} \leq 4d . 
  \end{equation}
\end{proposition}

\begin{proof}
  Let $x \in \cl{\M}$ and $\bgamma >0$ given by
  \Cref{lemma:lower_bound_prob}. For any $\gamma \in \ooint{0, \bgamma}$, we
  have using \Cref{lemma:lower_bound_prob}
  \begin{equation}
    \textstyle{\int_{\rset^d} \bm{1}_{x + \sqrt{\gamma} z \in \M} \varphi(z) \rmd z \geq 1/4 .}
  \end{equation}
  We also have that
  \begin{equation}
    \textstyle{
       \int_{\rset^d} \bm{1}_{x + \sqrt{\gamma} z \in \M} \normLigne{z}^2  \varphi(z) \rmd z \leq d ,
      }
    \end{equation}
    which concludes the proof.
  \end{proof}

  \subsection{Properties of large drift terms}
  \label{sec:large_drift_properties}

  Finally, we define for any $\gamma >0$,
$\hat{b}^\gamma: \ \cl{\M} \to \rset^d$ given for any $x \in \cl{\M}$ by
\begin{equation}
  \label{eq:intermediate_drift}
 \textstyle{ \hat{b}^\gamma(x) =  \gamma^{-1/2} \int_{\rset^d} \bm{1}_{x + \sqrt{\gamma} z \in \M}  z  \varphi(z) \rmd z  / \int_{\rset^d} \bm{1}_{x + \sqrt{\gamma} z \in \M} \varphi(z) \rmd z. }
\end{equation}
First, we show away from the boundary the drift $\hat{b}^\gamma$ converges to zero.

\begin{proposition}
  \label{prop:away_implies_small}
  There exists $\bgamma >0$ such that for any $\gamma \in \ooint{0, \bgamma}$,
  $r > 0$ and $x \in \cl{\M}$ such that $d(x, \partial \M) \geq r$ we have
  $\normLigne{\hat{b}^\gamma(x)} \leq 2d \Psi(r/\gamma,d)^{1/2}/\gamma^{1/2}$.
\end{proposition}

\begin{proof}
  Let $x \in \cl{\M}$ and $\bgamma >0$ given by
  \Cref{lemma:lower_bound_prob}. For any $\gamma \in \ooint{0, \bgamma}$ we have
  using \Cref{lemma:lower_bound_prob}
  \begin{equation}
    \textstyle{\int_{\rset^d} \bm{1}_{x + \sqrt{\gamma} z \in \M} \varphi(z) \rmd z \geq 1/4 .}
  \end{equation}
  We also have that
  \begin{align}
    \textstyle{\normLigne{\int_{\rset^d} \bm{1}_{x + \sqrt{\gamma} z \in \M}  z  \varphi(z) \rmd z}} &\leq \textstyle{\normLigne{\int_{\rset^d} \bm{1}_{\normLigne{z} \leq r/\gamma^{1/2}}  z  \varphi(z) \rmd z}} + \int_{\rset^d} \bm{1}_{\normLigne{z} \geq r/\gamma^{1/2}} \normLigne{z} \varphi(z) \rmd z \\
                                                                                                 &\leq \textstyle{2 \int_{\rset^d} \bm{1}_{\normLigne{z} \geq r/\gamma^{1/2}} \normLigne{z} \varphi(z) \rmd z} \leq 2d \Psi(r/\gamma,d)^{1/2}/\gamma^{1/2} ,
  \end{align}
  which concludes the proof. 
\end{proof}

We have the following corollary.
\begin{corollary}
  \label{coro:condition_c}
  There exists $\bgamma >0$ such that for any $\delta >0$ there exists
  $M_\delta >0$ such that for any $\gamma \in \ooint{0, \bgamma}$ and
  $x \in \cl{\M}$, $\normLigne{\hat{b}^\gamma(x)} \geq M_\delta$, then
  $\Phi(x) \leq \delta$.
\end{corollary}

\begin{proof}
  Let $\bgamma >0$ given by \Cref{lemma:lower_bound_prob}.  Let
  $f: \ \rset_+ \to \rset_+$ given for any $r > 0$ by
  $f(r) = \sup \ensembleLigne{\gamma > 0}{\Psi(r/\gamma,1)^{1/2}/\gamma^{1/2}}$. We
  have that $f$ is non-increasing and $\lim_{r \to 0} f(r)=+\infty$. Let
  $\delta > 0$ and $M_\delta =2df(\delta/C)$ with
  $C = \sup \ensembleLigne{\normLigne{\nabla \Phi(x)}}{x \in \cl{\M}}$. Let
  $\gamma \in \ooint{0, \bgamma}$ and $x \in \cl{\M}$ such that
  $\normLigne{\hat{b}^\gamma(x)} \geq M_\delta$ then using
  \Cref{prop:away_implies_small} we have that
  $d(x, \partial \M) \leq \delta / C$. Let $\bar{x} \in \partial \M$ such that
  $\normLigne{x - \bar{x}} = d(x, \partial \M)$. We have
  \begin{equation}
    \textstyle{\Phi(x) \leq \Phi(\bar{x}) + \int_0^1 \langle \nabla \Phi(\bar{x} + t(x-\bar{x})), x - \bar{x} \rangle \rmd t \leq \delta,}
  \end{equation}
  which concludes the proof.
\end{proof}

For ease of notation, for any $\gamma >0$, we define
$\bar{b}^\gamma = \gamma^{1/2} \hat{b}^\gamma$, the \emph{renormalized} version
of the drift.  First, we have the following result which will ensure that the
drift projected on the normal component does not vanish.

\begin{lemma}
  \label{lemma:complement_computation}
  There exists $\bgamma > 0$ such that for any $\gamma \in \ooint{0, \bgamma}$ and $x \in \msv$ we have
  \begin{equation}
    \langle \bar{b}^\gamma(x), \nabla \Phi(\bar{x}) \rangle \geq \normLigne{\bar{b}^\gamma(x)} - \psi(\gamma) ,
  \end{equation}
  with $\psi: \ \rset_+ \to \rset_+$ such that $\limsup_{\gamma \to 0} \psi(\gamma)/\sqrt{\gamma} <+\infty$.
\end{lemma}

\begin{proof}
  Let $x \in \cl{\M}$ and $\bgamma >0$ given by
  \Cref{lemma:lower_bound_prob}. For any $\gamma \in \ooint{0, \bgamma}$ we have
  using \Cref{lemma:lower_bound_prob}
  \begin{equation}
    \label{eq:lowr}
    \textstyle{\int_{\rset^d} \bm{1}_{x + \sqrt{\gamma} z \in \M} \varphi(z) \rmd z \geq 1/4 .}
  \end{equation}
  In addition, we have 
  \begin{align}
    \textstyle{\int_{\rset^d} \bm{1}_{x + \sqrt{\gamma}z \in \M} \langle z, \nabla \Phi(\bar{x}) \rangle \varphi(z) \rmd z}  & \geq  \textstyle{ \int_{\rset^d} \bm{1}_{x + \sqrt{\gamma}z \in \msc(x,\gamma)} \langle z, \nabla \Phi(\bar{x}) \rangle \varphi(z) \rmd z}  \\
       & \qquad \textstyle{-\int_{\rset^d} \bm{1}_{x + \sqrt{\gamma}z \in \M \cap \msc(x,\gamma)^\complementary} \langle z, \nabla \Phi(\bar{x}) \rangle \varphi(z)  }.
  \end{align}
  Using \Cref{lemma:control_outside}, we get that
  \begin{equation}
    \label{eq:ineq_remove}
    \textstyle{\int_{\rset^d} \bm{1}_{x + \sqrt{\gamma}z \in \M} \langle z, \nabla \Phi(\bar{x}) \rangle \varphi(z) \rmd z}   \geq  \textstyle{ \int_{\rset^d} \bm{1}_{x + \sqrt{\gamma}z \in \msc(x,\gamma)} \langle z, \nabla \Phi(\bar{x}) \rangle \varphi(z) \rmd z} - \psi(\gamma) . 
  \end{equation}
  Let $\{e_i \}_{i=1}^{d-1}$ a basis of $\nabla \Phi(\bar{x})^\perp$.
  Using \Cref{thm:tubul-neighb}-\ref{item:c}, we have that for any $i \in \{1, \dots, d-1\}$
  \begin{align}
    \textstyle{ \int_{\rset^d} \bm{1}_{x + \sqrt{\gamma}z \in \msc(x,\gamma)} \langle z, e_i \rangle \varphi(z) \rmd z}&\textstyle{= \int_{-\bar{\alpha}/\gamma^{1/2}}^{r/\gamma^{1/2}} \int_{\nabla \Phi(\bar{x})^\perp} \bm{1}_{\normLigne{v}^2 \leq (\gamma^{1/2} \alpha + \bar{\alpha})/\gamma} \langle v, e_i \rangle \varphi(v) \varphi(\alpha) \rmd v \rmd \alpha}. 
  \end{align}
Hence, combining this result and the Cauchy-Schwarz inequality we have for any $i \in \{1, \dots, d-1\}$
\begin{align}  \textstyle
  &\textstyle (\int_{\rset^d} \bm{1}_{x + \sqrt{\gamma}z \in \msc(x,\gamma)} \langle z, e_i \rangle \varphi(z) \rmd z)^2 = \textstyle (\int_{-\bar{\alpha}/\gamma^{1/2}}^{r/\gamma^{1/2}} \int_{\nabla \Phi(\bar{x})^\perp} \bm{1}_{\normLigne{v}^2 \geq (\gamma^{1/2} \alpha + \bar{\alpha})/\gamma} \langle v, e_i \rangle \varphi(v) \varphi(\alpha) \rmd v \rmd \alpha)^2 \\
  &\qquad \qquad \leq \textstyle \int_{\nabla \Phi(\bar{x})^\perp} \langle v, e_i \rangle^2 \varphi(v) \rmd v  (\int_{-\bar{\alpha}/\gamma^{1/2}}^{r/\gamma^{1/2}} \Psi((\bar{\alpha} + \alpha\gamma^{1/2})/\gamma, d-1)^{1/2} \varphi(\alpha) \rmd \alpha )^2 \\
    &\qquad \qquad \textstyle \leq (\int_{-\bar{\alpha}/\gamma^{1/2}}^{r/\gamma^{1/2}} \Psi((\bar{\alpha} + \alpha\gamma^{1/2})/\gamma, d-1)^{1/2} \varphi(\alpha) \rmd \alpha )^2 . 
\end{align}
Hence, using \Cref{lemma:tail_chi_square_int}, we get that
\begin{equation}
  \textstyle{\sum_{i=1}^{d-1}( \int_{\rset^d} \bm{1}_{x + \sqrt{\gamma}z \in \msc(x,\gamma)} \langle z, e_i \rangle \varphi(z) \rmd z)^2} \leq (d-1) \psi^2(\gamma) ,
\end{equation}
with $\psi$ given by \Cref{lemma:tail_chi_square_int} with $\beta_0=1/2$.  Therefore, we get that 
\begin{align}
  &\textstyle{(\int_{\rset^d} \bm{1}_{x + \sqrt{\gamma}z \in \msc(x,\gamma)} \langle z, \nabla \Phi(\bar{z}) \rangle \varphi(z) \rmd z)^2} \\
  &\qquad = \textstyle (\int_{\rset^d} \bm{1}_{x + \sqrt{\gamma} z \in \M} \varphi(z) \rmd z)^2 \textstyle{\normLigne{\bar{b}^\gamma(x)}^2 - \sum_{i=1}^{d-1}( \int_{\rset^d} \bm{1}_{x + \sqrt{\gamma}z \in \msc(x,\gamma)} \langle z, e_i \rangle \varphi(z) \rmd z)^2} \\
  &\qquad \geq \textstyle (\int_{\rset^d} \bm{1}_{x + \sqrt{\gamma} z \in \M} \varphi(z) \rmd z)^2 \normLigne{\bar{b}^\gamma(x)}^2  - \psi(\gamma)^2.
\end{align}
We conclude the proof upon using that for any $a,b \geq 0$, $(a+b)^{1/2} \leq a^{1/2} + b^{1/2}$ and \eqref{eq:lowr}.
\end{proof}




We are now ready to state the following lower bound on the drift.

\begin{proposition}
  \label{prop:lower-bound-drift}
  There exist $\bgamma >0$, $M \geq 0$ and $c >0$ such that for any
  $x \in \cl{\M}$ and $\gamma \in \ooint{0, \bgamma}$ if
  $\normLigne{\hat{b}^\gamma(x)} \geq M$ then $x \in \msv$ and
  \begin{equation}
    \min(\langle \hat{b}^\gamma(x), \nabla \Phi(x) \rangle, \langle \hat{b}^\gamma(x), \nabla \Phi(\bar{x}) \rangle) \geq c \normLigne{\hat{b}^\gamma(x)} .
  \end{equation}
\end{proposition}

\begin{proof}
  Let $\bgamma >0$ given by \Cref{lemma:lower_bound_prob} and
  $M_0 = 4 \sup \ensembleLigne{\psi(\gamma)/\gamma^{1/2}}{\gamma \in \ocint{0,
      \bgamma}}$.  In addition, let $c = 1/4$.  Using
  \Cref{prop:away_implies_small} and \Cref{thm:tubul-neighb}-\ref{item:a},
  there exists $M_1 \geq 0$ such that for any any $x \in \cl{\M}$, if
  $\normLigne{\hat{b}^\gamma(x)} \geq M_1$ then $x \in \msv$ and
  $x = \bar{x} + \alpha \nabla \Phi(\bar{x})$ with $\alpha \leq 1/(4C)$ and
  $C = \sup \ensembleLigne{\normLigne{\nabla^2 \Phi(x)}}{x \in \cl{\M}}$.  We
  denote $M = \max(M_0, M_1)$. Let $\gamma \in \ooint{0, \bgamma}$ and
  $x \in \cl{\M}$ such that $\normLigne{\hat{b}^\gamma(x)} \geq M$. Using
  \Cref{lemma:complement_computation}, we have that
  \begin{equation}
    \langle \hat{b}^\gamma(x), \nabla \Phi(\bar{x}) \rangle \geq \normLigne{\hat{b}^\gamma(x)} - \psi(\gamma)/\gamma^{1/2} .
  \end{equation}
  Using that $\psi(\gamma)/\gamma^{1/2} \leq M/2 \leq \normLigne{\hat{b}^\gamma(x)}/2$, we have 
  \begin{equation}
    \langle \hat{b}^\gamma(x), \nabla \Phi(\bar{x}) \rangle \geq (1/2) \normLigne{\hat{b}^\gamma(x)} .
  \end{equation}
  Since $\normLigne{x - \bar{x}} \leq \alpha \leq 1/(4C)$ we have
  $\langle \hat{b}^\gamma(x), \nabla \Phi(x) \rangle \geq (1/2 - C \alpha )
  \normLigne{\hat{b}^\gamma(x)} \geq \normLigne{\hat{b}^\gamma(x)} / 4$, which concludes the
  proof.
\end{proof}

\subsection{Convergence on compact sets}
\label{sec:convergence_on_compact}

In this section, we show the convergence of the drift and diffusion matrix on
compact sets. We recall that $\M$ does \emph{not} include its boundary
$\partial \M$.

\begin{proposition}
  \label{prop:condition_uno}
  For any compact set $\msk \subset \M$ and $\vareps >0$, there exists
  $\bgamma >0$ such that for any $\gamma \in \ooint{0, \bgamma}$ we have for any
  $x \in \msk$
  \begin{equation}
    \normLigne{\hat{b}^\gamma(x)} \leq \vareps , \qquad \normLigne{\hat{\Sigma}^\gamma(x) - \Id} \leq \vareps . 
  \end{equation}
\end{proposition}

\begin{proof}
  Let $\msk \subset \M$ be a compact set and $\gamma >0$. Since
  $\msk \cap \partial \M = \emptyset$, there exists $r > 0$ such that for any
  $x \in \msk$, $d(x, \partial \M) > r$. Therefore, we have that for any $x \in \msk$
  \begin{equation}
    \textstyle{ \normLigne{\hat{b}^\gamma(x)} = \gamma^{-1/2} \normLigne{\int_{x + \sqrt{\gamma}z \in \M} z \varphi(z) \rmd z} / \int_{x + \sqrt{\gamma}z \in \M}  \varphi(z) \rmd z . }
  \end{equation}
  In addition, using the Cauchy-Schwarz inequality we have 
  \begin{align}
    \textstyle{\normLigne{\int_{x + \sqrt{\gamma}z \in \M} z \varphi(z) \rmd z}} &\leq \normLigne{ \textstyle{\int_{\rset^d} z \varphi(z) \rmd z}} + \int_{\M^\complementary} \normLigne{z} \varphi(z) \rmd z \\
    &\leq \textstyle{\int_{\rset^d} \bm{1}_{\normLigne{z} \geq r / \gamma^{1/2}} \normLigne{z} \varphi(z) \rmd z \leq \sqrt{d} \Psi(r^2/\gamma, d)^{1/2} . }
  \end{align}
  Using \Cref{lemma:tail_chi_square} and \Cref{lemma:lower_bound_prob}, there
  exists $\bgamma_0 >0$ such that for any $\gamma \in \ooint{0, \bgamma_0}$ we
  have that for any $x \in \msk$
  \begin{equation}
    \normLigne{\hat{b}^\gamma(x)} \leq 4d\Psi(r^2/\gamma,1)^{1/2}/\gamma^{1/2} \leq \vareps ,
  \end{equation}
  which concludes the first part of the proof.
  Similarly, we have that for any $x \in \msk$
  \begin{align}
    \textstyle{\normLigne{\int_{x + \sqrt{\gamma}z \in \M} (z \otimes z - \Id) \varphi(z) \rmd z}} &\leq \normLigne{ \textstyle{\int_{\rset^d} (z\otimes z - \Id) \varphi(z) \rmd z}} + \int_{\M^\complementary} \normLigne{z} \varphi(z) \rmd z \\
                                                                                                   &\leq \textstyle{\int_{\rset^d} \bm{1}_{\normLigne{z} \geq r / \gamma^{1/2}} \normLigne{z \otimes z - \Id} \varphi(z) \rmd z} \\
    & \leq \sqrt{2}(1 + 3d^2)^{1/2} \Psi(r^2/\gamma, d)^{1/2} . 
  \end{align}
  Using \Cref{lemma:tail_chi_square} and \Cref{lemma:lower_bound_prob}, there
  exists $\bgamma_1 >0$ such that for any $\gamma \in \ooint{0, \bgamma_1}$, we
  have that for any $x \in \msk$
    \begin{equation}
      \normLigne{\hat{\Sigma}^\gamma(x) - \Id} \leq 4 \sqrt{2}(1 + 3d^2)^{1/2} \Psi(r^2/\gamma, 1)^{1/2} \leq \vareps , 
    \end{equation}
    which concludes the proof upon letting $\bgamma = \min(\bgamma_0, \bgamma_1)$.
\end{proof}

\subsection{Convergence on the boundary}
\label{sec:convergence_boundary}

Finally, we investigate the behavior at the boundary of the diffusion matrix and
the drift. First, we show that there is a lower bound to the diffusion matrix
near the boundary. Second, we show that the renormalized drift converges to the
outward normal.

\begin{proposition}
  \label{prop:condition_duo_iii}
  There exist $c > 0$ and $\bgamma >0$ such that for any
  $\gamma \in \ooint{0, \bgamma}$, $u \in \rset^d$ and $x \in \msv$ we have
  \begin{equation}
    \label{eq:lower_bound_diffusion}
   \langle u, \hat{\Sigma}^\gamma(x) u \rangle \geq c \normLigne{u}^2 . 
 \end{equation}
 In particular, there exist $r, \vareps >0$ such that for any
 $\gamma \in \ooint{0, \bgamma}$ and $x \in \cl{\M}$ with
 $d(x, \partial \M) \leq r$
 \begin{equation}
   \label{eq:lower_boundary}
  \langle \nabla \Phi(x), \hat{\Sigma}^\gamma(x) \nabla \Phi(x) \rangle \geq \vareps .
 \end{equation}
\end{proposition}

\begin{proof}
  First, we show \eqref{eq:lower_bound_diffusion}. Let $x \in \msv$. We have for any $u \in \rset^d$
  \begin{align}
    \label{eq:first-lower}
    \langle u, \hat{\Sigma}^\gamma(x) u \rangle &= \textstyle{\int_{\rset^d} \bm{1}_{x +\sqrt{\gamma}z \in \M} \langle z, u\rangle^2 \varphi(z) \rmd z / \int_{\rset^d} \bm{1}_{x +\sqrt{\gamma}z \in \M}  \rmd z } \\
    &\geq \textstyle{\int_{\rset^d} \bm{1}_{x+\sqrt{\gamma} z \in \msc(x,\gamma)} \langle z, u\rangle^2 \varphi(z) \rmd z} .
  \end{align}
  For any $u \in \rset^d$, let $\alpha_u = \langle u, \nabla \Phi(\bar{x}) \rangle$.
  Using \Cref{thm:tubul-neighb}-\ref{item:c} we have for any $u \in \rset^d$
  \begin{align}
    &\textstyle \int_{\rset^d} \bm{1}_{x+\sqrt{\gamma} z \in \msc(x,\gamma)} \langle z, u\rangle^2 \varphi(z) \rmd z \\
    &\qquad \textstyle =   \int_{-\bar{\alpha}/\gamma^{1/2}}^{r/\gamma^{1/2}} \int_{\nabla \Phi(\bar{x})^\perp} (\langle u, v \rangle + \alpha \alpha_u)^2 
      \bm{1}_{\norm{v}^2 \leq (\alpha \gamma^{1/2} + \bar{\alpha})/\gamma} \varphi(v) \varphi(\alpha) \rmd v \rmd \alpha \\
                                                                                                                 &\qquad \qquad \textstyle \geq  \int_{0}^{r/\gamma^{1/2}} \int_{\nabla \Phi(\bar{x})^\perp} (\langle u, v \rangle^2 + \alpha^2 \alpha_u^2)  \bm{1}_{\normLigne{v}^2 \leq (\alpha \gamma^{1/2} + \bar{\alpha})/\gamma} \varphi(v) \varphi(\alpha) \rmd v \rmd \alpha \\
    &\qquad \qquad\textstyle \geq \alpha_u^2 \int_0^{r/\gamma^{1/2}} \alpha^2 \varphi(\alpha) \rmd \alpha + \int_{-\bar{\alpha}/\gamma^{1/2}}^{r/\gamma^{1/2}} \int_{\nabla \Phi(\bar{x})^\perp} \langle u, v \rangle^2 \bm{1}_{\normLigne{v}^2 \leq (\alpha \gamma^{1/2} + \bar{\alpha})/\gamma} \varphi(v) \varphi(\alpha) \rmd v \rmd \alpha .
      \label{eq:lower_bound_decomp}
  \end{align}
Using Cauchy-Schwarz inequality, we have
\begin{equation}
  \label{eq:lower_bound_easy}
    \textstyle\int_0^{r/\gamma^{1/2}} \alpha^2 \varphi(\alpha) \rmd \alpha = (1/2) - \int_{r/\gamma^{1/2}}^{+\infty} \alpha^2 \varphi(\alpha) \rmd \alpha \geq (1/2) - 3 \Phi(r^2/\gamma,1)^{1/2}.
  \end{equation}
  In addition, using the Cauchy-Schwarz inequality, we have that
\begin{align}
  &\textstyle \int_{-\bar{\alpha}/\gamma^{1/2}}^{r/\gamma^{1/2}} \int_{\nabla \Phi(\bar{x})^\perp} \langle u, v \rangle^2 \bm{1}_{\normLigne{v}^2 \leq (\alpha \gamma^{1/2} + \bar{\alpha})/\gamma} \varphi(v) \varphi(\alpha) \rmd v \rmd \alpha\\
  & \qquad \qquad  \qquad = \textstyle \int_{\nabla \Phi(\bar{x})^\perp} \langle u, v \rangle^2 \varphi(v) \rmd v \int_{-\bar{\alpha}/\gamma^{1/2}}^{r/\gamma^{1/2}} \varphi(\alpha) \rmd \alpha \\
  &\qquad \qquad  \qquad \qquad \qquad \qquad \qquad  \textstyle - \int_{-\bar{\alpha}/\gamma^{1/2}}^{r/\gamma^{1/2}} \int_{\nabla \Phi(\bar{x})^\perp} \langle u, v \rangle^2 \bm{1}_{\normLigne{v}^2 \geq (\alpha \gamma^{1/2} + \bar{\alpha})/\gamma} \varphi(v) \varphi(\alpha) \rmd v \rmd \alpha \\
  & \qquad \geq \textstyle (\normLigne{u}^2 - \alpha_u^2) ((1/2) - \Phi(r^2/\gamma,1)) \\
  & \qquad \qquad \textstyle - \sqrt{3}(d-1) \normLigne{u}^2 \int_0^{+\infty} \Phi(\alpha/\gamma^{1/2},d-1)^{1/2} \varphi(\alpha - \bar{\alpha}/\gamma^{1/2}) \rmd \alpha .
\end{align}
Combining this result, \eqref{eq:lower_bound_easy}, \eqref{eq:lower_bound_decomp} and \Cref{lemma:tail_chi_square_int} there exists $\bgamma > 0$ such that for any $\gamma \in \ocint{0, \bgamma}$ and $ u \in \rset^d$
\begin{equation}
  \label{eq:inter_lower_sigma}
  \textstyle{\int_{\rset^d} \bm{1}_{x+\sqrt{\gamma} z \in \msc(x,\gamma)} \langle z, u\rangle^2 \varphi(z) \rmd z} \geq (1/4) \normLigne{u}^2 ,
\end{equation}
which concludes the proof of \eqref{eq:lower_bound_diffusion}.  Finally, using
\Cref{thm:tubul-neighb}-\ref{item:a}, we have that for any $x \in \cl{\M}$ if
$d(x, \partial \M) \leq R$ then $x \in \msv$. Let $r = \min(R, 1/(2C))$ with
$C = \sup \ensembleLigne{\normLigne{\nabla^2 \Phi(x)}}{x \in \cl{\M}}$.  We have
that for any $x \in \cl{\M}$ such that $d(x, \partial \M)\leq r$
\begin{equation}
  \normLigne{\nabla \Phi(x)} \geq \normLigne{\nabla \Phi(\bar{x}_0)} - C r \geq 1/2 ,
\end{equation}
where $\bar{x}_0$ is such that $\normLigne{x - \bar{x}_0} \leq r$ and
$\bar{x}_0 \in \partial \M$. Combining this result and
\eqref{eq:inter_lower_sigma} concludes the proof upon letting
$\vareps = 1/16$.
\end{proof}

Finally, we investigate the behavior of the normalized drift near the boundary.

\begin{proposition}
  \label{prop:condition_duo_iv}
  For any $\bar{x}_0 \in \partial \M$ and $\vareps >0$, there exist $\bgamma, r, M >0$ such that for any
  $x \in \cl{\M}$ and $\gamma \in \ooint{0, \bgamma}$ with
  $\normLigne{x - \bar{x}_0} \leq r$ and $\normLigne{\hat{b}^\gamma(x)} \geq M$ 
  \begin{equation}
    \normLigne{\hat{b}^\gamma(x)/\langle \hat{b}^\gamma(x), \nabla \Phi(x) \rangle - \nabla \Phi(\bar{x}_0)} \leq \vareps .
  \end{equation}
\end{proposition}

\begin{proof}
  Let $\bgamma$ be given by \Cref{prop:lower-bound-drift}.  Let $\psi$ given by
  \Cref{lemma:tail_chi_square_int} and
  $M_0 = \sup \ensembleLigne{\psi(\gamma)/\gamma^{1/2}}{\gamma \in \ooint{0,
      \bgamma}} < +\infty$. Let $M = 16 M_0 / (c\vareps^{1/2})$ with $c$ given in
  \Cref{prop:lower-bound-drift}. Let $R > 0$ given by
  \Cref{thm:tubul-neighb}-\ref{item:a} such that for any $x \in \cl{\M}$
  with $d(x, \partial \M)$ there exist $\bar{x} \in \partial \M$ and
  $\alpha \in \ccint{0, c \vareps /(4C)}$ such that
  $x = \bar{x} + \alpha \nabla \Phi(\bar{x})$ with
  $C = \sup \ensembleLigne{\normLigne{\nabla^2 \Phi(x)}}{x \in \cl{\M}}$ and $c$
  given in \Cref{prop:lower-bound-drift}.  Let
  $r = \min(\bar{r}, c \vareps/4, R)$ and $x \in \cl{M}$ with
  $\normLigne{x - \bar{x}_0} \leq r$.  First, since
  $d(x, \partial \M) \leq R$, there exist $\bar{x} \in \partial \M$ and
  $\alpha \in \ccint{0,\vareps/(4C)}$ such that
  $x = \bar{x} + \alpha \nabla \Phi(\bar{x})$. Therefore, we get that
  $\normLigne{\bar{x} - \bar{x}_0} \leq \vareps/(2C)$ and therefore
  $\normLigne{\nabla \Phi(\bar{x}_0) - \nabla \Phi(\bar{x})} \leq \vareps /2$.
  In addition, we have that
  \begin{align}
    &\normLigne{\hat{b}^\gamma(x)/\langle \hat{b}^\gamma(x), \nabla \Phi(x) \rangle - \hat{b}^\gamma(x)/\langle \hat{b}^\gamma(x), \nabla \Phi(\bar{x}) \rangle} \\
    & \qquad \qquad \leq  \normLigne{\hat{b}^\gamma(x)}^2 \normLigne{\nabla \Phi(x) - \nabla \Phi(\bar{x})} / (\langle \hat{b}^\gamma(x), \nabla \Phi(x) \rangle \langle \hat{b}^\gamma(x), \nabla \Phi(\bar{x}) \rangle) .
   \end{align}
   Using \Cref{prop:lower-bound-drift}, we get that
   \begin{equation}
     \normLigne{\hat{b}^\gamma(x)/\langle \hat{b}^\gamma(x), \nabla \Phi(x) \rangle - \hat{b}^\gamma(x)/\langle \hat{b}^\gamma(x), \nabla \Phi(\bar{x}) \rangle} \leq \vareps/4 .
   \end{equation}
   In what follows, we show that
  \begin{equation}
    \normLigne{\hat{b}^\gamma(x)/\langle \hat{b}^\gamma(x), \nabla \Phi(\bar{x}) \rangle - \nabla \Phi(\bar{x})}^2 \leq \vareps/2 .
  \end{equation}
  In particular, we show that for any $u \in \nabla \Phi(\bar{x})^\perp$ with $\normLigne{u}=1$,
  \begin{equation}
    \label{eq:ineq_b_gamma}
    \langle \hat{b}^\gamma(x), u \rangle ^2 \leq (\vareps/16) \langle \hat{b}^\gamma(x), \nabla \Phi(\bar{x}) \rangle^2.
  \end{equation}
  Assuming \eqref{eq:ineq_b_gamma}, letting 
  $u = (\hat{b}^\gamma(x) - \langle \hat{b}^\gamma(x), \nabla \Phi(\bar{x} \rangle)) /
  (\normLigne{\hat{b}^\gamma(x)}^2 - \langle \hat{b}^\gamma(x), \nabla \Phi(\bar{x}
  \rangle^2)^{1/2}$ and using that
  $\hat{b}^\gamma(x) = \langle \hat{b}^\gamma(x), u\rangle u + \langle \hat{b}^\gamma(x), \nabla
  \Phi(\bar{x}) \rangle \nabla \Phi(\bar{x})$ we have
  \begin{align}
    \normLigne{\hat{b}^\gamma(x) / \langle \hat{b}^\gamma(x), \nabla \Phi(x) \rangle - \nabla \Phi(\bar{x})} &\leq \normLigne{\hat{b}^\gamma(x) / \langle \hat{b}^\gamma(x), \nabla \Phi(\bar{x}) \rangle - \nabla \Phi(\bar{x})} \\
                                                                                                 & \qquad + \normLigne{\hat{b}^\gamma(x) / \langle \hat{b}^\gamma(x), \nabla \Phi(x) \rangle - \hat{b}^\gamma(x) / \langle \hat{b}^\gamma(x), \nabla \Phi(\bar{x})\rangle} \\
    &\leq \absLigne{\langle \hat{b}^\gamma(x), u\rangle / \langle \hat{b}^\gamma(x), \nabla \Phi(\bar{x}) \rangle} + \vareps / 4 \leq \vareps / 2 ,
  \end{align}
  which concludes the proof.  Let $u \in \nabla \Phi(\bar{x})^\perp$ with
  $\norm{u}=1$ and $\{e_i \}_{i=1}^{d-1}$ an orthonormal basis of
  $\nabla \Phi(\bar{x})^\perp$. There exist $\{a_i\}_{i=1}^{d-1}$ such that
  $\sum_{i=1}^{d-1} a_i^2 = 1$ and $u = \sum_{i=1}^{d-1} a_i e_i$.  Using
  \Cref{thm:tubul-neighb}-\ref{item:c}, we have that for any
  $i \in \{1, \dots, d-1\}$
  \begin{align}
    \textstyle{ \int_{\rset^d} \bm{1}_{x + \sqrt{\gamma}z \in \msc(x,\gamma)} \langle z, e_i \rangle \varphi(z) \rmd z}&\textstyle{= \int_{-\bar{\alpha}/\gamma^{1/2}}^{r/\gamma^{1/2}} \int_{\nabla \Phi(\bar{x})^\perp} \bm{1}_{\normLigne{v}^2 \leq (\gamma^{1/2} \alpha + \bar{\alpha})/\gamma} \langle v, e_i \rangle \varphi(v) \varphi(\alpha) \rmd v \rmd \alpha} \\
                                                                                                                   &= \textstyle{\int_{-\bar{\alpha}/\gamma^{1/2}}^{r/\gamma^{1/2}} \int_{\nabla \Phi(\bar{x})^\perp} \bm{1}_{\normLigne{v}^2 \geq (\gamma^{1/2} \alpha + \bar{\alpha})/\gamma} \langle v, e_i \rangle \varphi(v) \varphi(\alpha) \rmd v \rmd \alpha}\end{align}
Hence, combining this result and the Cauchy-Schwarz inequality we have for any $i \in \{1, \dots, d-1\}$
\begin{align}
  &\textstyle
  (\int_{\rset^d} \bm{1}_{x + \sqrt{\gamma}z \in \msc(x,\gamma)} \langle z, e_i \rangle \varphi(z) \rmd z)^2 = \textstyle (\int_{-\bar{\alpha}/\gamma^{1/2}}^{r/\gamma^{1/2}} \int_{\nabla \Phi(\bar{x})^\perp} \bm{1}_{\normLigne{v}^2 \geq (\gamma^{1/2} \alpha + \bar{\alpha})/\gamma} \langle v, e_i \rangle \varphi(v) \varphi(\alpha) \rmd v \rmd \alpha)^2 \\
  & \qquad \leq \textstyle \int_{\nabla \Phi(\bar{x})^\perp} \langle v, e_i \rangle^2 \varphi(v) \rmd v  (\int_{-\bar{\alpha}/\gamma^{1/2}}^{r/\gamma^{1/2}} \Psi((\bar{\alpha} + \alpha\gamma^{1/2})/\gamma, d-1)^{1/2} \varphi(\alpha) \rmd \alpha )^2 . 
\end{align}
Hence, we get that
\begin{equation}
  \textstyle{\sum_{i=1}^{d-1}a_i^2 ( \int_{\rset^d} \bm{1}_{x + \sqrt{\gamma}z \in \msc(x,\gamma)} \langle z, e_i \rangle \varphi(z) \rmd z)^2} \leq \normLigne{u}^2 \psi^2(\gamma) ,
\end{equation}
with $\psi$ given by \Cref{lemma:tail_chi_square_int}. Recalling that $\normLigne{\hat{b}^\gamma(x)} \geq M$ we have 
\begin{equation}
  \langle \hat{b}^\gamma(x), u \rangle^2 \leq 16\psi(\gamma)^2/\gamma \leq c^2 (\vareps / 16) M^2 \leq (\vareps / 16) \langle \hat{b}^\gamma(x), \nabla \Phi(\bar{x}) \rangle^2 ,
\end{equation}
which concludes the proof.
\end{proof}

\subsection{Submartingale problem and weak solution}
\label{sec:submartingale}

We are now ready to conclude the proof. 

\begin{theorem}
  \label{thm:martingale-problem}
  There exists $\Pbb^\star$ a distribution on $\rmD(\ccint{0,T}, \cl{\M})$ such
  that $\lim_{\gamma \to 0} \hat{\Pbb}^\gamma = \Pbb^\star$. In addition, for any
  $f \in \rmc^{1,2}(\ccint{0,T} \times \cl{\M}, \rset)$ with
  $\langle \nabla \Phi(\bar{x}), \nabla f(x) \rangle \geq 0$ for any
  $t \in \ccint{0,T}$ and $x \in \partial \M$, we have that the process
  $(f(t, \omega(t)))_{t \in \ccint{0,T}}$ given for any $t \in \ccint{0,T}$
  \begin{equation}
    \textstyle
    f(t, \omega(t)) - \int_0^t (\partial_s f(s, \omega(s) + \tfrac{1}{2} \Delta f(s, \omega(s))) \bm{1}_{\M}(\omega(s)) \rmd s ,
  \end{equation}
  is a $\Pbb$ submartingale.
\end{theorem}

\begin{proof}
  Condition (A) \cite[p.197]{stroock1971diffusion} is a consequence of
  \Cref{prop:condition_a}.  Condition (B) \cite[p.197]{stroock1971diffusion} is
  a consequence of \Cref{prop:lower-bound-drift}.  Condition (C)
  \cite[p.198]{stroock1971diffusion} is a consequence of
  \Cref{coro:condition_c}.  Condition (D) \cite[p.198]{stroock1971diffusion} is
  a consequence of \Cref{prop:condition_d}.  We fix $\rho =0$ and condition (1)
  \cite[p.203]{stroock1971diffusion} is a consequence of
  \Cref{prop:condition_uno}.  Condition (2)-(iii)
  \cite[p.203]{stroock1971diffusion} is a consequence of
  \Cref{prop:condition_duo_iii}.  Condition (2)-(iv)
  \cite[p.203]{stroock1971diffusion} is a consequence of
  \Cref{prop:condition_duo_iv}.  We conclude upon using \cite[Theorem
  6.3]{stroock1971diffusion} and \cite[Theorem
  5.8]{stroock1971diffusion}.
\end{proof}

We finally conclude the proof of \Cref{thm:weak_convergence_appendix} upon using the
results of \cite{kang2017submartingale} which establish the link between a weak
solution to the reflected SDE and the solution to a submartingale problem.

\begin{theorem}
  \label{thm:subm-probl-weak}
  For any $T \geq 0$, $(\hat{\bfX}_t^\gamma)_{t \in \ccint{0,T}}$ weakly converges to
  $(\bfX_t)_{t \in \ccint{0,T}}$ such that for any $t \in \ccint{0,T}$
  \begin{equation}
    \textstyle
    \bfX_t = x + \bfB_t - \bfk_t , \qquad \absLigne{\bfk}_t = \int_0^t \bm{1}_{\bfX_s \in \partial \M} \rmd \absLigne{\bfk}_s , \qquad \bfk_t = \int_0^t \bfn(\bfX_s) \rmd \absLigne{\bfk}_s . 
  \end{equation}
\end{theorem}

\begin{proof}
  Using \Cref{thm:martingale-problem} and \cite[Theorem 1, Proposition
  2.12]{kang2017submartingale}, we have that $\Pbb$ in
  \Cref{thm:martingale-problem}is associated with a solution to the extended
  Skorokhod problem. We conclude that a solution to the extended Skorokhod
  problem is a solution to the Skorokhod problem using \cite[Corollary
  2.10]{ramanan2006reflected}.
\end{proof}

\subsection{Extension to the Metropolis process}
\label{sec:extens-lazy-proc}

We recall that the Metropolis process is defined as follows. Let
$(X_k^\gamma)_{k \in \nset}$ given for any $\gamma >0$ and $k \in \nset$ by
$X_0^\gamma = x \in \cl{\M}$ and for
$X_{k+1}^\gamma = X_k^\gamma + \sqrt{\gamma} Z_k$ if
$X_k^\gamma + \sqrt{\gamma} Z_k^\gamma \in \cl{\M}$ and $X_k^\gamma$ otherwise,
$Z_k \sim \mathrm{N}(0, \Id)$.  We recall that $\hat{b}^\gamma$,
$\hat{\Sigma}^\gamma$ and $\hat{\Delta}^\gamma$ are given by
\eqref{eq:intermediate_Delta}, \eqref{eq:intermediate_diffusion} and
\eqref{eq:intermediate_drift}. In particular, denoting $\hat{\Kker}^\gamma$ the
Markov kernel associated with $(\hat{X}_k^\gamma)_{k \in \nset}$, i.e.~
$\hat{\Kker}^\gamma: \ \M \times \mcb{\M} \to \ccint{0,1}$ such that for any
$x \in \M$, $\hat{\Kker}^\gamma(x, \cdot)$ is a probability measure, for any
$\msa \in \mcb{\M}$, $\hat{\Kker}^\gamma(\cdot, \msa)$ is a measurable function
and
$\expeLigne{\bm{1}_{\msa}(\hat{X}_1^\gamma) \ | \ \hat{X}_0^\gamma=x} =
\hat{\Kker}^\gamma(x, \msa)$. We have that for any $\gamma > 0$ and $x \in \M$
\begin{align}
  &\textstyle \hat{b}^\gamma(x) = (1/\gamma) \int_{\M} (y - x) \hat{\Kker}^\gamma(x, \rmd y) , \\
  &\textstyle \hat{\Sigma}^\gamma(x) = (1/\gamma)\int_{\M} (y - x)^{\otimes 2} \hat{\Kker}^\gamma(x, \rmd y) , \\
  &\textstyle \hat{\Delta}^\gamma(x) = (1/\gamma)\int_{\M} \normLigne{y - x}^4 \hat{\Kker}^\gamma(x, \rmd y) .
\end{align}
In what follows, we denote
$a^\gamma(x) = \expeLigne{\bm{1}_{x + \sqrt{\gamma} Z_0 \in \M}}$. Denote
$\Kker^\gamma$ the kernel associated with $(X_k^\gamma)_{k \in \nset}$. We have
that for any $\msa \in \mcb{\M}$, $\gamma >0$ and $x \in \M$
\begin{align}
  \Kker^\gamma(x,\msa) &= \expeLigne{\bm{1}_{X_{k+1}^\gamma \in \msa} \bm{1}_{x + \sqrt{\gamma} Z_{k+1} \in \M}} + (1-a^\gamma(x)) \bm{1}_{\msa}(x) \\
                &= a^\gamma(x) \hat{\Kker}^\gamma(x, \msa) + (1-a^\gamma(x)) \bm{1}_{\msa}(x) .                  \label{eq:new_kernel}
\end{align}
We define for any $\gamma > 0$ and $x \in \M$
\begin{align}
  &\textstyle b^\gamma(x) = (1/\gamma) \int_{\M} (y - x) \Kker^\gamma(x, \rmd y) , \\
  &\textstyle \Sigma^\gamma(x) = (1/\gamma)\int_{\M} (y - x)^{\otimes 2} \Kker^\gamma(x, \rmd y) , \\
  &\textstyle \Delta^\gamma(x) = (1/\gamma)\int_{\M} \normLigne{y - x}^4 \Kker^\gamma(x, \rmd y) .
\end{align}
Using \eqref{eq:new_kernel}, we get that for any $\gamma >0$ and $x \in \M$
\begin{equation}
  \label{eq:relation_coeff}
  b^\gamma(x) = a^\gamma(x) \hat{b}^\gamma(x) , \qquad \Sigma^\gamma(x) = a^\gamma(x) \hat{\Sigma}^\gamma(x) , \qquad \Delta^\gamma(x) = a^\gamma(x) \hat{\Delta}^\gamma(x) .
\end{equation}
Using \Cref{lemma:lower_bound_prob}, we have that for any
$\gamma \in \ooint{0, \bgamma}$ and $x \in \M$, $a^\gamma(x) \geq 1/4$.

In order to conclude for the convergence of the Metropolis process we adapt
\Cref{thm:martingale-problem} and \Cref{thm:subm-probl-weak}. We define
$\bfX^\gamma : \rset_+ \to \cl{\M}$ given for any $k \in \nset$ by
$\bfX^\gamma_{k\gamma} = X_k^\gamma$ and for any
$t \in \coint{k\gamma, (k+1)\gamma}$, $\bfX^\gamma_t = X_{k}^\gamma$. Note
that $(\bfX_t)_{t \in \ccint{0,T}}$ is a $\rmD(\ccint{0,T}, \cl{\M})$ valued
random variable, where $\rmD(\ccint{0,T}, \cl{\M})$ is the space of
right-continuous with left-limit processes which take values in $\cl{\M}$. We
denote $\Pbb^\gamma$ the distribution of $(\bfX_t)_{t \in \ccint{0,T}}$ on
$\rmD(\ccint{0,T}, \cl{\M})$.

\begin{theorem}
  \label{thm:martingale-problem-lazy}
  There exists $\Pbb^\star$ a distribution on $\rmD(\ccint{0,T}, \cl{\M})$ such
  that $\lim_{\gamma \to 0} \Pbb^\gamma = \Pbb^\star$. In addition, for any
  $f \in \rmc^{1,2}(\ccint{0,T} \times \cl{\M}, \rset)$ with
  $\langle \nabla \Phi(\bar{x}), \nabla f(x) \rangle \geq 0$ for any
  $t \in \ccint{0,T}$ and $x \in \partial \M$, we have that the process
  $(f(t, \omega(t)))_{t \in \ccint{0,T}}$ given for any $t \in \ccint{0,T}$
  \begin{equation}
    \textstyle
    f(t, \omega(t)) - \int_0^t (\partial_s f(s, \omega(s) + \tfrac{1}{2} \Delta f(s, \omega(s))) \bm{1}_{\M}(\omega(s)) \rmd s ,
  \end{equation}
  is a $\Pbb$ submartingale.
\end{theorem}

\begin{proof}
Condition (A) \cite[p.197]{stroock1971diffusion} is a consequence of
\Cref{prop:condition_a} and \eqref{eq:relation_coeff}.  Condition (B)
\cite[p.197]{stroock1971diffusion} is a consequence of
\Cref{prop:lower-bound-drift} and \eqref{eq:relation_coeff}.  Condition (C)
\cite[p.198]{stroock1971diffusion} is a consequence of \Cref{coro:condition_c}
and \eqref{eq:relation_coeff}.  Condition (D) \cite[p.198]{stroock1971diffusion}
is a consequence of \Cref{prop:condition_d} and \eqref{eq:relation_coeff}.  We
fix $\rho =0$ and condition (1) \cite[p.203]{stroock1971diffusion} is a
consequence of \Cref{prop:condition_uno} and that $\lim_{\gamma \to 0} a^\gamma = 1$
uniformly on compact subsets $\msk \subset \M$.  Condition (2)-(iii)
\cite[p.203]{stroock1971diffusion} is a consequence of
\Cref{prop:condition_duo_iii} and \eqref{eq:relation_coeff}.  Condition (2)-(iv)
\cite[p.203]{stroock1971diffusion} is a consequence of
\Cref{prop:condition_duo_iv} and \eqref{eq:relation_coeff}.  We conclude upon
using \cite[Theorem 6.3]{stroock1971diffusion} and \cite[Theorem
5.8]{stroock1971diffusion}.
\end{proof}

\begin{theorem}
  \label{thm:subm-probl-weak-lazy}
  For any $T \geq 0$, $(\bfX_t^\gamma)_{t \in \ccint{0,T}}$ weakly converges to
  $(\bfX_t)_{t \in \ccint{0,T}}$ such that for any $t \in \ccint{0,T}$
  \begin{equation}
    \textstyle
    \bfX_t = x + \bfB_t - \bfk_t , \qquad \absLigne{\bfk}_t = \int_0^t \bm{1}_{\bfX_s \in \partial \M} \rmd \absLigne{\bfk}_s , \qquad \bfk_t = \int_0^t \bfn(\bfX_s) \rmd \absLigne{\bfk}_s . 
  \end{equation}
\end{theorem}

\begin{proof}
The proof is identical to \Cref{thm:subm-probl-weak}.  
\end{proof}


\section{Modelling geospatial data within non-convex boundaries}
\label{app_sec:wildfire_data}

To demonstrate the ability of the proposed method to model distributions whose support is restricted to manifolds with highly non-convex boundaries, we derived a geospatial dataset based on the historical wildfire incidence rate within the continental United States (described in in~\Cref{app_sec:wildfire_data_dataset}) and, using the corresponding country borders, trained a constrained diffusion model by adapting the point-in-spherical-polytope conditions outlined in \cite{ketzner2022robust} (described in~\Cref{app_sec:wildfire_data_algo}).

\subsection{Derivation of bounded geospatial dataset}
\label{app_sec:wildfire_data_dataset}

\begin{figure}[h]
    \centering
    \includegraphics[width=0.5\linewidth]{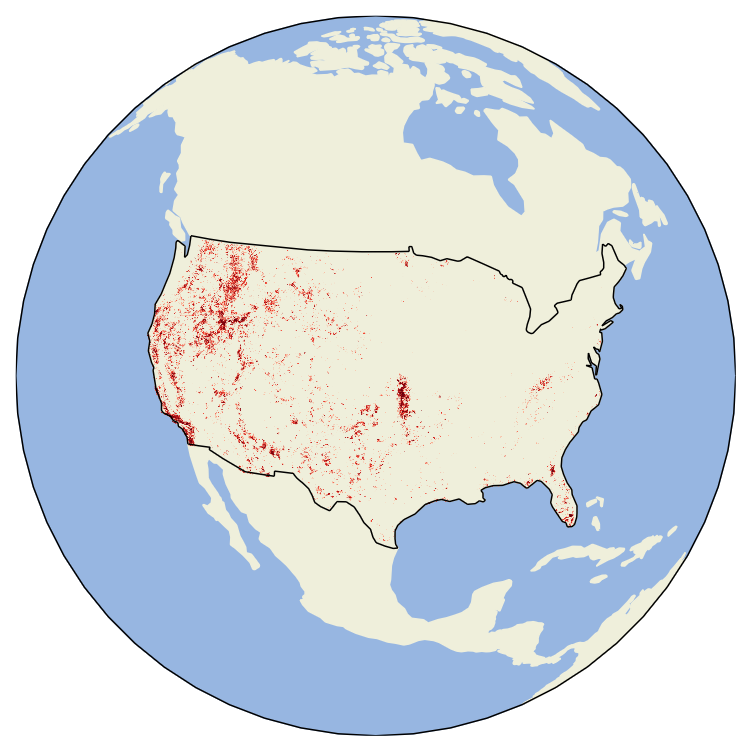}
    \caption{Orthographic projection of the wildfire dataset described in~\Cref{app_sec:wildfire_data}. The projection is aligned with the centroid of the continental United States and zoomed in ten-fold for visual clarity. All visualisations of geospatial data were generated using the \textsc{GeoViews} \citep{geoviews} and \textsc{DataShader} \cite{datashader} libraries.}
    \label{fig_app:wildfire_data_subsample}
\end{figure}

Specifically, we retrieved the rasterised version of the wildfire data provided by \citet{welty2020combined}, converted it to a spherical geodetic coordinate system using the \textsc{Cartopy} library \citep{Cartopy}, and drew a weighted subsample of size \num{1e6}. We then retrieved the country borders of the continental United States from \cite{natural_earth_data} and mapped them to the same geodetic reference frame as the wildfire data. A visualization of the resulting dataset is presented in~\Cref{fig_app:wildfire_data_subsample}. 

\subsection{Point-in-spherical-polytope algorithms}
\label{app_sec:wildfire_data_algo}

The support of the data-generating distribution we aim to approximate is thus restricted to a highly non-convex spherical polytope $\mathbb{P}\in\mathcal{S}^2$ given by the country borders of the continental United States. To determine whether a query point $q\in\mathcal{S}^2$ is within $\mathbb{P}$, we adapt an efficient reformulation of the point-in-spherical-polygon algorithm \citep{bevis1989locating} presented in \cite{ketzner2022robust}.
The algorithm requires the provision of a reference point $r\in\mathcal{S}^2$ known to be located in $\mathbb{P}$ and determines whether $q$ is inside or outside the polygon by checking whether the geodesic between $r$ and $q$ crosses the polygon an even or odd number of times. 
Letting $\Hat{x}\in\mathbb{R}^3$ denote the Cartesian coordinates of a point $x\in\mathcal{S}^2$, \cite{ketzner2022robust}  rely on a Cartesian reference coordinate system $\Hat{Q}$ (with its $z$-axis given by $\Hat{r}$) and the corresponding spherical coordinate system $Q$ to decompose the edge-crossing condition of \citet{bevis1989locating} into two efficiently computable parts. That is, the geodesic between $q$ and $r$ crosses an edge $e_i=(v_i, v_j)$ of the polygon if:
\begin{enumerate}[label=(\roman*)]
    \item the longitude of $q$ in $Q$ is bounded by the longitudes of $v_i$ and $v_j$ in $Q$, i.e. 
    $$
    \phi_{Q}(q)\in[\min(\phi_{Q}(v_i), \phi_{Q}(v_j)), \max(\phi_{Q}(v_i), \phi_{Q}(v_j))],
    $$
    \item the plane specified by the normal vector $\Hat{p}_i=\Hat{v}_i\times \Hat{v}_j$ represents an equator that separates ${q}$ and ${r}$ into two different hemispheres, i.e.
    $$
    \operatorname{sign}(\langle \Hat{p}_i, \Hat{r} \rangle \cdot \langle \Hat{p}_i, \Hat{q} \rangle) = -1.
    $$
\end{enumerate}

Especially when $\mathbb{P}$ is fixed and the corresponding coordinate transformations and normal vectors can be precomputed for each edge, this algorithm affords an efficient and parallelisable approach to determining whether any given point on $\mathcal{S}^2$ is contained by a spherical polytope.

\section{Supplementary Experimental Results}
\label{sec:experimental_appendix}

\subsection{Evaluating log-barrier and Euclidean models}
\label{sec:app_exp_barrier_results}

Following \cite{fishman2023diffusion}, we approached the empirical evaluation of our Metropolis model by computing the maximum mean discrepancy (MMD) \citep{gretton2012kernel} between samples from the true distribution and the trained diffusion models. The MMD is a statistic that quantifies the similarity of two samples by computing the distance of their respective mean embeddings in a reproducing kernel Hilbert space. For this, we use an RBF kernel with the same length scales as the standard deviations of the normal distributions used to generate the synthetic distribution. We sum these RBF kernels by the weights of the corresponding components of the synthetic Gaussian mixture model.

This is essential to be able to include the log-barrier in the comparison since the log-barrier methods suffer severe instabilities around the boundary, as the space is stretched to more and more. These instabilities cause the problems in fitting the log-barrier model and in computing the likelihood using the log-barrier model.  


\begin{table}[H]
    \caption{
    Maximum mean discrepancy (MMD) ($\downarrow$) of a held-out test set from a synthetic bimodal distribution over convex subsets of $\Rbb^d$ bounded by the hypercube \([-1,1]^d\) and unit simplex \(\Delta^d\). Means and standard deviations are computed over 3 different runs. 
    }
    \label{tab:synthetic_polytope_mmd}
    \vspace{1em}
    \centering

\setlength{\tabcolsep}{15.0pt}
\begin{adjustbox}{max width=\textwidth}
\begin{tabular}{lllrrr}
\toprule
\multirow{2}{*}{Manifold}&\multirow{2}{*}{Dimension}&\multirow{2}{*}{Process}& \multicolumn{2}{c}{MMD} & \% in Manifold \\
          &    &        &  mean &   std &             mean \\
\midrule
\multirow{12}{*}{$\Delta^d$}& \multirow{4}{*}{2}  & Euclidean & 0.027 & 0.011 &            0.969 \\
          &    & Log-Barrier & 0.050 & 0.012. &         1.000 \\
          &    & Reflected & 0.041 & 0.008 &            1.000 \\
          &    & Rejection & 0.030 & 0.002 &               1.000 \\\cmidrule{2-6}
          & \multirow{4}{*}{3}  & Euclidean & 0.032 & 0.015 &            0.969 \\
          &    & Log-Barrier & 0.238 & 0.009 &          1.000 \\
          &    & Reflected & 0.179 & 0.013 &            1.000 \\
          &    & Rejection & 0.111 & 0.002 &               1.000 \\\cmidrule{2-6}
          & \multirow{4}{*}{10} & Euclidean & 0.028 & 0.001 &            0.946 \\
          &    & Log-Barrier & 0.275 & 0.0015 &         1.000 \\
          &    & Reflected & 0.233 & 0.004 &            1.000 \\
          &    & Rejection & 0.226 & 0.005 &               1.000 \\\midrule
\multirow{12}{*}{$[0, 1]^d$}& \multirow{4}{*}{2}  & Euclidean & 0.069 & 0.004 &            0.992 \\
          &    & Log-Barrier & 0.66 & 0.006 &           1.000 \\
          &    & Reflected & 0.048 & 0.012 &            1.000 \\
          &    & Rejection & 0.025 & 0.005 &               1.000 \\\cmidrule{2-6}
          & \multirow{4}{*}{3}  & Euclidean & 0.074 & 0.004 &            0.991 \\
          &    & Log-Barrier & 0.209 & 0.0077 &         1.000 \\
          &    & Reflected & 0.085 & 0.006 &            1.000 \\
          &    & Rejection & 0.049 & 0.006 &               1.000 \\\cmidrule{2-6}
          & \multirow{4}{*}{10} & Euclidean & 0.086 & 0.007 &            0.968 \\
          &    & Log-Barrier & 0.330 & 0.004 &          1.000 \\
          &    & Reflected & 0.314 & 0.049 &            1.000 \\
          &    & Rejection & 0.138 & 0.007 &               1.000 \\
\bottomrule
\end{tabular}
\end{adjustbox}
\end{table}

From the results in~\Cref{tab:synthetic_polytope_mmd}, it is clear that the log-barrier approach performs significantly worse than the Reflected model and the Metropolis models across all settings. This, in conjunction with numerical instabilities we encountered when attempting to evaluate sample likelihoods with the log-barrier models as presented in \cite{fishman2023diffusion}, motivated us to focus on the Reflected and Metropolis models in the main text.

Additionally, we note that the unconstrained Euclidean models outperform the constrained methods on both the simplex and the hypercube as the dimensionality of the problem space increases. Especially on the simplex, we attribute this performance primarily to the fact that the synthetic distribution is simply a standard Normal with only a small portion close to the boundary. The amount of reflection needed to model the distribution decreases in higher dimensions, as the mass of the Normal distribution gets increasingly concentrated---which Euclidean diffusion models will fit well. This same dynamic is partially responsible for the hypercube performance.

\subsection{Implementational details}
\label{sec:app_implementational_details}

All source code that is needed to reproduce the results presented below is made available under \href{https://github.com/oxcsml/score-sde/tree/metropolis}{https://github.com/oxcsml/score-sde/tree/metropolis}, which requires a supporting package to handle the different geometries that is available under \href{https://github.com/oxcsml/geomstats/tree/polytope}{https://github.com/oxcsml/geomstats/tree/polytope}.

We use the same architecture in all of our experiments: a 6-layer MLP with 512 hidden units and sine activation functions, except in the output layer, which uses a linear activation function. Following \cite{fishman2023diffusion}, we implement a simple linear function that scales the score by the distance to the boundary, approaching zero within $\epsilon = 0.01$ of the boundary. This ensures the score obeys the Neumann boundary conditions required by the reflected Brownian Motion. For the geospatial dataset within non-convex country borders, we do not use distance rescaling. Instead, we substitute it with a series of step functions to rescale the score. This is a proof-of-concept to show that even when computing the distance is hard, simple and efficient approximations suffice. When constructing Riemannian diffusion models on the torus and sphere for the protein and geospatial datasets, we follow \cite{debortoli2022riemannian} and include an additional preconditioner for the score on the manifold. We \emph{do not} use the residual trick or the standard deviation trick, which are both common score-rescaling functions in image model architectures; in our setting, we find that they adversely affect model training.

For the forward/reverse process we always set $T=1$, $\beta_0=\num{1e-3}$ and then tune $\beta_1$ to ensure that the forward process just reaches the invariant distribution with a linear $\beta$-schedule. At sampling time we use $N=100$ steps of the discretised process. We discretise the training process by selecting a random $N$ between 0 and 100 for each example, rolling out to that time point. This lets us cheaply implement a simple variance reduction technique: we take multiple samples from this trajectory by selecting multiple random $N$ to save for each example. This technique was originally described in \cite{fishman2023diffusion} and we find it is also helpful for our Metropolis models. For all experiments, we use the $\mathrm{ism}$ loss with a modified weighting function of $(1 + t)$, which we found to be essential to model training. All experiments use a batch size of 256 with 8 repeats per batch. For training, we use a learning rate of \num{2e-4} with a cosine learning rate schedule. We trained for 100,000 batches on the synthetic examples and 300,000 batches on the real-world examples (robotics, proteins, wildfires). 

We selected these hyperparameters from a systematic search over learning rates (\num{6e-4}, \num{2e-4}, \num{6e-5}, \num{2e-5}), learning rate schedules (cosine, log-linear), and batch sizes (128, 256, 512, 1024) on synthetic examples for the reflected and log-barrier models. Similar parameters worked well for both, and we used those for our Metropolis models to allow a straightforward comparison. We tried $N=100,1000$ for several synthetic examples but found that very large rollout times actually hurt performance for the Metropolis model, though the log-barrier performed a bit better with longer rollouts and the reflected was the same.

All models were trained on a single NVIDIA GeForce GTX 1080 GPU. All of the Metropolis models presented here can easily be trained on this hardware in under 4 hours. The runtime for the log-barrier and reflected models is considerably longer.

\subsection{Synthetic Distributions on Constrained Manifolds of Increasing Dimensionality}
\label{sec:app_exp_synthetic_tasks}

\begin{figure}[H]
\centering
\begin{subfigure}{.23\textwidth}
  \centering
  \includegraphics[width=\textwidth]{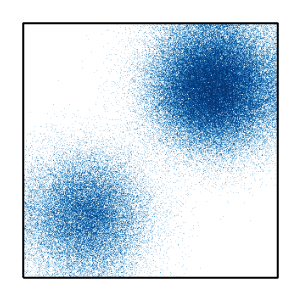}
  \caption{Data distribution}
  \label{fig:app:synthetic_hypercube_data}
\end{subfigure}%
\begin{subfigure}{.23\textwidth}
  \centering
  \includegraphics[width=\textwidth]{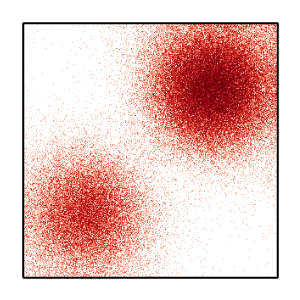}
  \caption{Metropolis model}
  \label{fig:app:synthetic_hypercube_rejection}
\end{subfigure}
\begin{subfigure}{.23\textwidth}
  \centering
  \includegraphics[width=\textwidth]{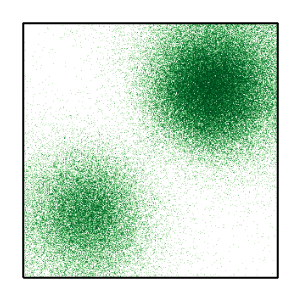}
  \caption{Reflected model}
  \label{fig:app:synthetic_hypercube_reflected}
\end{subfigure}%
\begin{subfigure}{.23\textwidth}
  \centering
  \includegraphics[width=\textwidth]{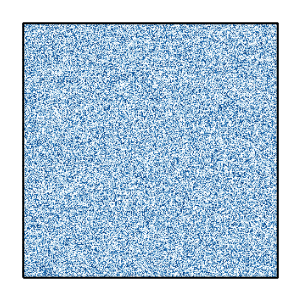}
  \caption{Uniform distribution}
  \label{fig:app:synthetic_hypercube_uniform}
\end{subfigure}
  \caption{Qualitiative comparison of samples from the data distribution, our Metropolis model, a Reflected model and the uniform distribution for a synthetic bimodal distribution on $[-1, 1]^2$.}
  \label{fig:app:synthetic_hypercube}
\end{figure}

\begin{figure}[H]
\centering
\begin{subfigure}{.23\textwidth}
  \centering
  \includegraphics[width=\textwidth]{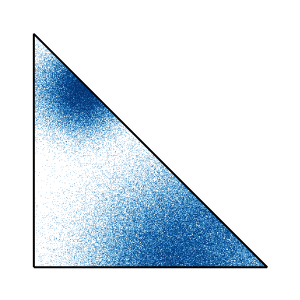}
  \caption{Data distribution}
  \label{fig:app:synthetic_simplex_data}
\end{subfigure}%
\begin{subfigure}{.23\textwidth}
  \centering
  \includegraphics[width=\textwidth]{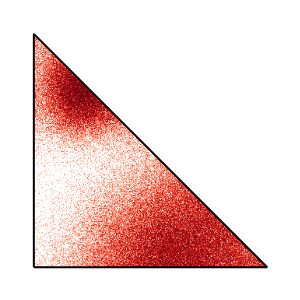}
  \caption{Metropolis model}
  \label{fig:app:synthetic_simplex_rejection}
\end{subfigure}
\begin{subfigure}{.23\textwidth}
  \centering
  \includegraphics[width=\textwidth]{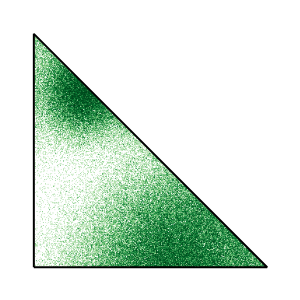}
  \caption{Reflected model}
  \label{fig:app:synthetic_simplex_reflected}
\end{subfigure}%
\begin{subfigure}{.23\textwidth}
  \centering
  \includegraphics[width=\textwidth]{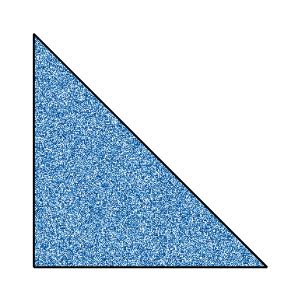}
  \caption{Uniform distribution}
  \label{fig:app:synthetic_simplex_uniform}
\end{subfigure}
  \caption{Qualitiative comparison of samples from the data distribution, our Metropolis model, a Reflected model and the uniform distribution for a synthetic bimodal distribution on $\Delta^2$.}
  \label{fig:app:synthetic_simplex}
\end{figure}

\subsection{Constrained Configurational Modelling of Robotic Arms}
\label{sec:app_robotic_arms_spd}

The following univariate marginal and pairwise bivariate plots visualise the distribution of different samples in
\begin{enumerate}[label=(\roman*)]
    \item the three dimensions needed to describe an ellipsoid \(M=\begin{bmatrix}
    l_{1} & l_{2}\\
    l_{2} & l_3
    \end{bmatrix}\in\c{S}_{++}^2\) and 
    \item the two dimensions needed to describe a location in \(\Rbb^2\).
\end{enumerate}

\subsubsection{Visualisation of samples from the data distribution}

\begin{figure}[H]
  \centering
  \begin{subfigure}[t]{0.45\textwidth}
    \centering
    \includegraphics[width=\textwidth]{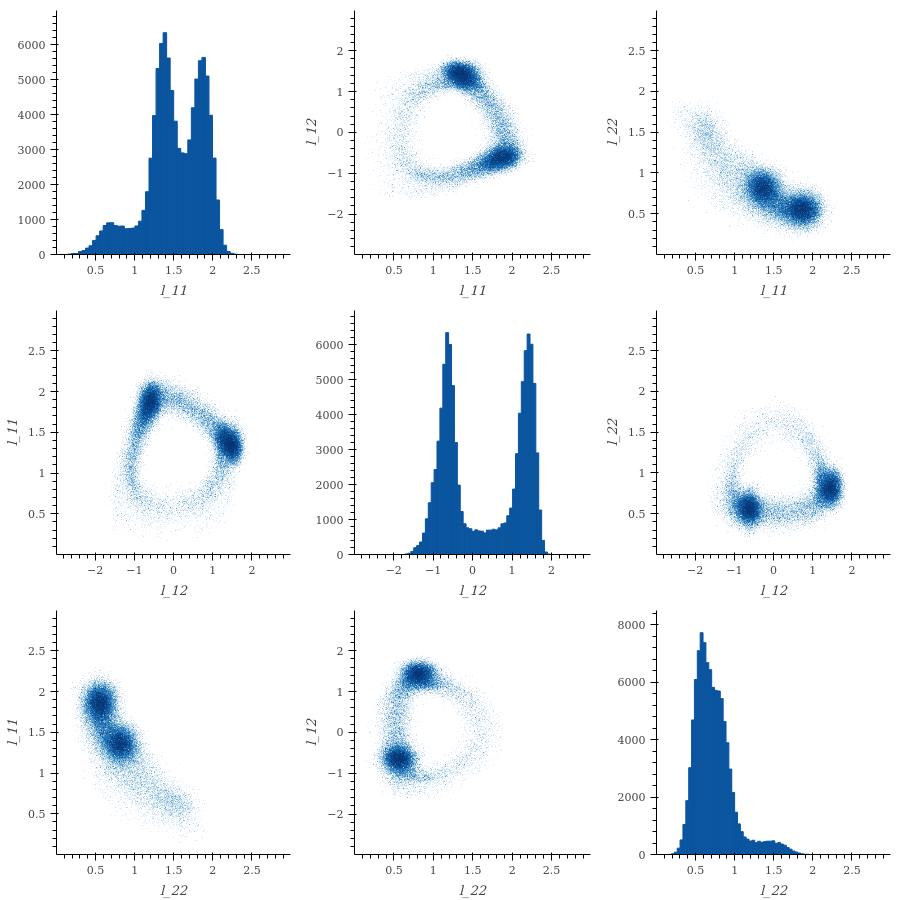}
    \caption{Plots of the univariate marginal and pairwise bivariate distributions of \num{1e5} samples from the data distribution in \(\c{S}_{++}^2\).}
    \label{fig:app_pairplots_robot_data_ellips}
  \end{subfigure}
  \hfill
  \begin{subfigure}[t]{0.45\textwidth}
    \centering
    \includegraphics[width=\textwidth]{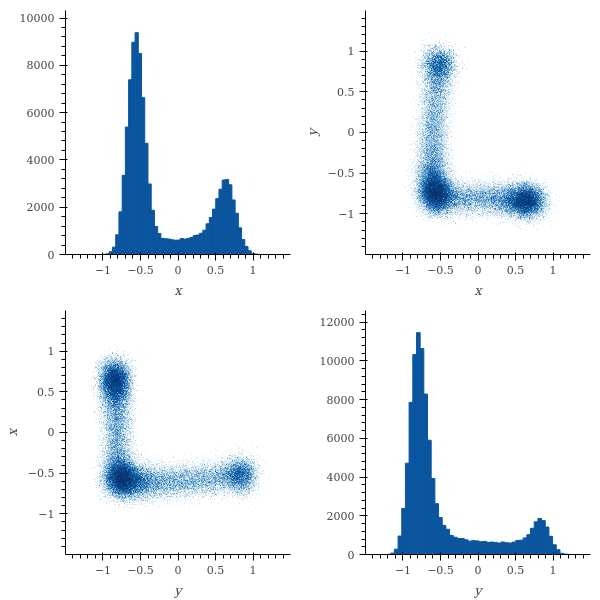}
    \caption{Plots of the univariate marginal and pairwise bivariate distributions of \num{1e5} samples from the data distribution in \(\Rbb^2\).}
    \label{fig:app_pairplots_robot_data_location}
  \end{subfigure}
  \caption{Visualisation of the data distribution in \(\c{S}_{++}^2\times\Rbb^2\) using univariate marginal and pairwise bivariate plots.}
  \label{fig:app_pairplots_robot_data}
\end{figure}

\subsubsection{Visualisation of samples from our Metropolis sampling-based diffusion model}

\begin{figure}[H]
  \centering
  \begin{subfigure}[t]{0.45\textwidth}
    \centering
    \includegraphics[width=\textwidth]{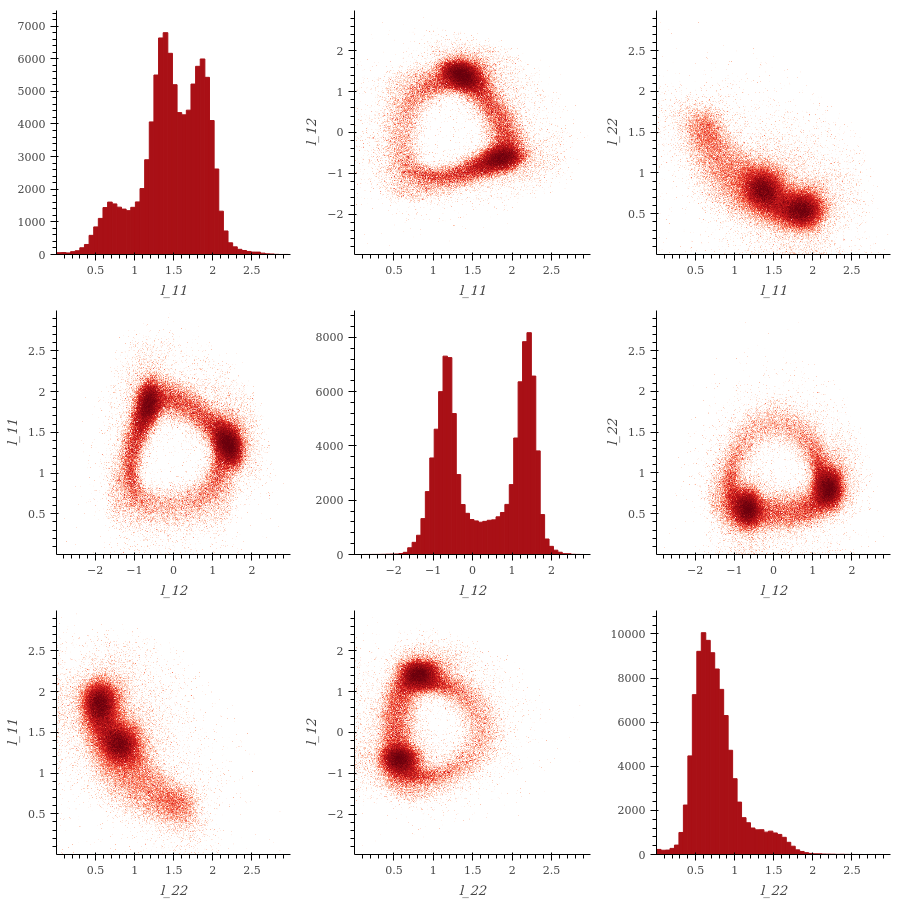}
    \caption{Plots of the univariate marginal and pairwise bivariate distributions of \num{1e5} samples from our Metropolis sampling-based diffusion model in \(\c{S}_{++}^2\).}
    \label{fig:app_pairplots_robot_rejection_ellips}
  \end{subfigure}
  \hfill
  \begin{subfigure}[t]{0.45\textwidth}
    \centering
    \includegraphics[width=\textwidth]{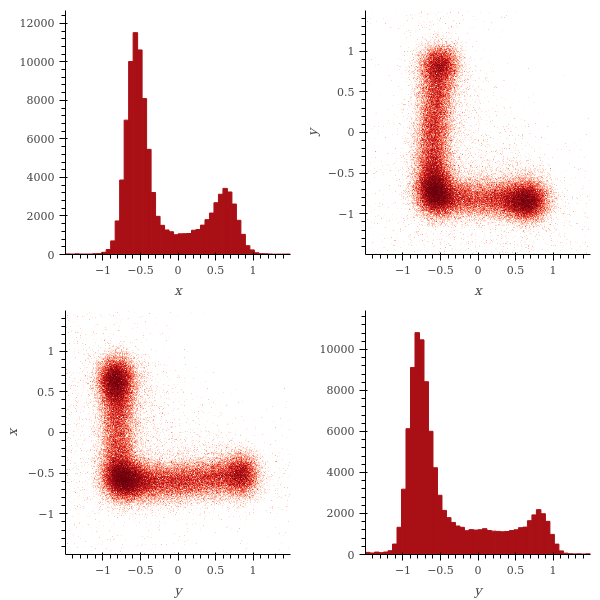}
    \caption{Plots of the univariate marginal and pairwise bivariate distributions of \num{1e5} samples from our Metropolis sampling-based diffusion model in \(\Rbb^2\).}
    \label{fig:app_pairplots_robot_rejection_location}
  \end{subfigure}
  \caption{Visualisation of the distribution learned by our Metropolis sampling-based diffusion model in \(\c{S}_{++}^2\times\Rbb^2\) using univariate marginal and pairwise bivariate plots.}
  \label{fig:app_pairplots_robot_rejection}
\end{figure}

\subsubsection{Visualisation of samples from a reflected Brownian motion-based diffusion model}

\begin{figure}[H]
  \centering
  \begin{subfigure}[t]{0.45\textwidth}
    \centering
    \includegraphics[width=\textwidth]{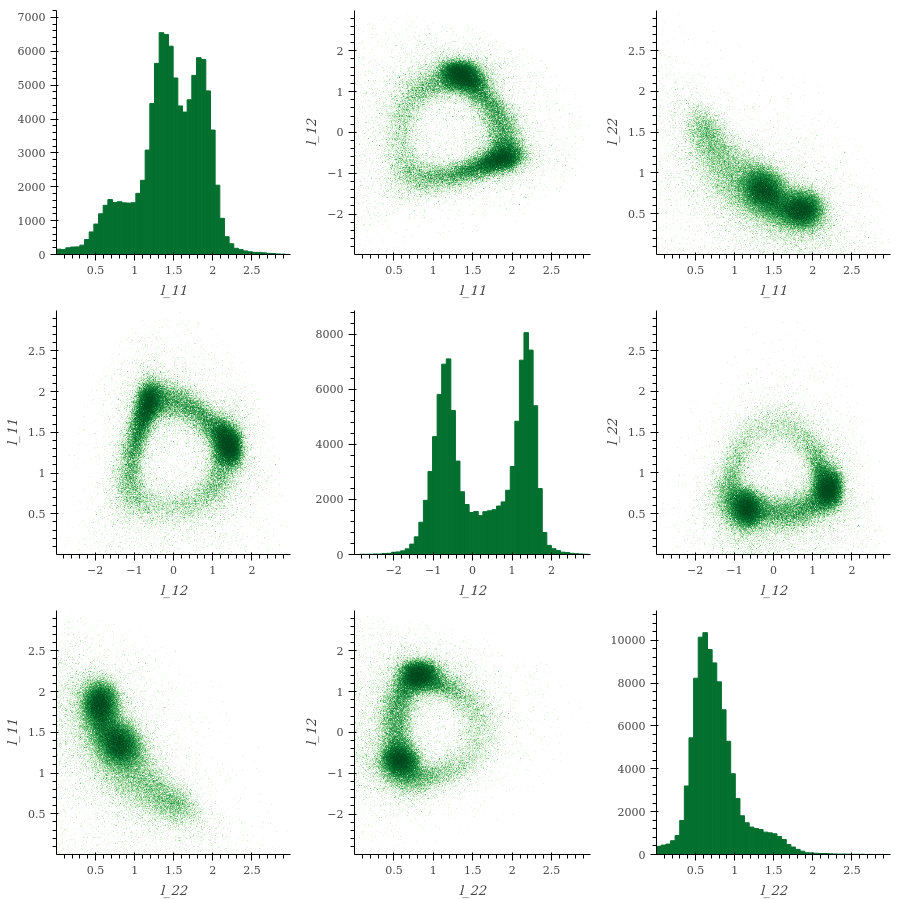}
    \caption{Plots of the univariate marginal and pairwise bivariate distributions of \num{1e5} samples from a reflected Brownian motion-based diffusion model in \(\c{S}_{++}^2\).}
    \label{fig:app_pairplots_robot_reflected_ellips}
  \end{subfigure}
  \hfill
  \begin{subfigure}[t]{0.45\textwidth}
    \centering
    \includegraphics[width=\textwidth]{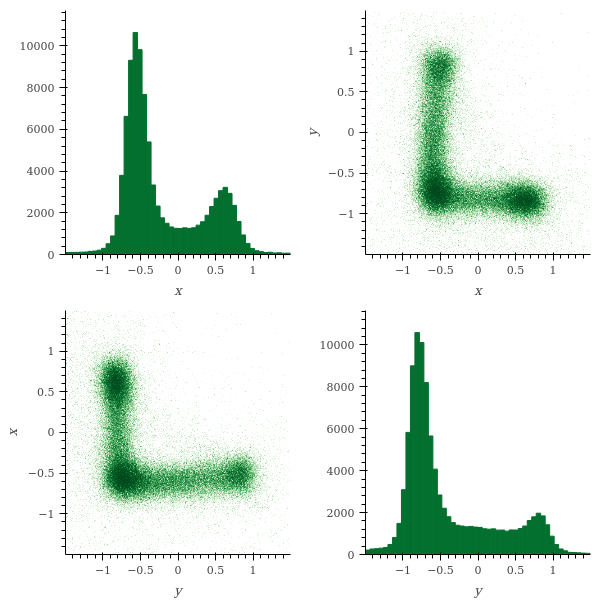}
    \caption{Plots of the univariate marginal and pairwise bivariate distributions of \num{1e5} samples from a reflected Brownian motion-based diffusion model in \(\Rbb^2\).}
    \label{fig:app_pairplots_robot_reflected_location}
  \end{subfigure}
  \caption{Visualisation of the distribution learned by a reflected Brownian motion-based diffusion model in \(\c{S}_{++}^2\times\Rbb^2\) using univariate marginal and pairwise bivariate plots.}
  \label{fig:app_pairplots_robot_reflected}
\end{figure}

\subsubsection{Visualisation of samples from the uniform distribution}

\begin{figure}[H]
  \centering
  \begin{subfigure}[t]{0.45\textwidth}
    \centering
    \includegraphics[width=\textwidth]{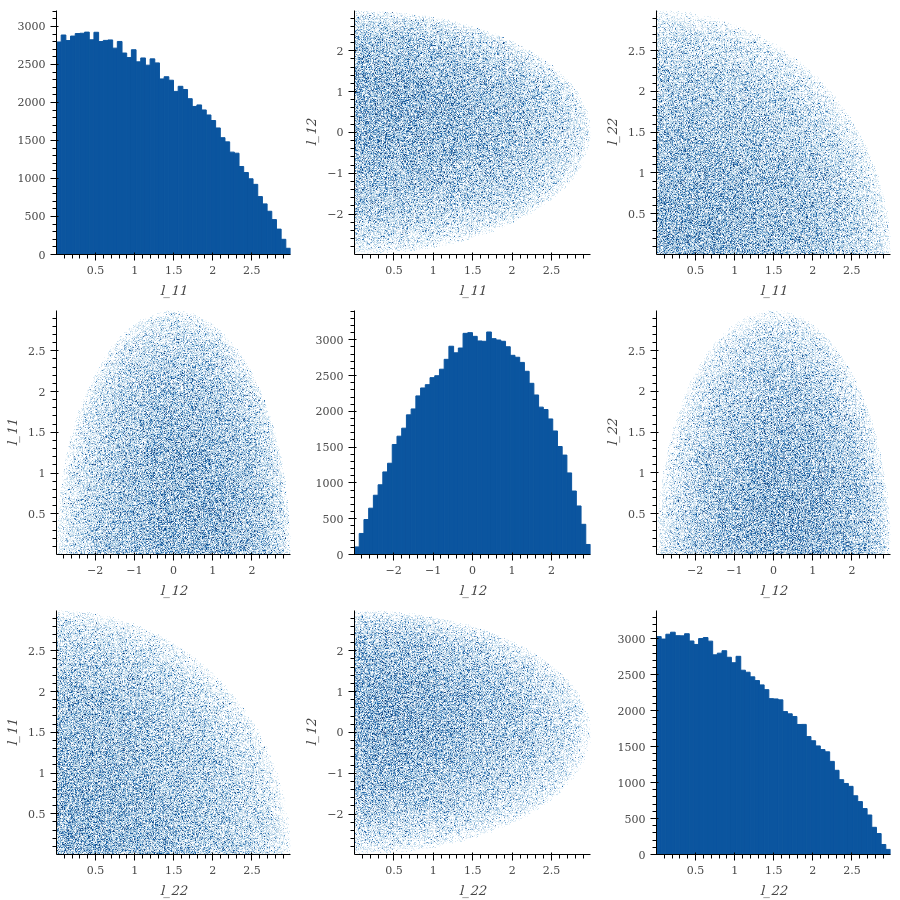}
    \caption{Plots of the univariate marginal and pairwise bivariate distributions of \num{1e5} samples from the uniform distribution in \(\c{S}_{++}^2\).}
    \label{fig:app_pairplots_robot_reflected_ellips}
  \end{subfigure}
  \hfill
  \begin{subfigure}[t]{0.45\textwidth}
    \centering
    \includegraphics[width=\textwidth]{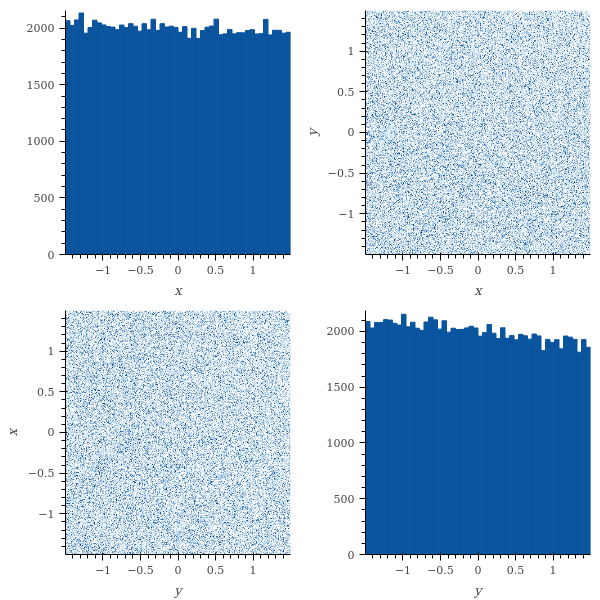}
    \caption{Plots of the univariate marginal and pairwise bivariate distributions of \num{1e5} samples from the uniform distribution in \(\Rbb^2\).}
    \label{fig:app_pairplots_robot_reflected_location}
  \end{subfigure}
  \caption{Visualisation of the uniform distribution in \(\c{S}_{++}^2\times\Rbb^2\) using univariate marginal and pairwise bivariate plots.}
  \label{fig:app_pairplots_robot_uniform}
\end{figure}

\subsection{Conformational Modelling of Protein Backbones}
\label{sec:app_loop_pairplots}

The following univariate marginal and pairwise bivariate plots visualise the distribution of different samples in
\begin{enumerate*}[label=(\roman*)]
    \item the polytope \(\mathbb{P}\subset\Rbb^3\) and
    \item the torus \(\mathbb{T}^4\)
\end{enumerate*}
used to parametrise the conformations of a polypeptide chain of length \(N=6\) with coinciding endpoints. We refer to \cite{han2006inverse} for full detail on the reparametrisation and to \cite{fishman2023diffusion} for a full description of the dataset.

\subsubsection{Visualisation of samples from the data distribution}

\begin{figure}[H]
  \centering
  \begin{subfigure}[t]{0.45\textwidth}
    \centering
    \includegraphics[width=\textwidth]{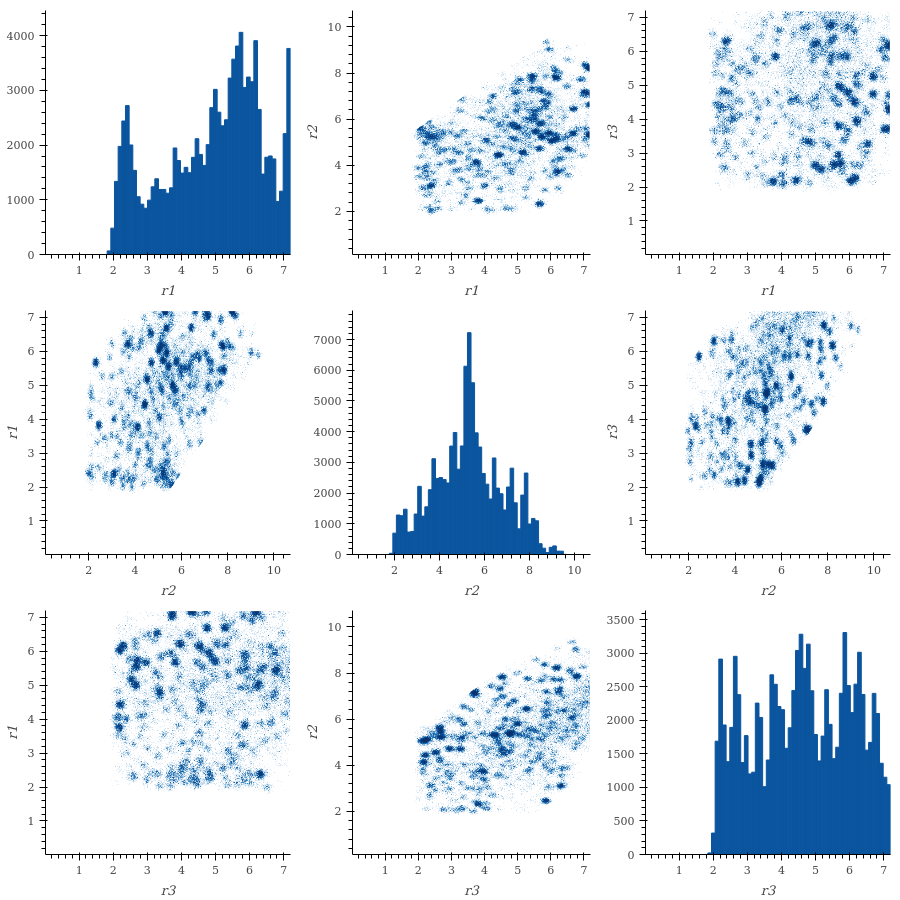}
    \caption{Plots of the univariate marginal and pairwise bivariate distributions of \num{1e5} samples from the data distribution in \(\mathbb{P}\subset\Rbb^3\).}
    \label{fig:app_pairplots_loop_polytope_data}
  \end{subfigure}
  \hfill
  \begin{subfigure}[t]{0.45\textwidth}
    \centering
    \includegraphics[width=\textwidth]{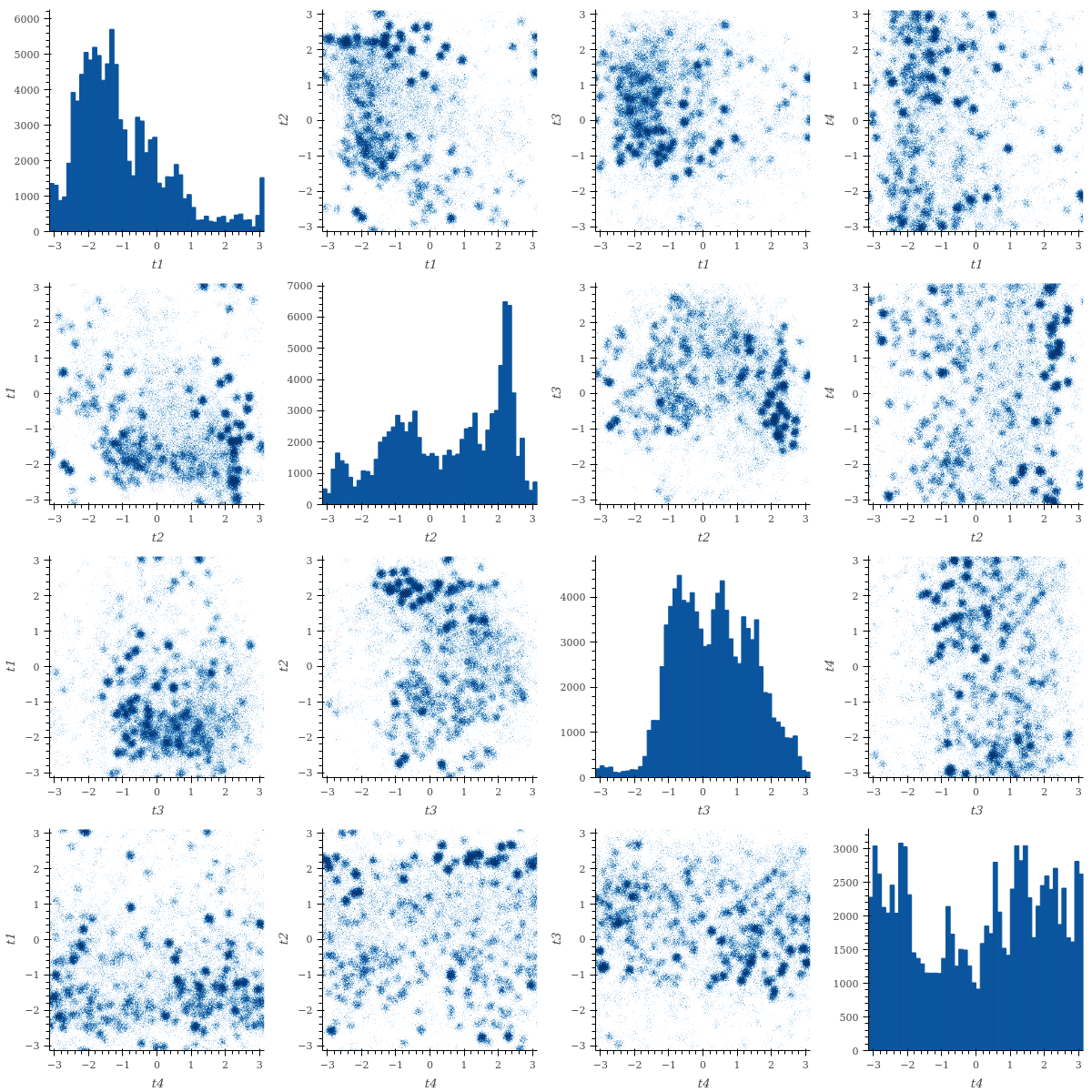}
    \caption{Plots of the univariate marginal and pairwise bivariate distributions of \num{1e5} samples from the data distribution in \(\mathbb{T}^4\).}
    \label{fig:app_pairplots_loop_torus_data}
  \end{subfigure}
  \caption{Visualisation of the data distribution in \(\mathbb{P}\subset\Rbb^3\times\mathbb{T}^4\) using univariate marginal and pairwise bivariate plots.}
  \label{fig:app_pairplots_loop_data}
\end{figure}

\subsubsection{Visualisation of samples from our Metropolis sampling-based diffusion model}

\begin{figure}[H]
  \centering
  \begin{subfigure}[t]{0.45\textwidth}
    \centering
    \includegraphics[width=\textwidth]{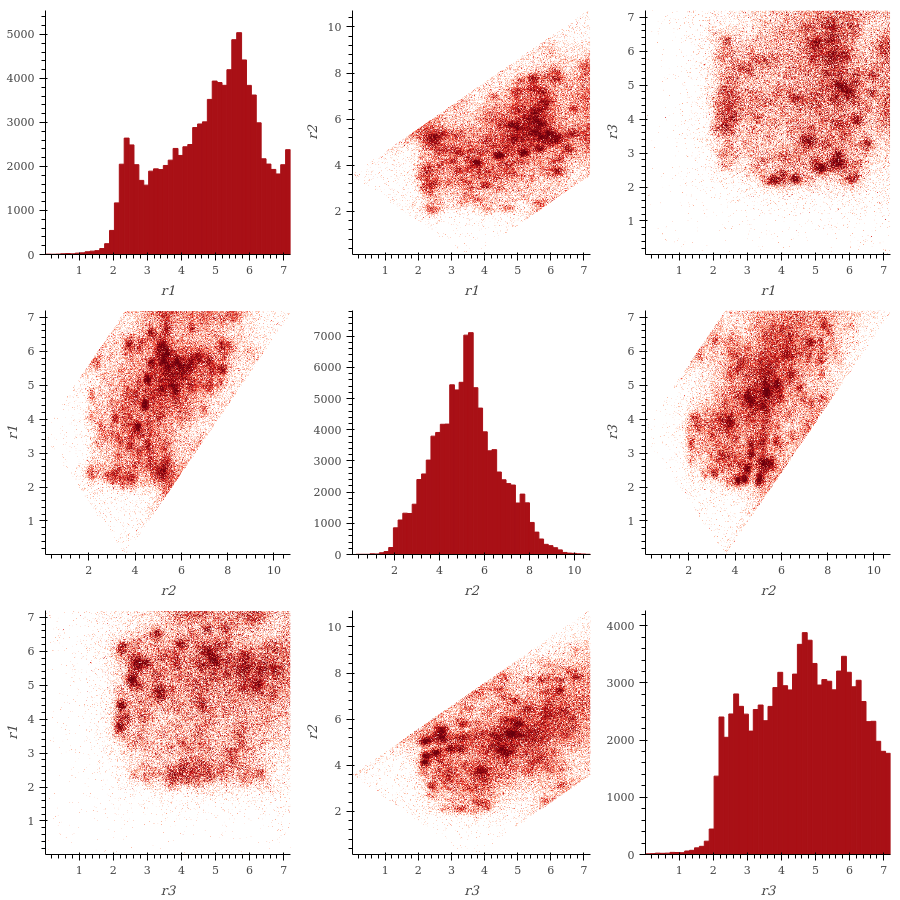}
    \caption{Plots of the univariate marginal and pairwise bivariate distributions of \num{1e5} samples from our Metropolis model in \(\mathbb{P}\subset\Rbb^3\).}
    \label{fig:app_pairplots_loop_polytope_rejection}
  \end{subfigure}
  \hfill
  \begin{subfigure}[t]{0.45\textwidth}
    \centering
    \includegraphics[width=\textwidth]{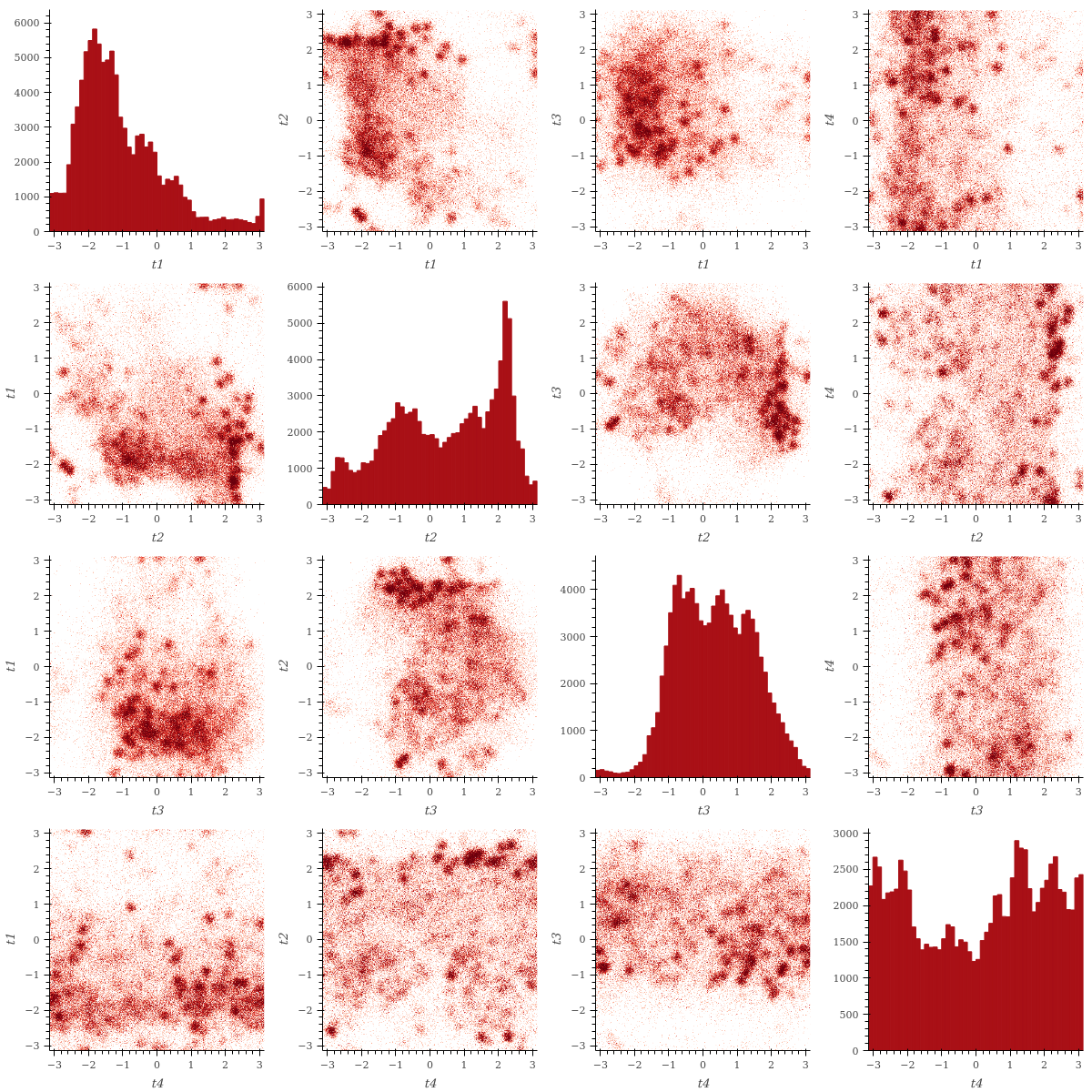}
    \caption{Plots of the univariate marginal and pairwise bivariate distributions of \num{1e5} samples from our Metropolis model in \(\mathbb{T}^4\).}
    \label{fig:app_pairplots_loop_torus_rejection}
  \end{subfigure}
  \caption{Visualisation of the distribution learned by our Metropolis model in \(\mathbb{P}\subset\Rbb^3\times\mathbb{T}^4\) using univariate marginal and pairwise bivariate plots.}
  \label{fig:app_pairplots_loop_rejection}
\end{figure}

\subsubsection{Visualisation of samples from a reflected Brownian motion-based diffusion model}

\begin{figure}[H]
  \centering
  \begin{subfigure}[t]{0.45\textwidth}
    \centering
    \includegraphics[width=\textwidth]{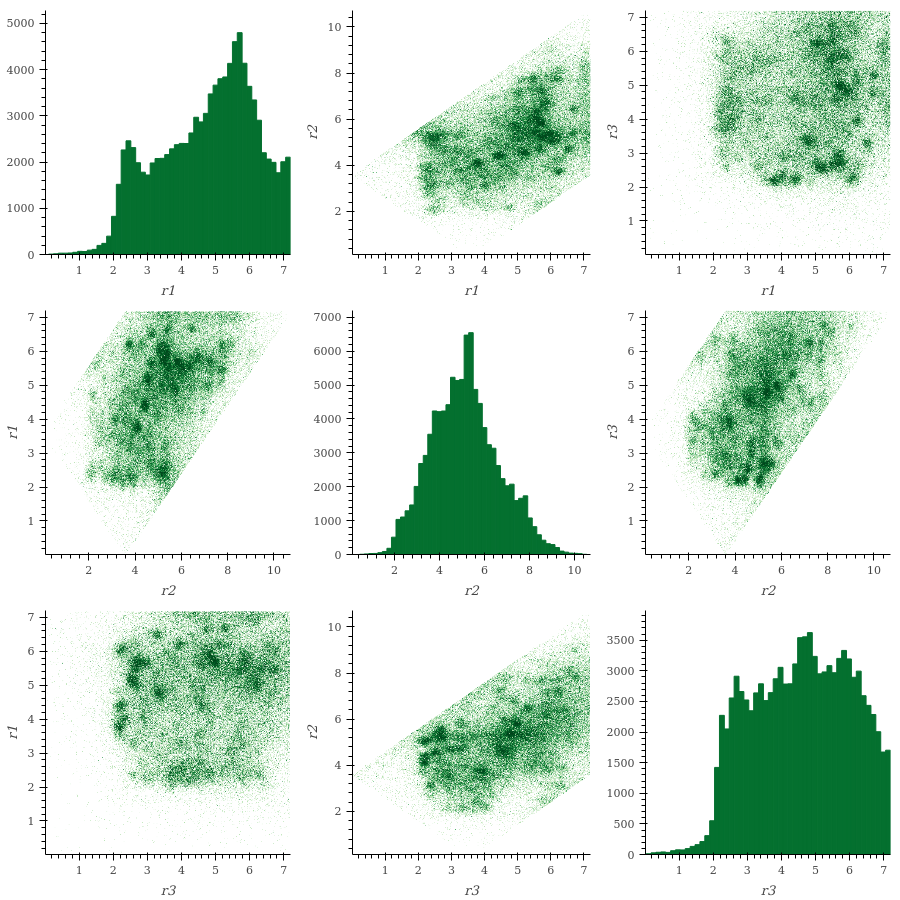}
    \caption{Plots of the univariate marginal and pairwise bivariate distributions of \num{1e5} samples from a reflected Brownian motion-based diffusion model in \(\mathbb{P}\subset\Rbb^3\).}
    \label{fig:app_pairplots_loop_polytope_reflected}
  \end{subfigure}
  \hfill
  \begin{subfigure}[t]{0.45\textwidth}
    \centering
    \includegraphics[width=\textwidth]{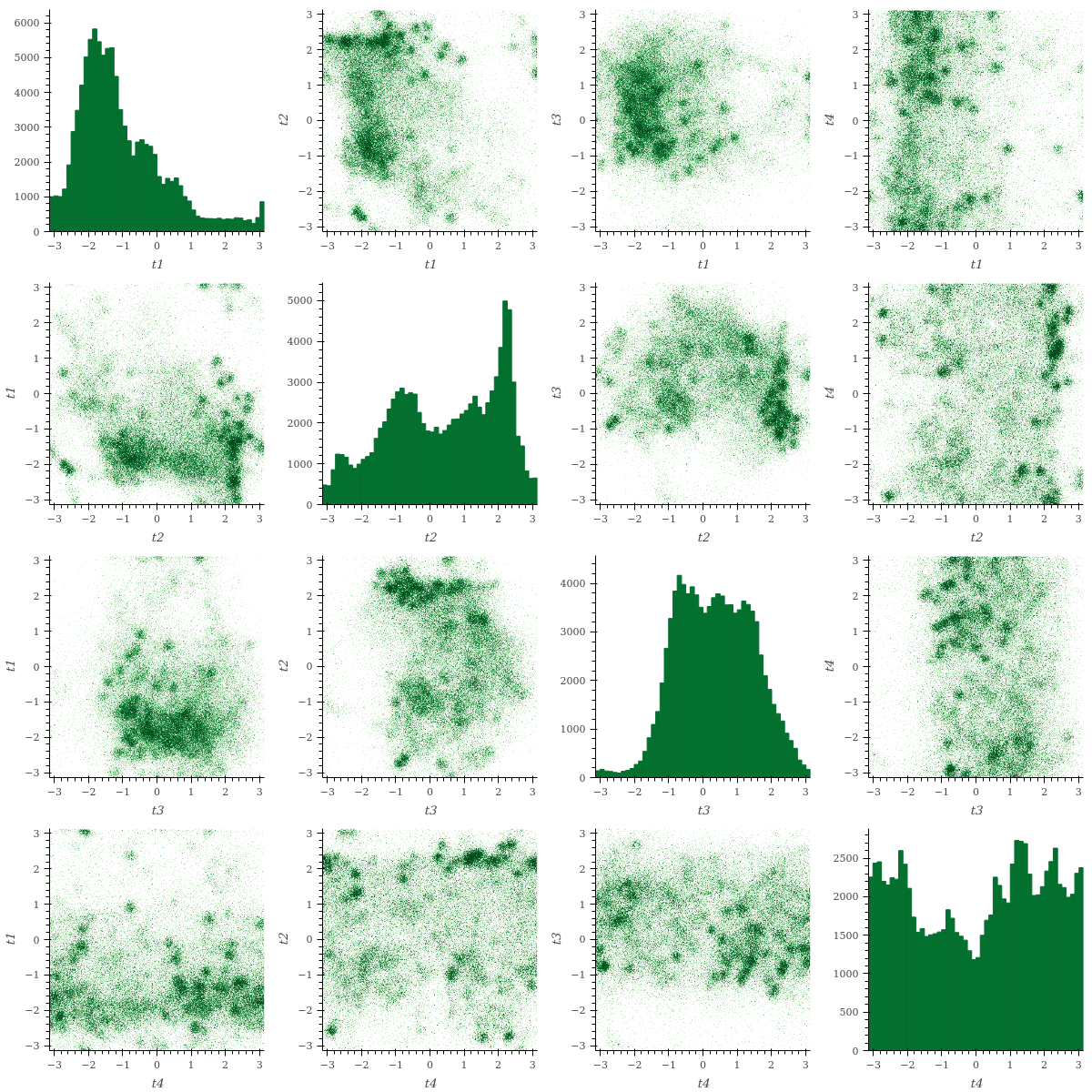}
    \caption{Plots of the univariate marginal and pairwise bivariate distributions of \num{1e5} samples from a reflected Brownian motion-based diffusion model in \(\mathbb{T}^4\).}
    \label{fig:app_pairplots_loop_torus_reflected}
  \end{subfigure}
  \caption{Visualisation of the distribution learned by a reflected Brownian motion-based diffusion model in \(\mathbb{P}\subset\Rbb^3\times\mathbb{T}^4\) using univariate marginal and pairwise bivariate plots.}
  \label{fig:app_pairplots_loop_reflected}
\end{figure}

\subsubsection{Visualisation of samples from the uniform distribution}

\begin{figure}[H]
  \centering
  \begin{subfigure}[t]{0.45\textwidth}
    \centering
    \includegraphics[width=\textwidth]{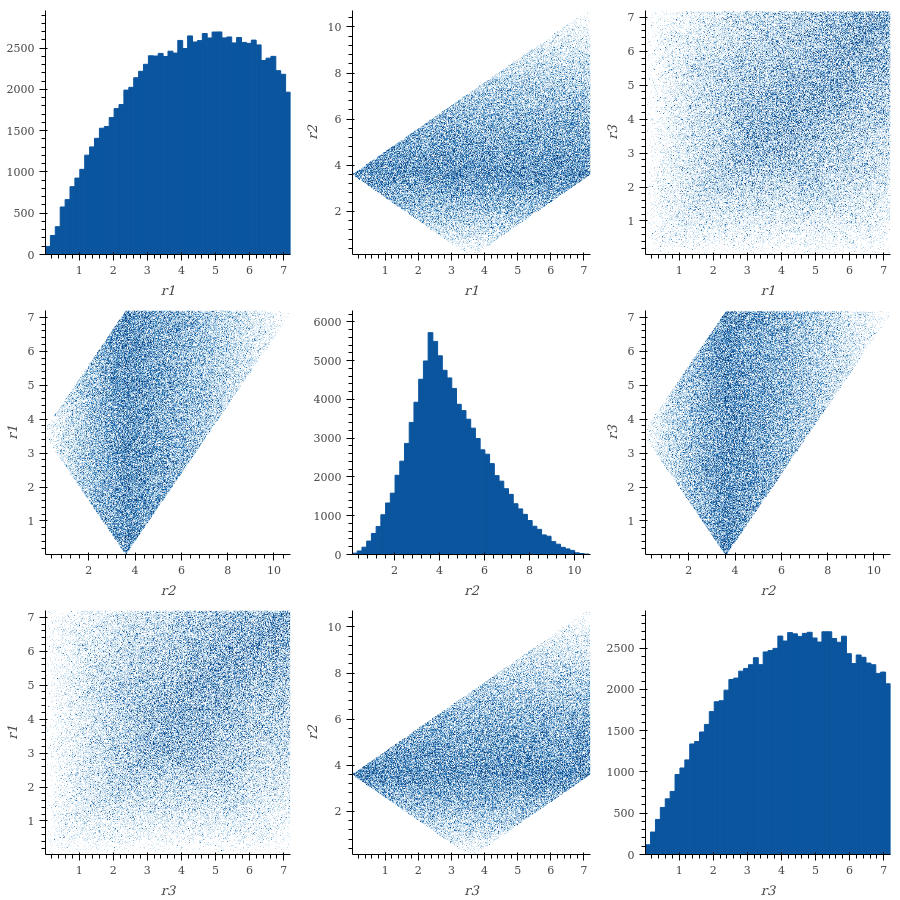}
    \caption{Plots of the univariate marginal and pairwise bivariate distributions of \num{1e5} samples from the uniform distribution in \(\mathbb{P}\subset\Rbb^3\).}
    \label{fig:app_pairplots_loop_polytope_uniform}
  \end{subfigure}
  \hfill
  \begin{subfigure}[t]{0.45\textwidth}
    \centering
    \includegraphics[width=\textwidth]{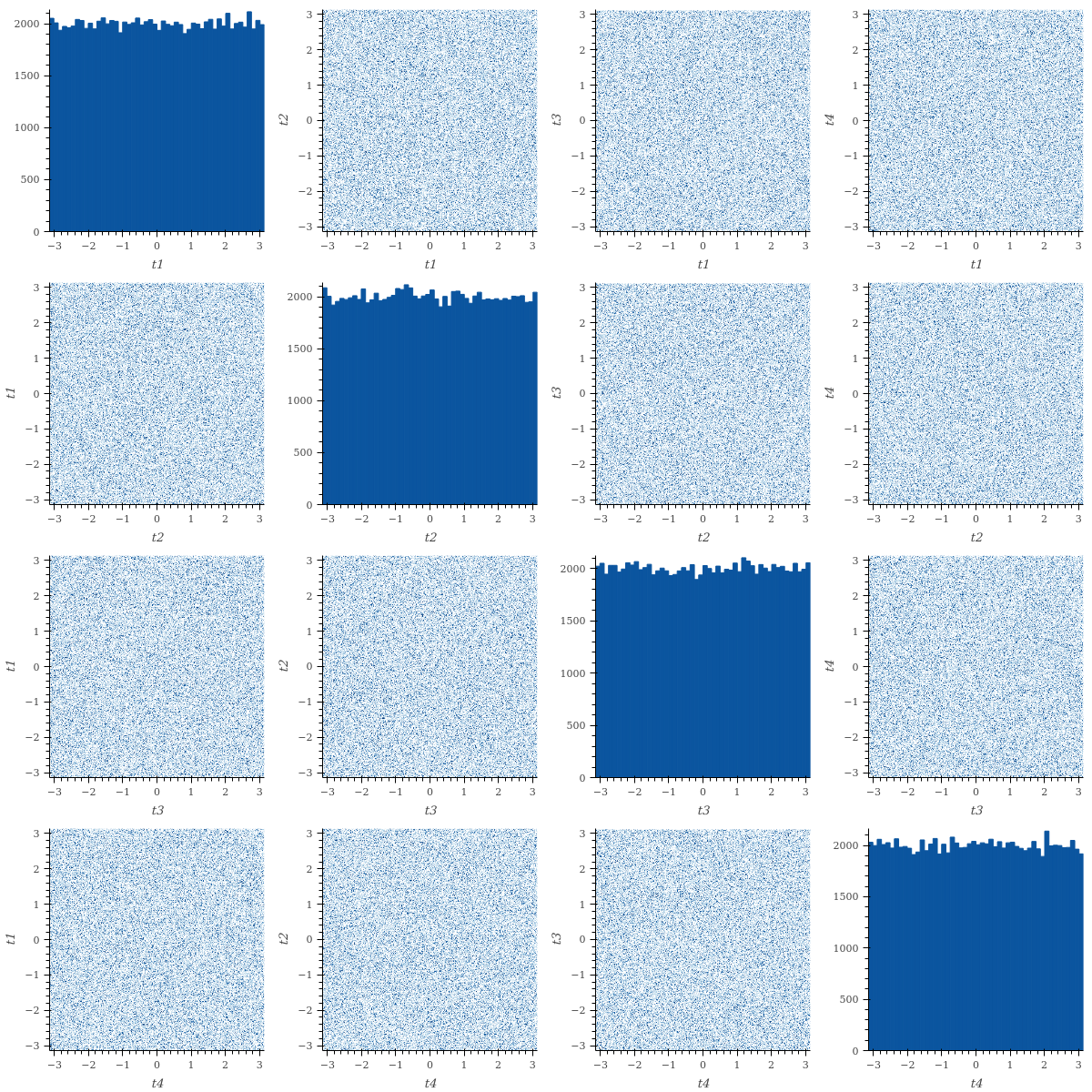}
    \caption{Plots of the univariate marginal and pairwise bivariate distributions of \num{1e5} samples from the uniform distribution in \(\mathbb{T}^4\).}
    \label{fig:app_pairplots_loop_torus_uniform}
  \end{subfigure}
  \caption{Visualisation of the uniform distribution in \(\mathbb{P}\subset\Rbb^3\times\mathbb{T}^4\) using univariate marginal and pairwise bivariate plots.}
  \label{fig:app_pairplots_loop_uniform}
\end{figure}

\end{document}
